\documentclass{article}

\usepackage{arxiv}

\usepackage{makecell, tabularx}
\usepackage[export]{adjustbox}

\usepackage{algorithm}
\usepackage{tabularx}
\usepackage{rotating}
\usepackage[algo2e]{algorithm2e} 
\usepackage{algpseudocode}
\usepackage{url}
\usepackage[table,xcdraw]{xcolor}
\usepackage{hyperref}
\usepackage[numbers]{natbib}
\usepackage{amsmath,amssymb}
\usepackage{bbm}
\usepackage{xcolor, soul}
\usepackage[utf8]{inputenc}
\usepackage[english]{babel}
\usepackage{graphicx}
\usepackage{booktabs}
\usepackage{multirow}
\usepackage{subcaption}
\usepackage{algpseudocode}
\usepackage{algorithm}
\usepackage[frozencache,cachedir=.]{minted}
\usepackage{tabularray}
\usepackage[font=footnotesize]{caption}  
\newcolumntype{C}[1]{>{\centering\arraybackslash}p{#1}}
\renewcommand{\shorttitle}{MineNetCD}

\usepackage{authblk}

\author[1,2]{Weikang Yu}
\author[3]{Xiaokang Zhang\thanks{Corresponding author}}
\author[1,4]{Xiao Xiang Zhu}
\author[2]{Richard Gloaguen}
\author[2,5]{Pedram Ghamisi}
\affil[1]{Technical University of Munich, 80333 Munich, Germany} 
\affil[2]{Helmholtz-Zentrum Dresden-Rossendorf (HZDR), 09599 Freiberg, Germany} 
\affil[3]{Wuhan University of Science and Technology, 430081 Wuhan, China}
\affil[4]{Munich Center for Machine Learning, 80539 Munich, Germany}
\affil[5]{Lancaster University, LA1 4YR Lancaster, U.K.}

\date{}    
\title{MineNetCD: A Benchmark for Global Mining Change Detection on Remote Sensing Imagery}

\begin{document}
\maketitle
\begin{abstract}
Monitoring changes triggered by mining activities is crucial for industrial controlling, environmental management and regulatory compliance, yet it poses significant challenges due to the vast and often remote locations of mining sites. Remote sensing technologies have increasingly become indispensable to detect and analyze these changes over time. We thus introduce MineNetCD, a comprehensive benchmark designed for global mining change detection using remote sensing imagery. The benchmark comprises three key contributions. First, we establish a global mining change detection dataset featuring more than 70k paired patches of bi-temporal high-resolution remote sensing images and pixel-level annotations from 100 mining sites worldwide. Second, we develop a novel baseline model based on a change-aware Fast Fourier Transform (ChangeFFT) module, which enhances various backbones by leveraging essential spectrum components within features in the frequency domain and capturing the channel-wise correlation of bi-temporal feature differences to learn change-aware representations. Third, we construct a unified change detection (UCD) framework that integrates over 13 advanced change detection models. This framework is designed for streamlined and efficient processing, utilizing the cloud platform hosted by HuggingFace. Extensive experiments have been conducted to demonstrate the superiority of the proposed baseline model compared with 12 state-of-the-art change detection approaches. Empirical studies on modularized backbones comprehensively confirm the efficacy of different representation learners on change detection. This contribution represents significant advancements in the field of remote sensing and change detection, providing a robust resource for future research and applications in global mining monitoring. Dataset and Codes are available via the \href{https://github.com/EricYu97/MineNetCD}{link}.
\end{abstract}


\section{Introduction}
%
%
%
%
\begin{figure}[!htp]
    \captionsetup{singlelinecheck=false}
    \centering
    \includegraphics[width=.5\linewidth]{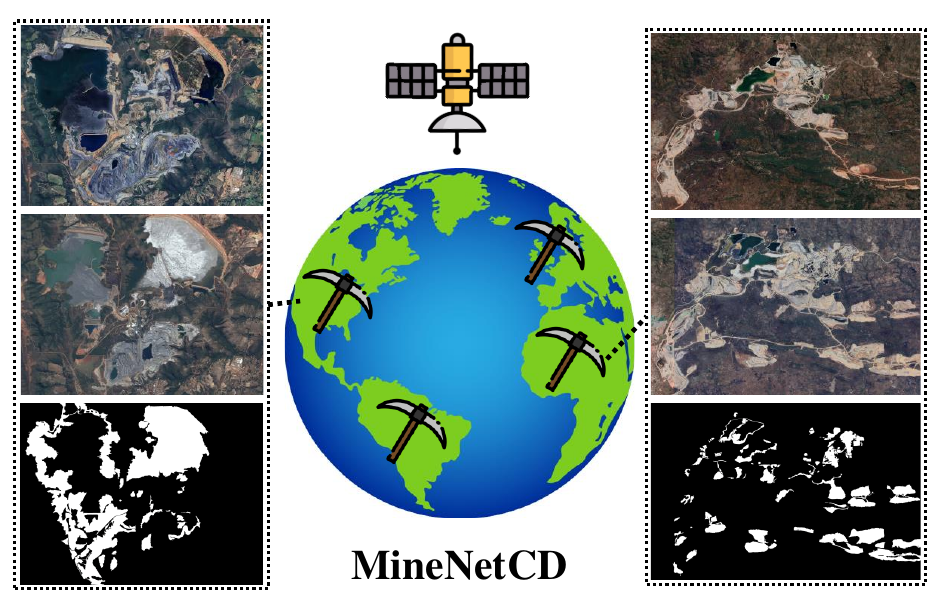}
    \caption{The MineNetCD.}
    \label{fig:intro}
\end{figure}
Mining operations are crucial to provide the materials required for infrastructure, technology, and wellbeing and ensure the energy transition but have significant environmental impacts \cite{sonter2017mining}. Open-pit mine areas, where minerals are extracted from a pit in the ground, induce large excavations and produce massive wastes in the form of tailings and waste piles during their life cycle \cite{du2022open,maus2020global}. These activities are closely linked to adverse impacts such as ecosystem fragmentation, degradation, geotechnical disasters, and biodiversity loss \cite{ sonter2020renewable}. Land degradation is a primary concern due to mining-induced alterations such as excavations and the removal of topsoil and vegetation. This results in soil erosion, landscape instability, and ecosystem disruption \cite{HE2020111742,CHEN2020111663}. Sustainability aims to minimize environmental impact and promote a better relationship between humans and our planet. While mining is inherently not sustainable, metal sourcing can be part of a sustainable society. Sourcing metals sustainably will require considering wellbeing as a purpose and preserving natural capital \cite{gloaguen2022mineral}. Sourcing the raw materials increasingly demanded by our societies will need transparent and inclusive stakeholder participation as well as a holistic understanding of the impact of extractive activities to reach sustainability status \cite{renn2022metal}. 
Geospatial data collected through satellite, drone, and aerial imagery plays a vital role in achieving metal sourcing sustainability by providing real-time, detailed perspectives on mining sites \cite{GALLWEY2020111970, WITKOWSKI2024113934}. 
Tools like machine learning and artificial intelligence (AI) help analyze this data to enhance the accuracy and speed of data analysis by transforming raw geospatial data into actionable insights \cite{9527092}. This improved analysis enables better anticipation and mitigation of environmental risks, thereby reducing mining's environmental footprint and aligning with the goals of AI for Social Good in building responsible AI \cite{ghamisi2024responsible}. 
Nowadays, remote sensing change detection technology using multitemporal images has become a key tool in tracking changes in the landscape caused by mining operations \cite{li2023multi,10002353,10329996}.
\begin{figure*}[h]
		\centering
		\includegraphics[width=0.99\linewidth]{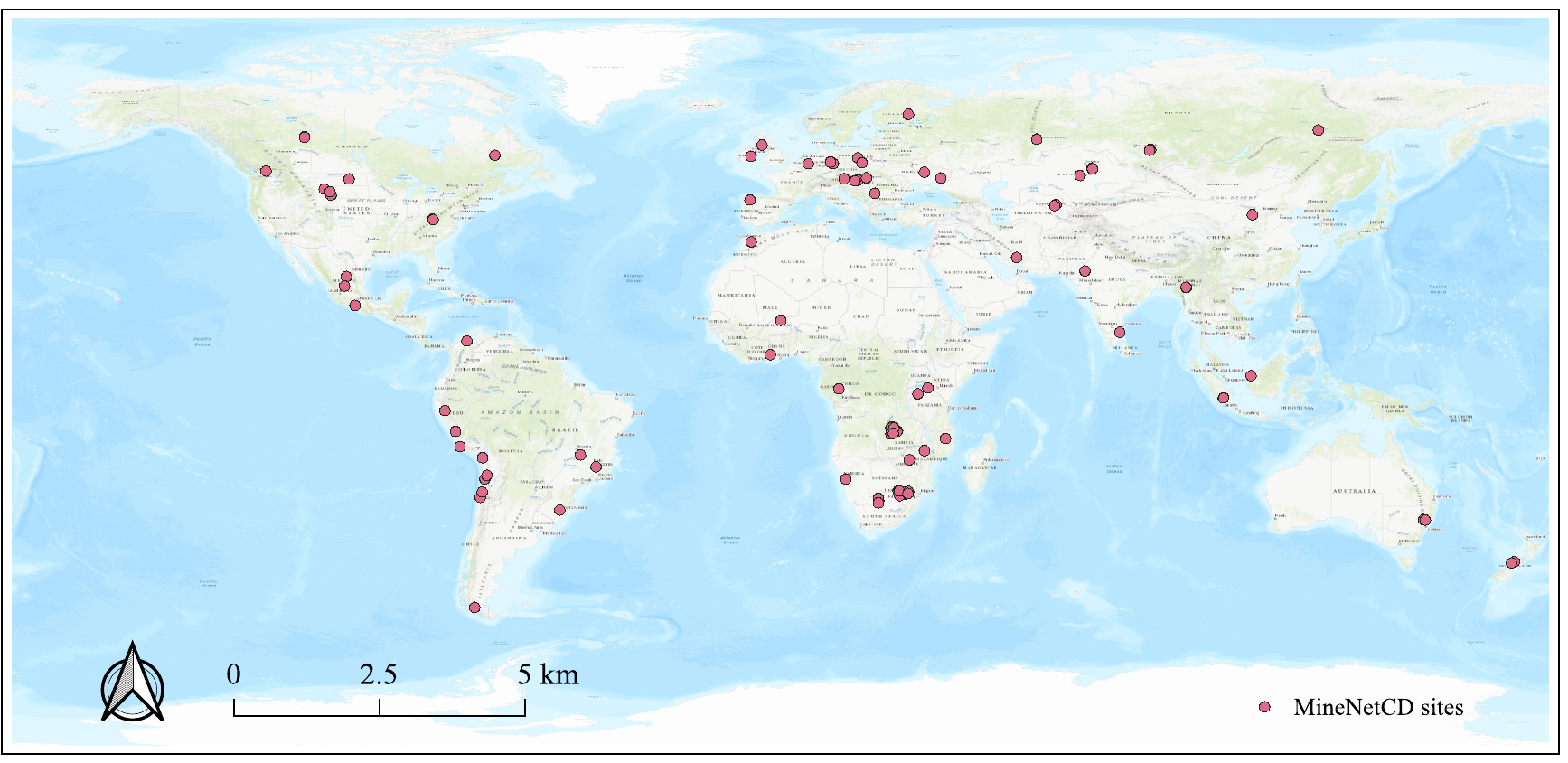}
		\caption{Spatial distribution of the mining sites in MineNetCD dataset}
		\label{MinPlaces}
\end{figure*}

Mining change detection aims to quantify the changes caused by mining operations on the Earth’s surface by comparing the multitemporal images acquired in the same geographical areas \cite{lv2021land,shi2020change}. It is essentially a dense labeling task for pixel-wise classification. In recent years, deep learning-based methods have emerged as the dominant approach for change detection, leveraging encoder-decoder structures based on convolutional neural networks (CNNs) and transformers \cite{8915802,9690575}. Nevertheless, the success of these models heavily relies on large-scale, high-quality training datasets, posing challenges in terms of data annotation and diversity. \citet{10002353} constructed the first open-pit mine change detection dataset, OMCD, with a resolution of $2$ m containing bi-temporal images from $39$ open-pit mining sites in China. On this basis, a CNN-based change detection method for open-pit mines has been investigated, in which the multiscale difference features of bi-temporal images were extracted and exploited by a Siamese network in an encoder-decoder structure \cite{li2023multi}. To address the issue of data scarcity, \citet{10329996} proposed a novel sub-instance data augmentation method based on a generative adversarial network to generate realistic and diverse samples.

For the change detection model, various strategies have been proposed to improve the feature representation of CNNs by leveraging multi-scale feature interaction \cite{chang2024dsfernet,zhang2020deeply,fang2021snunet,feng2023change}. On this basis, attention mechanisms have been applied to refine feature maps along the spatial and channel dimensions \cite{shi2021deeply,peng2020optical}. To handle the limitations of global context modeling in CNNs, recent methods have introduced transformers into change detection \cite{chen2021remote, Hong2023SpectralGPTSR}. They either take Transformer as their backbone or integrate it with CNNs to capture global contextual information \cite{bandara2022transformer,zhang2022multilevel,9736956}. Seeking to combine the strengths of CNNs and transformers, attention-based multiscale transformer networks have been investigated to extract global-local contextual information \cite{10114976,liu2023attention,yu2024maskcd}.
Overall, it is still challenging to model long-range contextual dependencies in an efficient manner while preserving the local details. Since mining change detection tasks focus on mining operation-driven land cover changes, the spectral variability poses more challenges to change-aware representation learning.

Open science has emerged as a vital paradigm in the development of artificial intelligence applications for Earth Observation (AI4EO), emphasizing transparency and the sharing of scientific research, data, and methodologies with the broader community. This involves not only disseminating final research outcomes but also sharing underlying data, models and codes.  To streamline these processes, several open-source change detection frameworks $\footnote{https://github.com/walking-shadow/Simple-Remote-Sensing-Change-Detection-Framework}$ $\footnote{https://github.com/likyoo/change\_detection.pytorch}$ have been introduced, aiming to facilitate the application of various change detection methodologies on widely used datasets. However, these frameworks often require additional dataset preprocessing, and sharing pre-trained models remains cumbersome, leading to complex and repetitive operations. Additionally, the availability of datasets and trained model weights is limited, making it challenging to reproduce the results of different approaches. Consequently, current frameworks still demand a high level of specialized skills, setting a relatively high bar for researchers interested in this field.

In this paper, we attempt to build a benchmark for global mining change detection and to prompt the achievement of mining sustainability goals. The main contributions of this paper are summarized as follows:
\begin{itemize}
\item The global mining change detection dataset contains more than 70k paired patches of bi-temporal high-resolution remote sensing images and pixel-level annotations acquired from $100$ mining sites worldwide. 
\item We developed a novel baseline model based on change-aware Fast Fourier Transform (ChangeFFT). It aids various backbones in leveraging key spectrum components within the features and capturing the channel-wise correlation of bitemporal features to learn the change-aware representations.
\item A unified change detection framework (UCD) is constructed by integrating more than 13 advanced change detection models. It is designed for a more streamlined and efficient process, leveraging the cloud platform HuggingFace. 
\end{itemize}

This paper is organized as follows. Section \ref{sec3} describes the proposed mining change detection dataset. After that, a new baseline model and a new change detection framework are introduced in Section \ref{sec4} and Section \ref{sec:UCD}, respectively. The experimental performance is evaluated in Section \ref{sec5}. Finally, Section \ref{sec6} concludes the paper.

  \begin{figure}[h]
		\centering
  \subfloat[]{\includegraphics[width=1\linewidth]{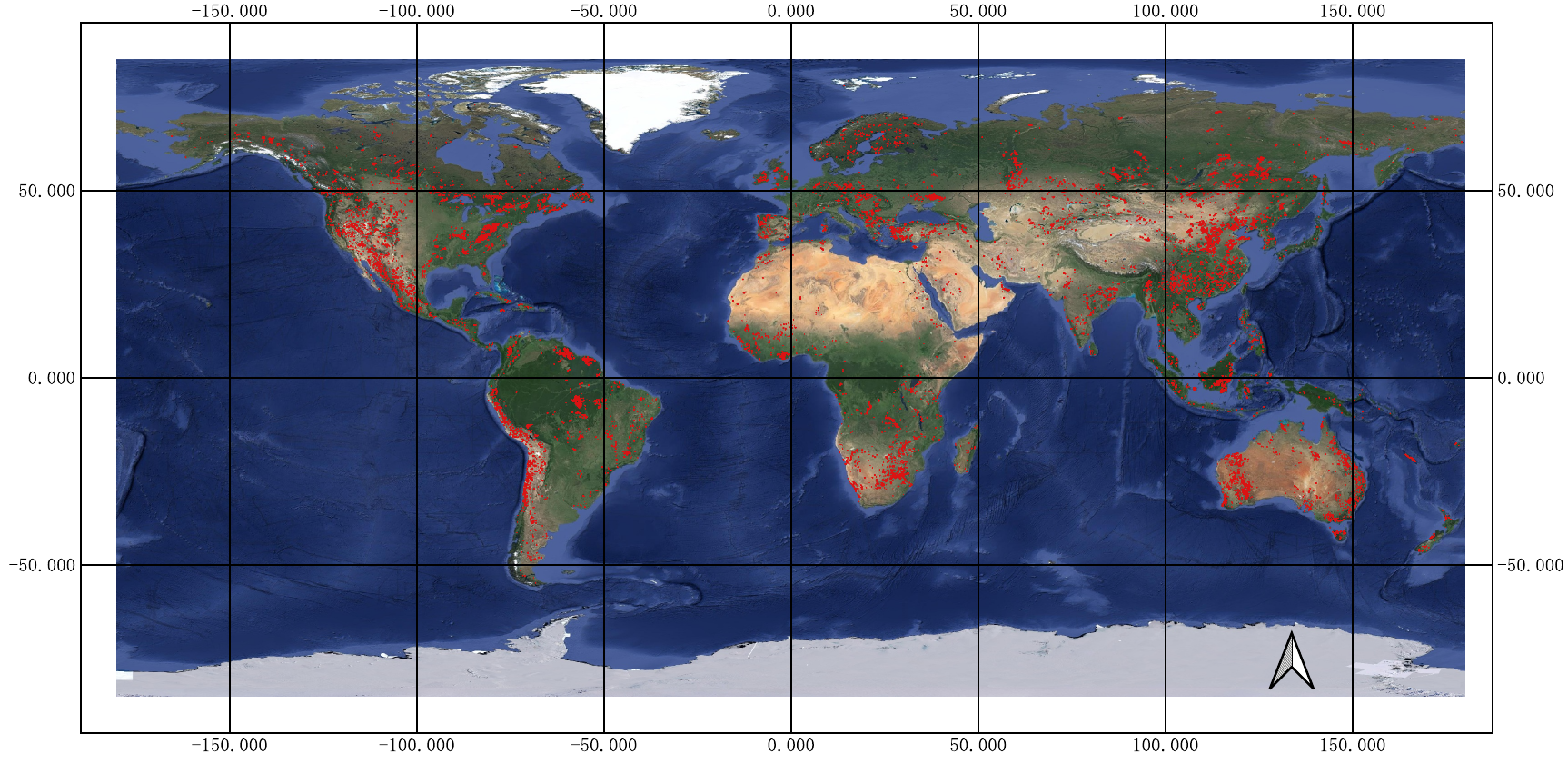}}
	\hspace{-2pt}
	\subfloat[]{\includegraphics[width=0.45\linewidth]{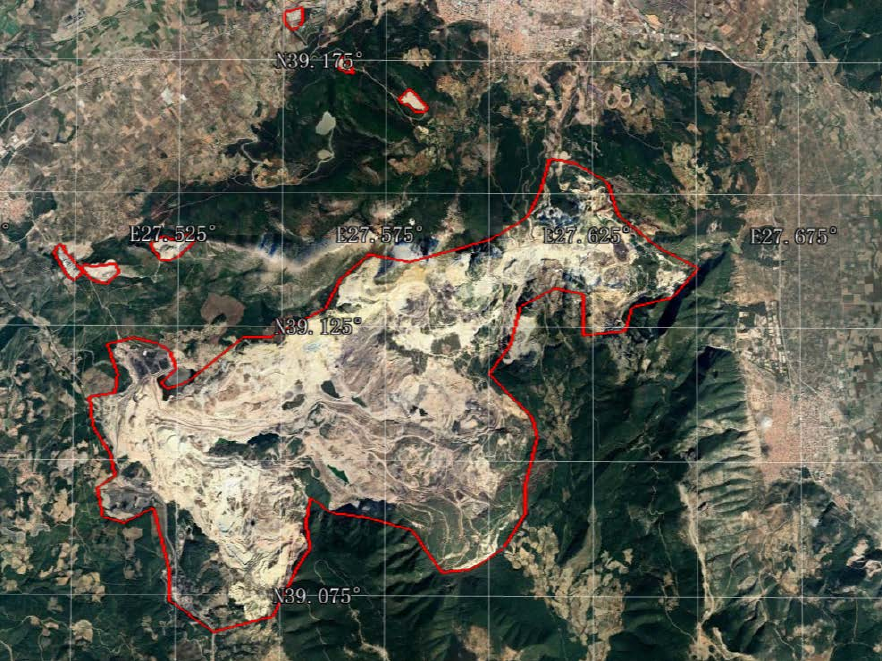}}
\hspace{-2pt}	
 \subfloat[]{\includegraphics[width=0.45\linewidth]{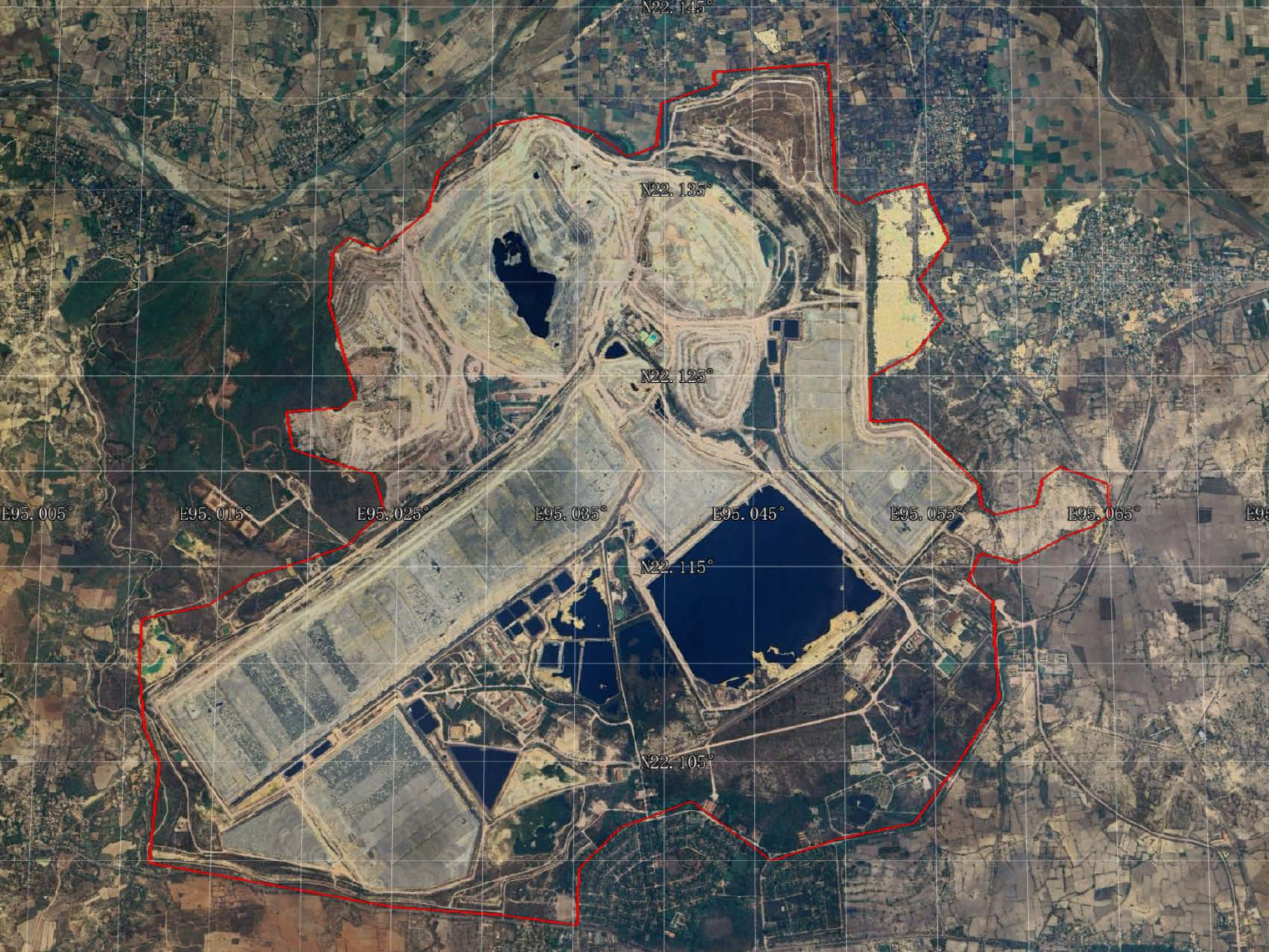}}
		\caption{To locate the mining sites: (a) Overlay the global mining polygon data \cite{maus2020global} onto Google Earth. This allows the sites to be accurately identified. (b) and (c) illustrate two such examples.}
		\label{sites}
	\end{figure}
 
\begin{figure}[h]
		\centering
	\includegraphics[width=0.7\linewidth]{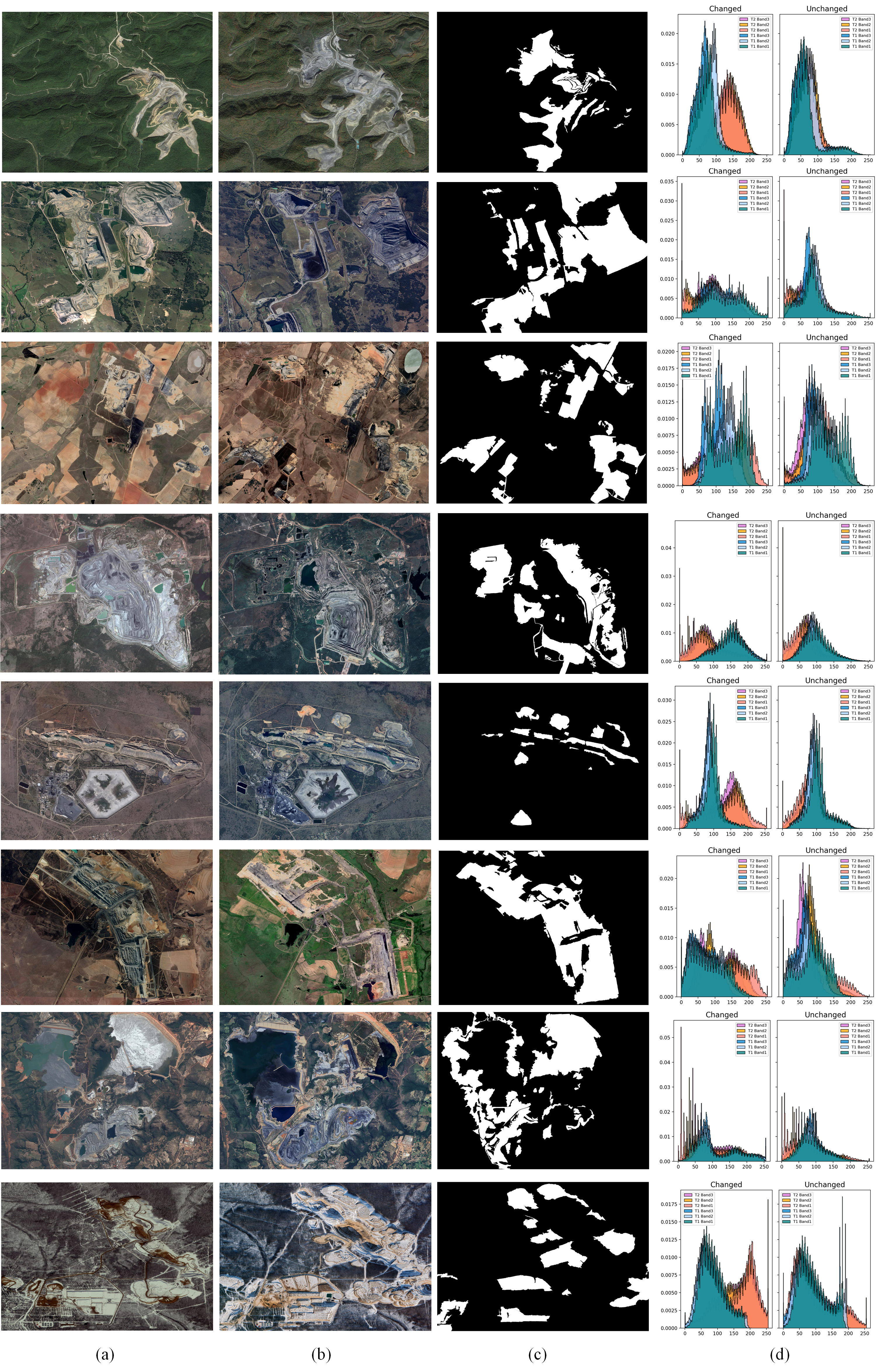}
		\caption{Typical samples in MineNetCD. (a) and (b) Pre- and post-event imagery. (c) Change masks. (d) Spectral intensity
histograms.}
		\label{MinSamples}
	\end{figure}
 
 \begin{figure}[h]
		\centering
  \subfloat[]{\includegraphics[width=.7\linewidth]{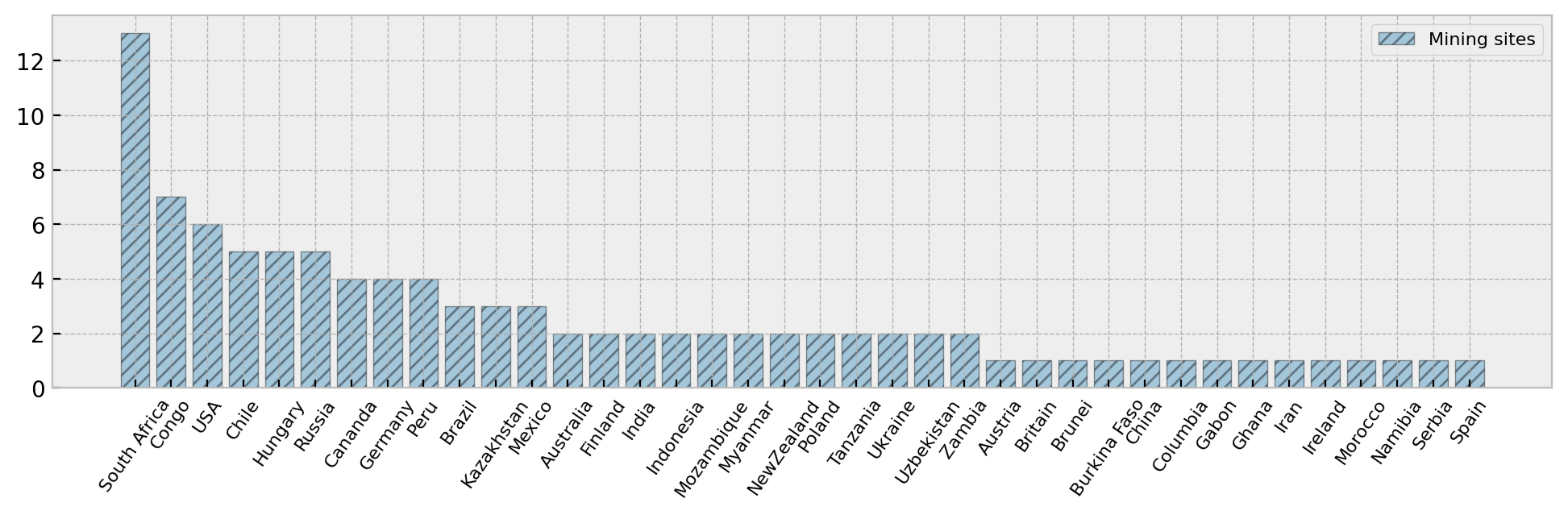}}
	\hspace{-2pt}
	\subfloat[]{\includegraphics[width=.7\linewidth]{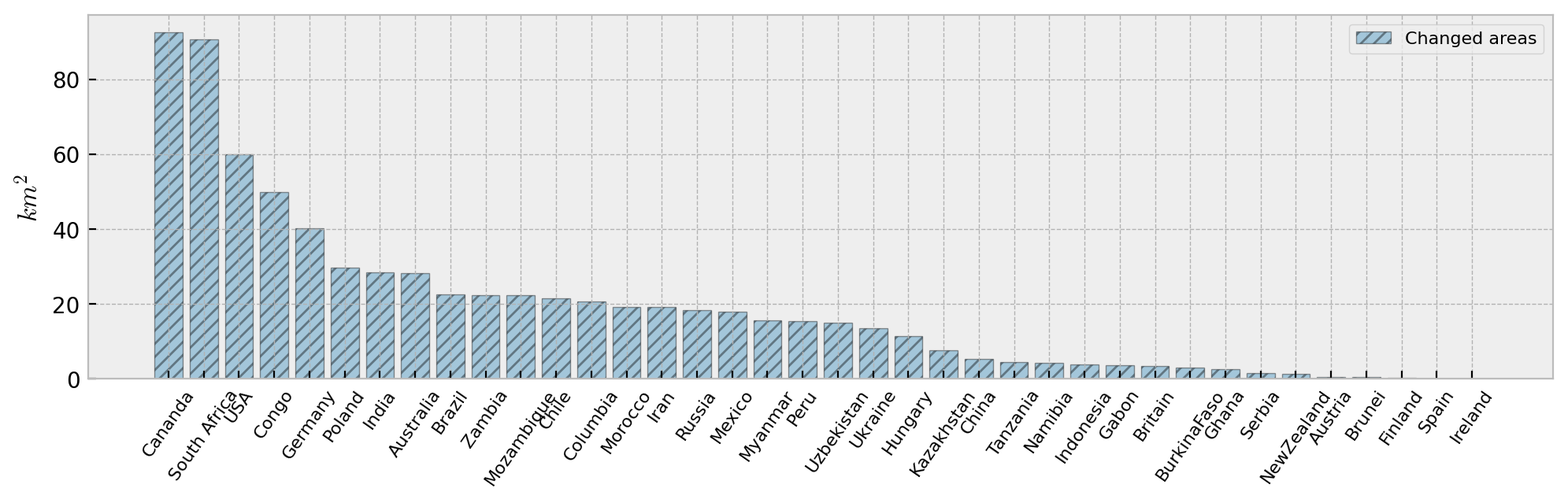}}
	\hspace{-2pt}
	\subfloat[]{\includegraphics[width=.7\linewidth]{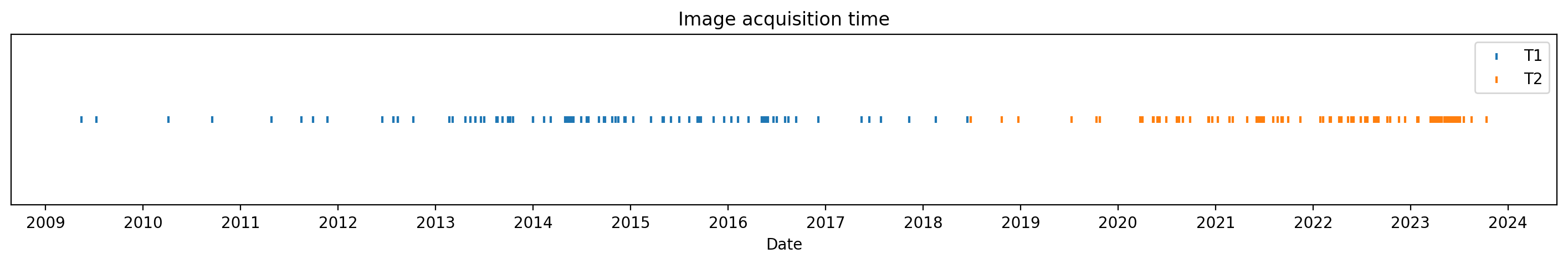}}
		\caption{Statistics of MineNetCD. (a) No. of mining sites in each country. Changed areas of (b) each country and (c) each site. (d) Image acquisition time.}
		\label{statistics}
	\end{figure}

\section{MineNetCD Dataset}\label{sec3}
Utilizing artificial intelligence and deep learning for open-pit mining detection from remote sensing images has demonstrated its potential to enhance the precision of open-pit mining detection under various environmental conditions. Nevertheless, the success of these models heavily relies on large-scale, high-quality training datasets, posing challenges in terms of data annotation and diversity. The main challenges in constructing a global mining change detection dataset are twofold. First, it is difficult to accurately locate mining sites in countries worldwide, hindering image acquisition. Second, compared with other scenes, dense labeling of mining changes is more challenging due to the spectral similarity between mining areas and other ground objects, as well as the irregular shapes and fragmented boundaries of changed areas.


We sought to present a global high-resolution open-pit mining change detection dataset, MineNetCD. 
It has been recorded that more than six thousand active mining sites exist globally \cite{maus2020global}. We intentionally selected $100$ representative mining sites from different geographical regions to build the benchmark. The spatial distribution of mining sites in the MineNetCD dataset is illustrated in Fig.~\ref{MinPlaces}. The sites were chosen based on changes in the mining area and their spatial distribution. We aimed to capture representative mining locations from around the world. By overlaying the mining polygons onto remote sensing images, we can precisely locate the mining sites and retrieve the desired images, as shown in Fig.~\ref{sites}.
This dataset has the following properties:
\begin{enumerate}
\item  The largest: MineNetCD covers a wide range of open-pit mining sites located in different countries and regions across $6$ continents. Each site involves bitemporal high-resolution imagery pairs with a spatial resolution of $1.2 m$ acquired via Google Earth Engine service.  MineNetCD consists of more than 70k paired patches that have total coverage of approximately $6756.88$ $\mathrm{km}^2$.

\item  High heterogeneity:  The dataset includes bi-temporal image pairs that cover mining sites of various sizes, shapes, and industries. Additionally, the changes within the images are driven by multiple factors, leading to significant spectral heterogeneity and intensity variations. This high-diversity dataset allows for the evaluation of the generalization performance of change detection models. Some samples are displayed in Fig.~\ref{MinSamples}. Fig.~\ref{statistics}(a) depicts the number of mining sites per country, highlighting South Africa, Congo, and the USA as having the highest counts, with South Africa leading significantly. Fig.~\ref{statistics}(b) quantifies the impacted land areas due to mining activities. Canada emerges as the most affected, with an extensive $80 km^2$ of altered landscape, followed by South Africa and Congo.

\item  Towards mining-driven changes: Unlike current research that primarily focuses on urban or building-related changes \cite{chen2020spatial,shi2021deeply}, this study examines land use/cover changes driven by mining operations, such as deforestation, erosion, and industrial expansion. Since land changes caused by mining are not frequent, the average time interval between the bi-temporal images is approximately $89.4$ months or about seven years, with a minimum interval of four years to ensure an obvious change pattern exists in the mining areas. As shown in Fig.~\ref{statistics}(c), the acquisition time of the pre-change images ranges from 2009 to 2018, while the post-change images were taken from 2018 to 2023.


\item  Fine-grained data:  Semi-automatic labeling is employed for image annotations by utilizing the eCognition developer software. First, the segmentation map is created by co-segmenting the stacked bitemporal images using the fractal net evolution approach (FNEA) to generate spatially corresponding segments. Subsequently, image interpretation experts manually label the changed areas by comparing the pre- and post-change images. This process ultimately yields pixel-wise annotations with fine-grained boundaries. More details can be observed in Fig.~\ref{MinSamples}.
\end{enumerate}

\section{MineNetCD Model}\label{sec4}
\begin{figure*}[!htp]
    \captionsetup{singlelinecheck=false}
    \centering
    \includegraphics[width=.8\textwidth]{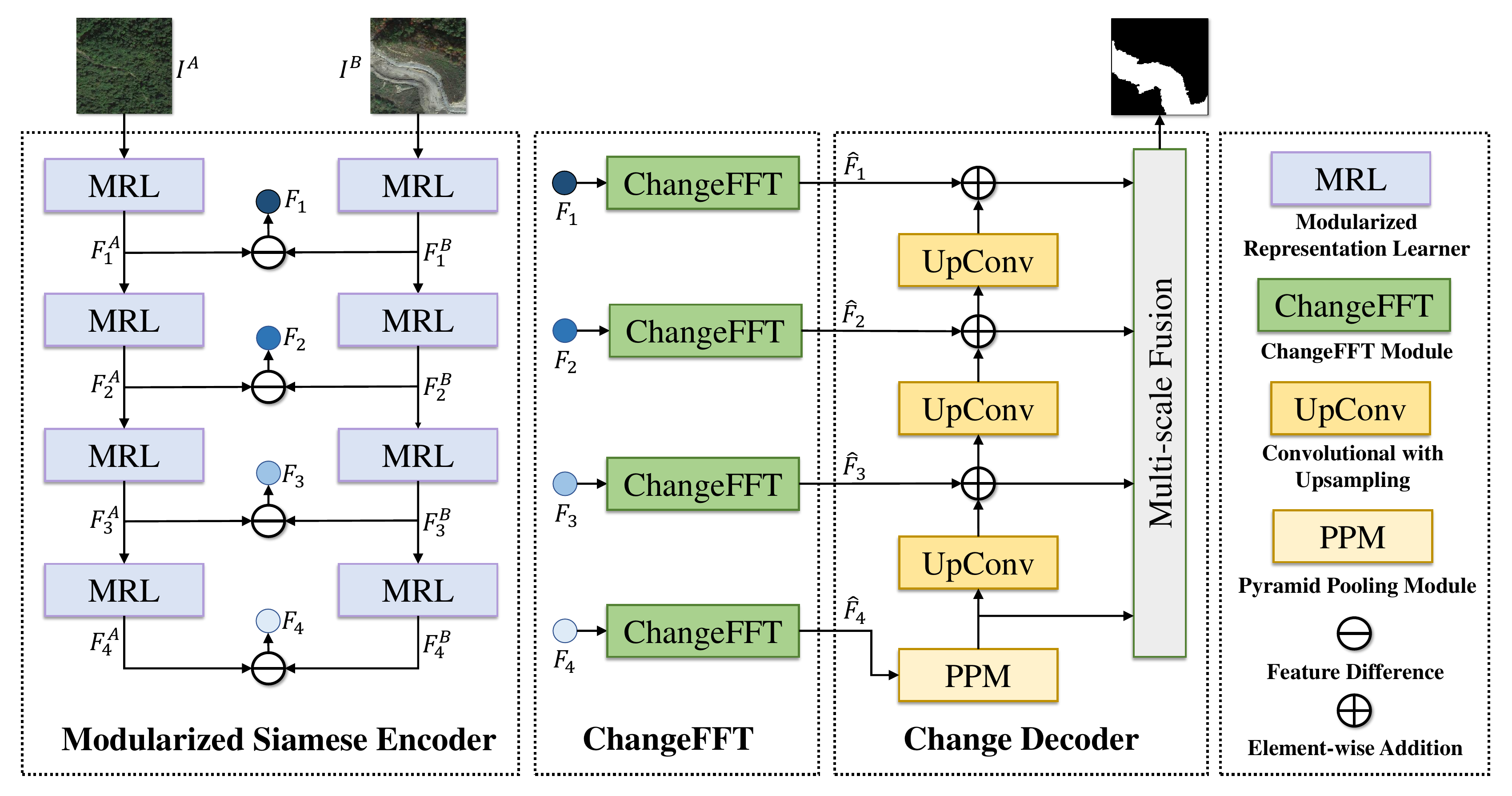}
    \caption{Proposed MineNetCD model for bi-temporal mining change detection.}
    \label{fig:changefft}
\end{figure*}

This section provides a detailed description of the proposed MineNetCD model, as shown in Fig \ref{fig:changefft}. First, the bi-temporal images are forwarded into a modularized Siamese encoder with two weight-shared backbones to extract the multi-level bi-temporal deep features, which are then subtracted into feature differences. After that, a change-aware fast Fourier transform (ChangeFFT) module is incorporated to enhance the multi-level feature differences into change-aware representations with a frequency domain learning strategy. Consequently, the enhanced feature maps are processed by a UperNet-based change decoder to generate a change map, in which the multi-level features are progressively processed and upsampled with a pyramid pooling module (PPM) and a feature pyramid network (FPN).

\subsection{Modularized Siamese Encoder}
\begin{figure}[!htp]
    \captionsetup{singlelinecheck=false}
    \centering
    \begin{subfigure}{.5\linewidth}
            \centering
            \includegraphics[width=\textwidth]{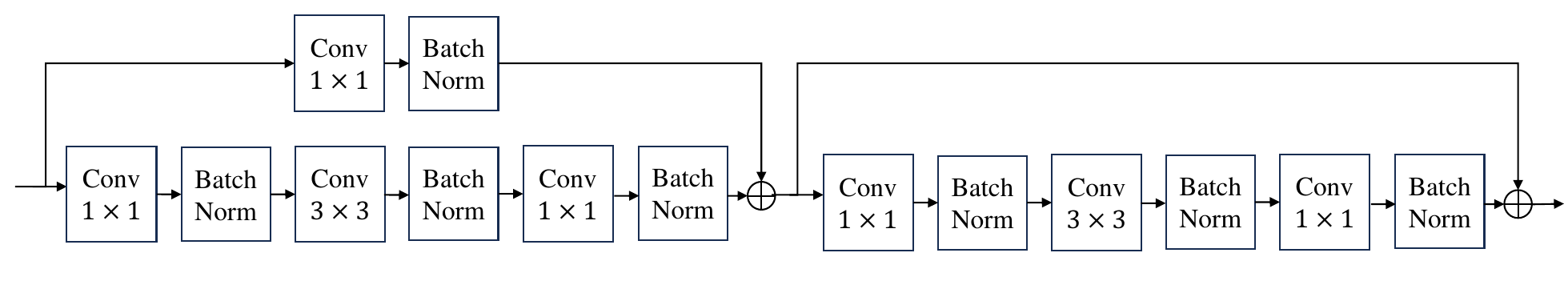}
            \caption{\centering ResNet Blocks}
    \end{subfigure}

    \begin{subfigure}{.5\linewidth}
            \centering
            \includegraphics[width=\textwidth]{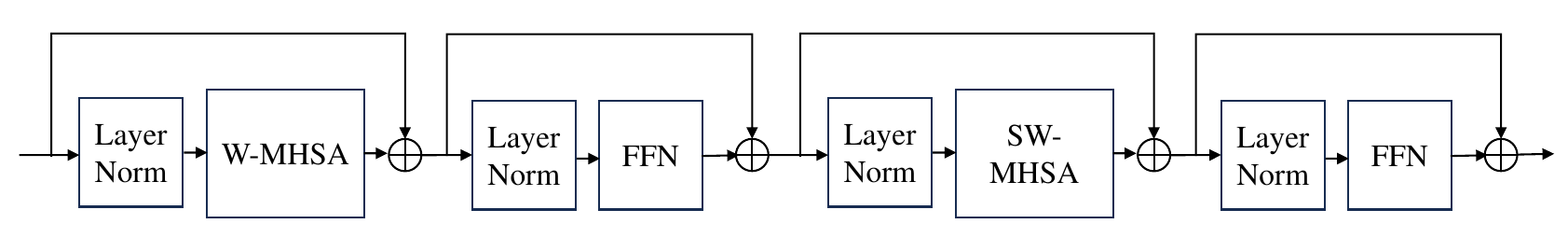}
            \caption{\centering Swin Transformer Blocks}
    \end{subfigure}

    \begin{subfigure}{.5\linewidth}
            \centering
            \includegraphics[width=\textwidth]{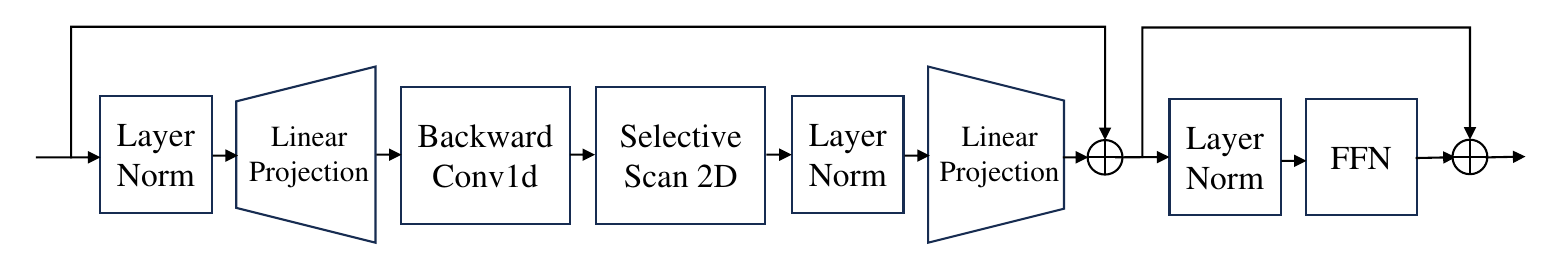}
             \caption{\centering VMamba Blocks}
    \end{subfigure}
    \caption{Illustration of different representation learners employed in the modularized Siamese encoder. The ResNet block consists of several convolutional layers with skip connections. The Swin-Transformer block utilizes a window multi-head self-attention (W-MHSA) and a subsequently shifted window multi-head self-attention (SW-MHSA) to capture the global dependencies and an MLP layer to enable channel-wise interactions. The VMamba block applies selective scan 2D to capture the contextual information and also an MLP layer for channel mixing.}
    \label{fig:rl}
\end{figure}
\begin{figure}[!htp]
    \captionsetup{singlelinecheck=false}
    \centering
    \includegraphics[width=.6\linewidth]{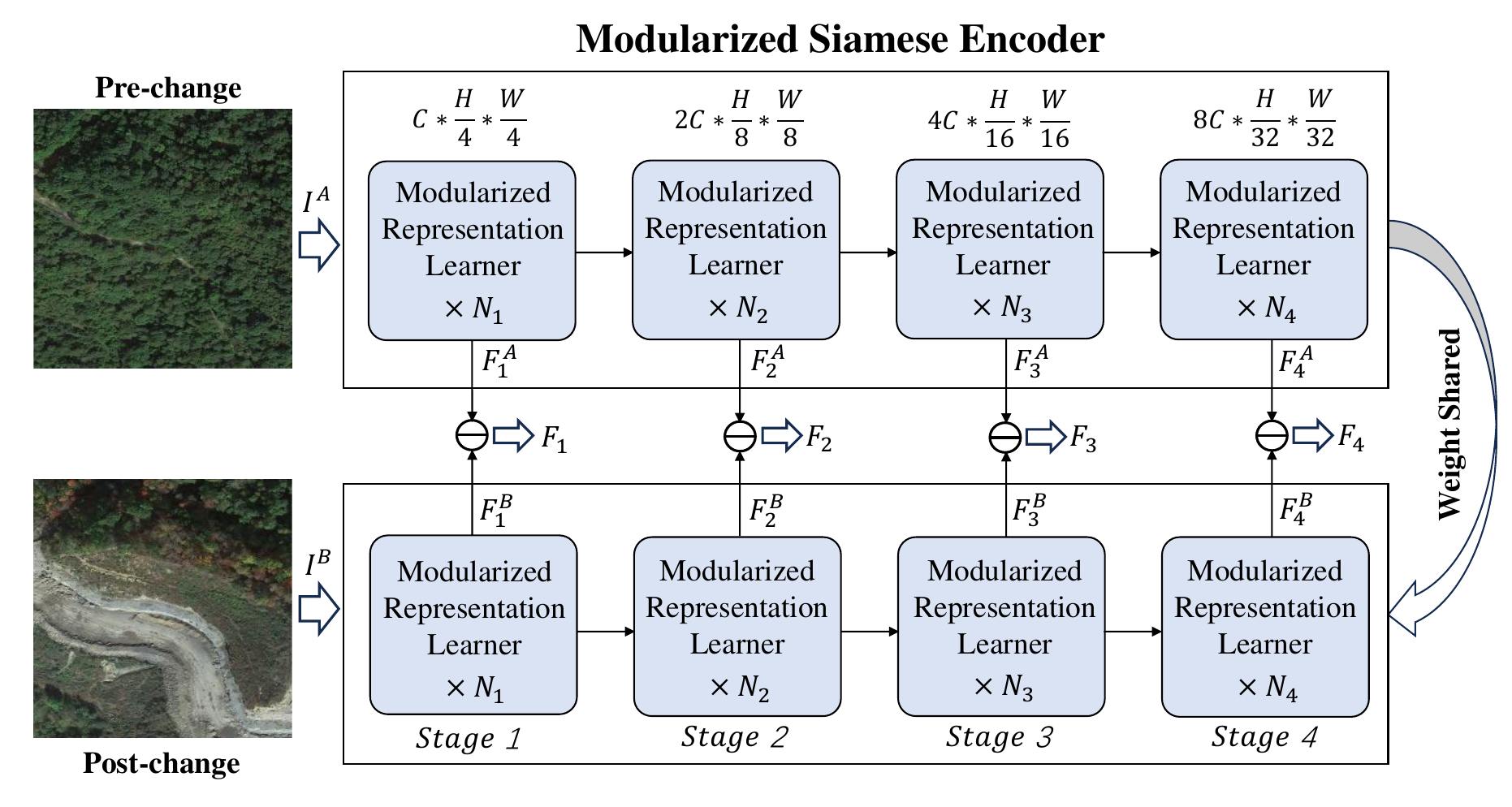}
    \caption{Illustration of the modularized Siamese encoder. The shapes of the output feature map of different levels are indicated on the top of the representation of learners.}
    \label{fig:encoder}
\end{figure}
In the literature, encoder-decoder-based change detection models typically use a fixed Siamese encoder to extract deep features from bitemporal images. This paradigm lacks scalability and adaptability for different change detection scenarios. To address this limitation, we propose a modularized Siamese encoder to build a more flexible change detection model. This encoder employs a weight-shared universal vision backbone with modularized representation learners that can be adjusted to different scales and incorporates various representation learning strategies. Consequently, our modularized Siamese encoder can adapt to heterogeneous remote sensing datasets with diverse scales and spectral characteristics.

Over the past decade, numerous vision backbones have been developed to enhance the efficacy of learning deep representations. These backbones can be broadly categorized into Convolutional Neural Networks (CNNs), Transformer-based models, and Mamba-based models, reflecting their architecture and  chronological development. As shown in Table \ref{table:backbone}, these backbones utilize convolutional networks, self-attention mechanisms, and selective state space models to learn deep representations, respectively. Furthermore, Fig. \ref{fig:rl} illustrates the architectures of the representation learners in these backbones. Each type of backbone has a distinct focus: CNNs concentrate on localized image areas, self-attention models capture long-range dependencies among all feature elements, and selective state space models reduce computational complexity for context modeling with a selective scan 2D mechanism.

In this paper, we select one representative model for each type of backbone: ResNet \cite{he2016deep} for CNNs, Swin Transformer \cite{liu2021swin} for transformer-based models, and VMamba \cite{zhu2024vision} for Mamba-based models. These vision backbones consist of four stages of representation learners, and we construct a modularized weight-shared Siamese structure to integrate these models to extract bi-temporal features simultaneously from remote sensing images, as shown in Fig. \ref{fig:encoder}. The $N_{1}$, $N_{2}$, $N_{3}$, and $N_{4}$ denote the number of blocks of the modularized representation learners at each stage of the encoder, which can be adjusted to construct the backbones with different scales. Let $I^{A}$ and $I^{B}$ be the pre-change and post-change remote sensing images, the pre-change and post-change deep features of four different levels $l=\{1,2,3,4\}$ are extracted by the modularized Siamese encoder as $\{F^{A}_{l}\}_{l=1}^{4}$ and $\{F^{B}_{l}\}_{l=1}^{4}$. After that, the multi-scale bi-temporal deep features are subtracted into feature differences $\{F_{l}\}_{l=1}^{4}$ as follows:
\begin{equation}\label{feature_diff}
    F_{l}=F_{l}^{A}-F_{l}^{B}.
\end{equation}

\begin{table}[]
\centering
\caption{Overview of the backbone models adopted in this study.}
\label{table:backbone}
\begin{tabular}{cccc}
\hline
Backbone Models & Learning Strategy & \#Parameters & FLOPs \\ \hline
\multicolumn{4}{l}{\textit{ResNet Series}} \\
ResNet-18 & \multirow{3}{*}{CNN} & 11M & 1.8G \\
ResNet-50 &  & 24M & 3.8G \\
ResNet-101 &  & 43M & 7.6G \\ \hline
\multicolumn{4}{l}{\textit{Swin Transformer Series}} \\
Swin-T & \multirow{3}{*}{Transformers} & 28M & 4.5G \\
Swin-S &  & 50M & 8.7G \\
Swin-B &  & 88M & 15.4G \\ \hline
\multicolumn{4}{l}{\textit{VMamba Series}} \\
VMamba-T & \multirow{3}{*}{Mamba} & 30M & 4.8G \\
VMamba-S &  & 50M & 8.7G \\
VMamba-B &  & 89M & 15.4G \\ \hline
\end{tabular}
\end{table}

Since these vision backbones have been widely adopted in a range of AI4EO applications, our objective is to conduct a comprehensive evaluation of these backbones for remote sensing change detection and validate the effectiveness of the proposed ChangeFFT module on various representation learners. Furthermore, we select three variants with different scales of parameters for each vision backbone to evaluate their scalability in the change detection task.

\subsection{Change-aware Fast Fourier Transform (ChangeFFT)}
\begin{figure}[!htp]
    \captionsetup{singlelinecheck=false}
    \centering
    \includegraphics[width=.5\linewidth]{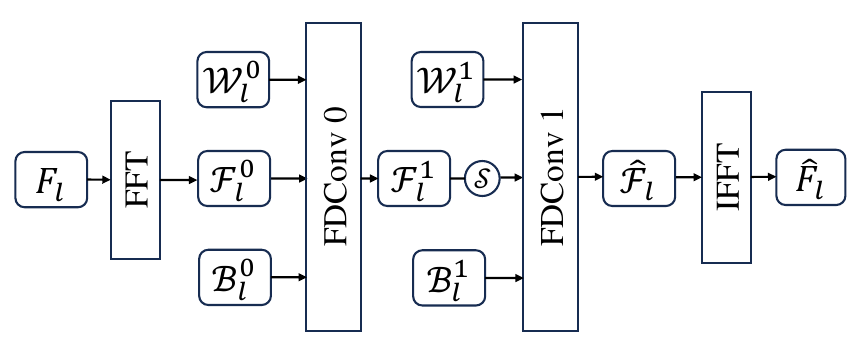}
    \caption{Illustration of the ChangeFFT module. The $\mathcal{S}$ denotes SiLU activation function.}
    \label{fig:changefftmodule}
\end{figure}
Currently, most encoder-decoder-based change detection approaches directly use feature differences as change representations to reconstruct a change map in the decoder. However, the subtraction operation for fusing bitemporal deep features can disrupt the well-learned deep representations due to the complex spatiotemporal relationships of the land surfaces, leading to challenges in distinguishing changes in the decoder. To overcome this challenge, we propose a change-aware Fast Fourier Transform (ChangeFFT) module that effectively learns change-aware representations from multi-level feature differences in the frequency domain.

Learning in the frequency domain has proven to be an efficient technique for capturing channel-wise dependencies of deep features \cite{qin2021fcanet}. By applying a Fast Fourier Transform (FFT), features are transformed from the time domain to the frequency domain, where change-aware representations can be precisely captured within the spectrum of feature differences. Let $\{F_{l}\}_{l=1}^{4} \in \mathbb{R}^{C_{l}\times H_{l} \times W_{l}}$ be the differenced features input, where $C_{l}$, $H_{l}$, and $W_{l}$ are the number of channels, height, and width of the $l$-th level of feature differences. The FFT can be calculated as follows:
\begin{align}
    \mathcal{F}_{l}(k)&=\int_{D}F(x)e^{-j2\pi kx}dx \notag\\ 
    =&\int_{D}F_{l}(x)cos(2\pi kx)dx+j\int_{D}F_{l}(x)sin(2\pi kx)dx \notag\\
    =&\mathcal{F}_{l}^{R}+j\mathcal{F}_{l}^{I},
\end{align}
where $k$ is the frequency variable and $x$ indexes the spatial coordinates. $\mathcal{F}_{l}^{R}$ and $\mathcal{F}_{l}^{I}$ is the real and imaginary part of the transformed feature $\mathcal{F}_{l}$ at the $l$-th level, respectively. Subsequently, two convolutional layers are applied to enable a change-aware representation learning in the frequency domain. According to the convolutional theorem, the Fourier transform of a convolution of two tensors is the pointwise product of their Fourier transforms, as follows: 
\begin{equation}
    \mathrm{FFT}\{H*G\}=\mathcal{H}\cdot \mathcal{G},
\end{equation}
where $*$ denotes the convolutional operator, and $\mathcal{H}$ and $\mathcal{G}$ are the Fourier transforms of tensors $H$ and $G$, respectively. On this basis, a convolutional layer in the frequency domain (FDConv) can be initialized with learnable parameters $\mathcal{W}$ and $\mathcal{B}$ in complex representing the weights and biases, respectively. Let $\mathcal{F}$ be a feature spectrum, the convolutional layer in the frequency domain can be computed as follows:
\begin{align}
    &\mathrm{FDConv}(\mathcal{F})=\mathrm{FFT}(F*W+B)=\mathcal{F}\cdot\mathcal{W}+\mathcal{B},\\ \notag 
    &=(\mathcal{F}^{R}\mathcal{W}^{R}-\mathcal{F}^{I}\mathcal{W}^{I}+\mathcal{B}_{R})+j(\mathcal{F}^{R}\mathcal{W}^{I}+j\mathcal{F}^{I}\mathcal{W}^{R}+\mathcal{B}_{I}),
\end{align}
where the superscript $R$ and $I$ denote the real and imaginary components of a complex tensor, respectively. As shown in Fig. \ref{fig:changefftmodule}, we apply two consecutive FDConvs to ensure a comprehensive change-aware representation learning for each level of the feature differences.

The FDConv can be viewed as a channel attention mechanism that learns the attention weights from the spectrum of feature differences in the frequency domain, which can be more effective in representing the channel-wise features than the global average pooling used in the traditional squeeze-and-excitation channel attention block. In the meantime, the number of parameters of the ChangeFFT module for each level is restricted to the amount of $C_{l}\cdot C_{l} +C_{l}$, which is relatively lightweight compared with other complicated channel-wise feature enhancement methods like multi-level perception (MLP). Our calculation indicates that the incorporation of the ChangeFFT module will only increase by about 10\% of the model parameters, indicating its ability to learn the change-aware representations in a balanced performance.

After the change-aware representation learning in two stages of FDConv, the obtained change representations are transformed back to the time domain with an inversed FFT (IFFT) function, as follows:
\begin{align}
    &\hat{F}_{l}(x)=\int_{D}\hat{\mathcal{F}}_{l}(k)e^{j2\pi kx}dk \notag\\ 
    &=\int_{D}(\hat{\mathcal{F}}_{l}^{R}(k)+j\hat{\mathcal{F}}_{l}^{I}(k))e^{j2\pi kx}dk,
\end{align}
where $\hat{\mathcal{F}}_{l}(k)$ is the spectrum output of the second stage of FDConv, as illustrated in Fig. \ref{fig:changefftmodule}.

Note that the FFT and IFFT are both performed along the channel dimension of the multi-scale features. In conclusion, the multi-level change-aware representations $\{\hat{F}_{l}\}_{l=1}^{4}$ are obtained from feature differences $\{F_{l}\}_{l=1}^{4}$ in the ChangeFFT module.
\subsection{Change Decoder}
To obtain a pixel-wise change map from the multi-scale change-aware representations, a multi-scale semantic decoder, named UperNet \cite{xiao2018unified}, is adopted that progressively fuses and upsample the multi-scale change-aware representations into a full-resolution change map. The UperNet has been widely employed as a plug-and-play semantic segmentor due to its superior compatibility with various vision backbones, which we considered also suitable for integrating with the modularized backbone and ChangeFFT module in this study. 

The UperNet change decoder processes multi-scale change-aware representations through a hierarchical approach that combines a Pyramid Pooling Module (PPM) and a Feature Pyramid Network (FPN). Initially, the PPM extracts global context information by applying pooling operations at multiple scales to the deepest change representations, creating a set of pooled features that capture various levels of contextual information. These pooled features are then concatenated and upsampled to match the original resolution. Subsequently, the FPN aggregates features from different layers of the backbone network, progressively refining and merging them to produce high-resolution feature maps. These multi-scale features from the PPM and FPN are then fused to generate a comprehensive feature representation, which is finally passed through convolutional layers to produce the change map, effectively leveraging both local details and global context for accurate change detection.

\subsection{Loss Function}
We employ a pixel-wise binary cross-entropy (BCE) loss $\mathcal{L}_{\mathrm{BCE}}$ to train the MineNetCD model. The BCE loss measures the discrepancy between the predicted change probabilities and the ground truths for each pixel as follows:
\begin{equation}
    \mathcal{L}_{\mathrm{BCE}}=-\frac{1}{N}\sum_{i=1}^{N}[y_{i}log(p_{i})+(1-y_{i})log(1-p_{i})],
\end{equation}
where $i$ indexes a pixel and $N$ is the number of pixels. $y_{i}$ denote a binary ground truth (0 for unchanged, 1 for changed) of pixel $i$, and $p_{i}$ is the predicted probability that pixel $i$ belongs to the changed class.

\section{Unified Change Detection Framework}\label{sec:UCD}
To promote deep learning-based change detection tasks for broader utilization by RS experts, we develop a unified change detection (UCD) framework based on the Huggingface platform with an open science prospect. The UCD framework currently includes more than 13 change detection approaches and 6 change detection datasets, including the MineNetCD dataset and model proposed in this paper. The UCD framework allows the user to simply specify the change detection method and dataset in the configuration file, on which the training and testing experiments can be automatically conducted without any professional skills. As a result, the change detection model can be initialized or loaded by using only one code line, respectively, as follows:
\begin{minted}{python}
    model=model.from_config(CONFIG)
    model=model.from_pretrained(MODEL_ID)
\end{minted}
where the CONFIG and MODEL\_ID are the model configurations and model paths, respectively. Furthermore, we leverage the Huggingface platform to facilitate the access of the dataset and model loading processes in a cloud-based paradigm. In particular, we uploaded the preprocessed datasets and pretrained model weights to the Huggingface Hub, which can be automatically loaded into the UCD framework by specifying their dataset and model ids in the configuration file. As a pioneer work, we have uploaded more than 70 model weights pretrained from 13 different approaches on 6 change detection datasets to our Huggingface Hub \url{https://huggingface.co/ericyu}. For future research, these pretrained models can be utilized to reproduce the results on these datasets, which avoids the efforts for re-training and maximizes the experiments' transparency.

The UCD is also designed with high compatibility that allows the users to easily integrate their own resources into the framework. We provide guidance and some tools for the users to preprocess and adapt their own datasets and models into standard formations that can be used in the UCD framework.
We hope this new approach to the UCD framework will motivate future change detection researchers to participate in these open science activities, as well as provide an opportunity for everyone with an interest to have a try in change detection without any professional skills.

\section{Experiments}\label{sec5}

\subsection{Dataset Preparation}
To facilitate the training and testing of the baseline models, the bi-temporal remote sensing image pairs from MineNetCD are cropped into a size of $256\times 256$. 
We randomly selected 60$\%$, 10$\%$, and 30$\%$ of mining sites as the training, validation, and testing sets, respectively. For computation convenience, the bi-temporal images and labels are cropped into patches of size $256 \times 256$. Finally, the training, testing, and validation sets include a total of $47,743$, $19,355$, and $4,613$ patches, respectively.

In accordance with the open-science paradigm of the proposed UCD framework, the MineNetCD dataset can be easily accessed and integrated into the framework via \url{https://huggingface.co/datasets/ericyu/MineNetCD256}. 
\subsection{Baseline Approaches}
The proposed MineNetCD baseline model is compared against several state-of-the-art deep learning-based RS-CD methods: a lightweight network with progressive aggregation and
supervised attention (A2Net) \cite{Li_2023_A2Net}, a bitemporal image transformer (BIT) network \cite{chen2021remote}, a transformer-based Siamese network for change detection (ChangeFormer) \cite{bandara2022transformer}, a dual-branch multi-level inter-temporal network (DMINet) \cite{feng2023change}, fully convolutional siamese networks for change detection (FC-EF) \cite{daudt2018fully}, a fully convolutional network within pyramid pooling (FCNPP) \cite{lei2019landslide}, an intra-scale cross-interaction and inter-scale feature fusion network (ICIFNet) \cite{feng2022icif}, a region detail preserving network (RDPNet) \cite{chen2022rdp}, a residual U-Net (ResUnet) \cite{9553995}, a fully convolutional Siamese concatenated U-Net (SiamUnet-Conc)  \cite{daudt2018fully}, a fully convolutional Siamese difference U-Net (SiamUnet-Diff) \cite{daudt2018fully}, an integrated Siamese network and nested U-Net (SNUNet) \cite{fang2021snunet}. 
\subsection{Experimental Settings}
The proposed MineNetCD model is implemented using the open-source \textit{transformers} package \cite{wolf-etal-2020-transformers} based on the UCD platform. For the model training, the Adam optimizer was adopted with an initial learning rate of $1e-4$, a momentum of $0.9$, and parameters $\beta_{1}, \beta_{2}$ as $0.9$ and $0.99$, respectively. The batch size was set to $32$ for each GPU.  In addition, we employ a cosine annealing scheduler that gradually reduces the learning rate to $1e-7$ for better model convergence.

In our experiments, the hyper-parameters for the methods considered for comparison were set to the default values according to their original publications. 
To ensure the reproduction of the experiments, the random seed is set to $8888$.
All experiments were conducted with a High-performance computing (HPC) system with a 128-core CPU and 8 NVIDIA Tesla A100 GPUs (40GB of RAM). In addition, the Accelerate \cite{accelerate} package is adopted for fully sharded data-parallel computing to speed up the computational of the models in our multi-GPU environment.

\begin{table*}[]
\centering
\caption{Quantitative Comparisons in terms of OA, Pre, Rec, F1, and cIoU on MineNetCD dataset. The best and second-best results are highlighted in red and blue, respectively.}
\label{table:quantitative}
\begin{tabular}{@{}m{3cm}<{\centering}|m{1.3cm}<{\centering}m{1.3cm}<{\centering}m{1.3cm}<{\centering}m{1.3cm}<{\centering}m{1.3cm}<{\centering}@{}}
\toprule
Methods & OA & Recall & Precision &F1  &  cIoU\\ \midrule
A2Net & {\color[HTML]{0000FF} 0.9188} & {\color[HTML]{0000FF} 0.7218} & 0.5766 & {\color[HTML]{0000FF} 0.6410} & {\color[HTML]{0000FF} 0.4718} \\
BIT & 0.9118 & 0.6732 & {\color[HTML]{0000FF} 0.5804} & 0.6233 &  0.4528\\
DMINet & 0.8967 & 0.6266 & 0.4412 & 0.5177  & 0.3493 \\
RDPNet & 0.8774 & 0.5128 & 0.4818 & 0.4968 & 0.3305 \\
ICIFNet & 0.8920 & 0.5966 & 0.4342 & 0.5024 & 0.3356 \\
SNUNet & 0.8927 & 0.5829 & 0.5124 & 0.5452 & 0.3750 \\
ChangeFormer & 0.8705 & 0.4857 & 0.516 & 0.5003 &  0.3337\\
SiamUnet-Conc & 0.8984 & 0.6469 & 0.4220 & 0.5108 &  0.3430\\
SiamUnet-Diff & 0.8960 & {\color[HTML]{FF0000} 0.7679} & 0.2478 & 0.3746 &  0.2305\\
ResUnet & 0.8671 & 0.4753 & 0.5512 &0.5104  & 0.3427 \\
FC-EF & 0.8836 & 0.5628 & 0.3293 & 0.4154 & 0.2622 \\
FCNPP & 0.8559 & 0.4022 & 0.3024 & 0.3452 & 0.2087 \\ \midrule
MineNetCD (Ours) & {\color[HTML]{FF0000} 0.9251} & 0.7120 & {\color[HTML]{FF0000} 0.6814} & {\color[HTML]{FF0000} 0.6963} & {\color[HTML]{FF0000} 0.5343} \\\bottomrule
\end{tabular}
\end{table*}
\subsection{Evaluation Metrics}
For quantitative evaluation, we adopt five commonly used metrics: overall accuracy (OA), precision (Pre), recall (Rec), F1-score, and change-class Intersection over Union (cIoU). The OA represents the percentage of pixels that are correctly detected among all samples. The precision is the number of correctly detected changed pixels divided by the number of all the pixels that are identified as changed in the change detection map. The recall is the percentage of correctly detected changed pixels among all the pixels that should be detected as changed. The F1-score is the harmonic mean of the precision and recall that also considers the class imbalance problem in change detection, and the cIoU reveals the exact coverage of the changed area of the detected change map compared with the ground truth. They are calculated according to the following equations:
\begin{align}
\mathrm{OA}&=(\mathrm{TP}+\mathrm{TN})/(\mathrm{TP}+\mathrm{TN}+\mathrm{FP}+\mathrm{FN})\\
\mathrm{Pre}&=\mathrm{TP}/(\mathrm{TP}+\mathrm{FP}),\\
\mathrm{Rec}&=\mathrm{TP}/(\mathrm{TP}+\mathrm{FN}),\\
\mathrm{F1}&=2\mathrm{TP}/\mathrm{(2TP+FP+FN)},\\
\mathrm{cIoU}&=\mathrm{TP}/(\mathrm{TP}+\mathrm{FP}+\mathrm{FN}),
\end{align}
where the TP is the number of correctly detected changed pixels, TN is the number of correctly detected unchanged pixels, FP is the number of false alarms, and FN is the number of missed change pixels.
\subsection{Benchmark Results with the State-of-the-art}
\subsubsection{Quantitative Results}
Table \ref{table:quantitative} displays the quantitative evaluation comparisons of different methods in terms of OA, Precision, Recall, F1 score, and cIoU on the MineNetCD dataset. The results indicate that the MineNetCD dataset is challenging for current change detection methods, with an average F1 score of less than $0.65$ and a cIoU of less than $0.48$. Although some methods achieve promising recall values, they suffer from low precision, which leads to a lot of false alarms. On the contrary, the proposed model can achieve an F1 score of $0.6963$ and a cIoU of $0.5343$, which gains an improvement of at least $0.0553$ and $0.0625$ in terms of F1 score and cIoU, respectively, compared to all the competitors. 
The overall results demonstrate the effectiveness and superiority of the proposed MineNetCD model.

\subsubsection{Qalitative Results}
\begin{figure*}
\hspace{1pt}
\subfloat[Pre-change]{%
              \includegraphics[width=.193\linewidth]{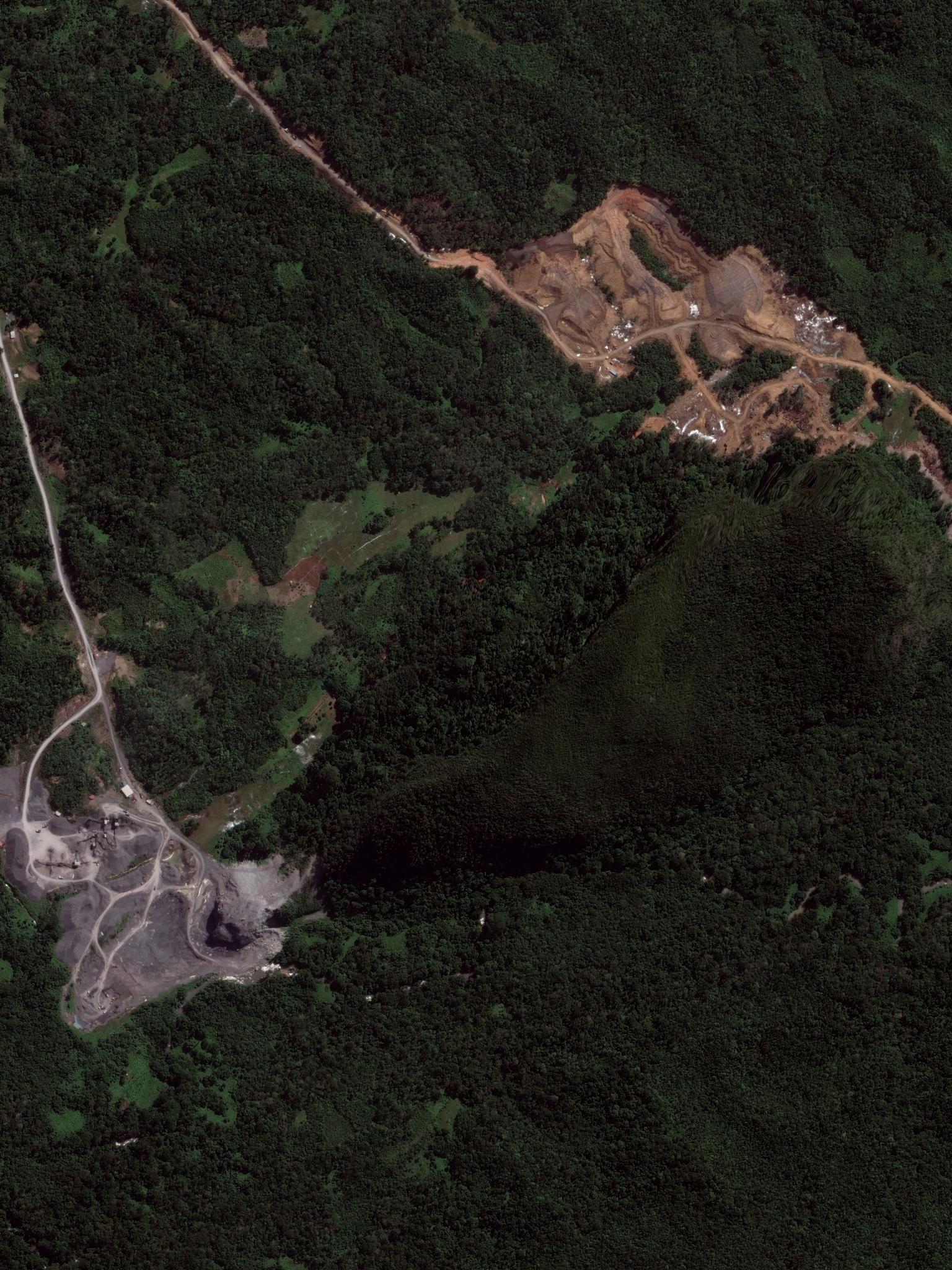}%
              \label{fig1:pre}%
           } 
\hspace{1pt}
\subfloat[Post-change]{%
              \includegraphics[width=.193\linewidth]{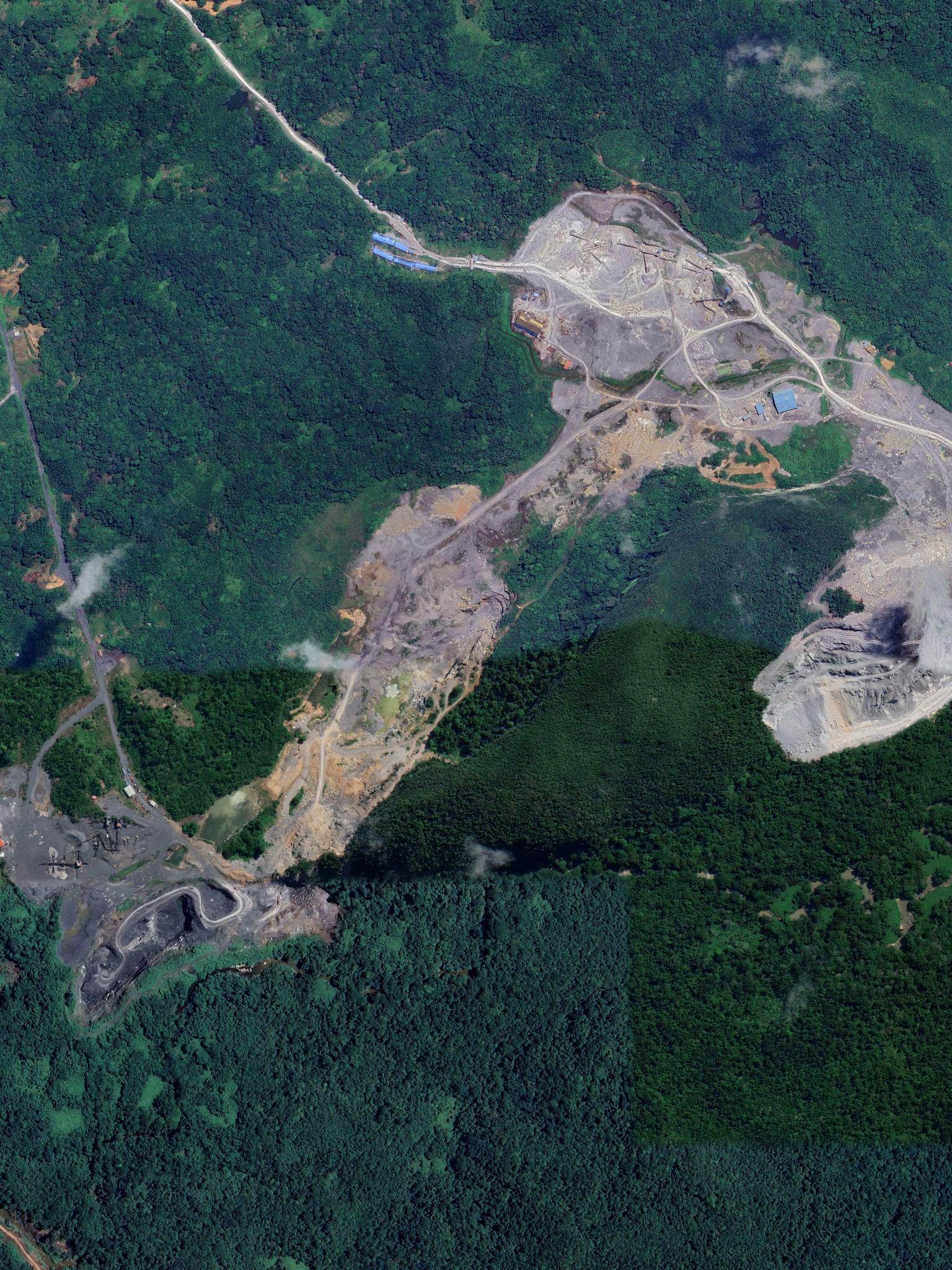}%
              \label{fig1:post}%
           } 
\hspace{1pt}
\subfloat[A2Net]{%
              \includegraphics[width=.193\linewidth]{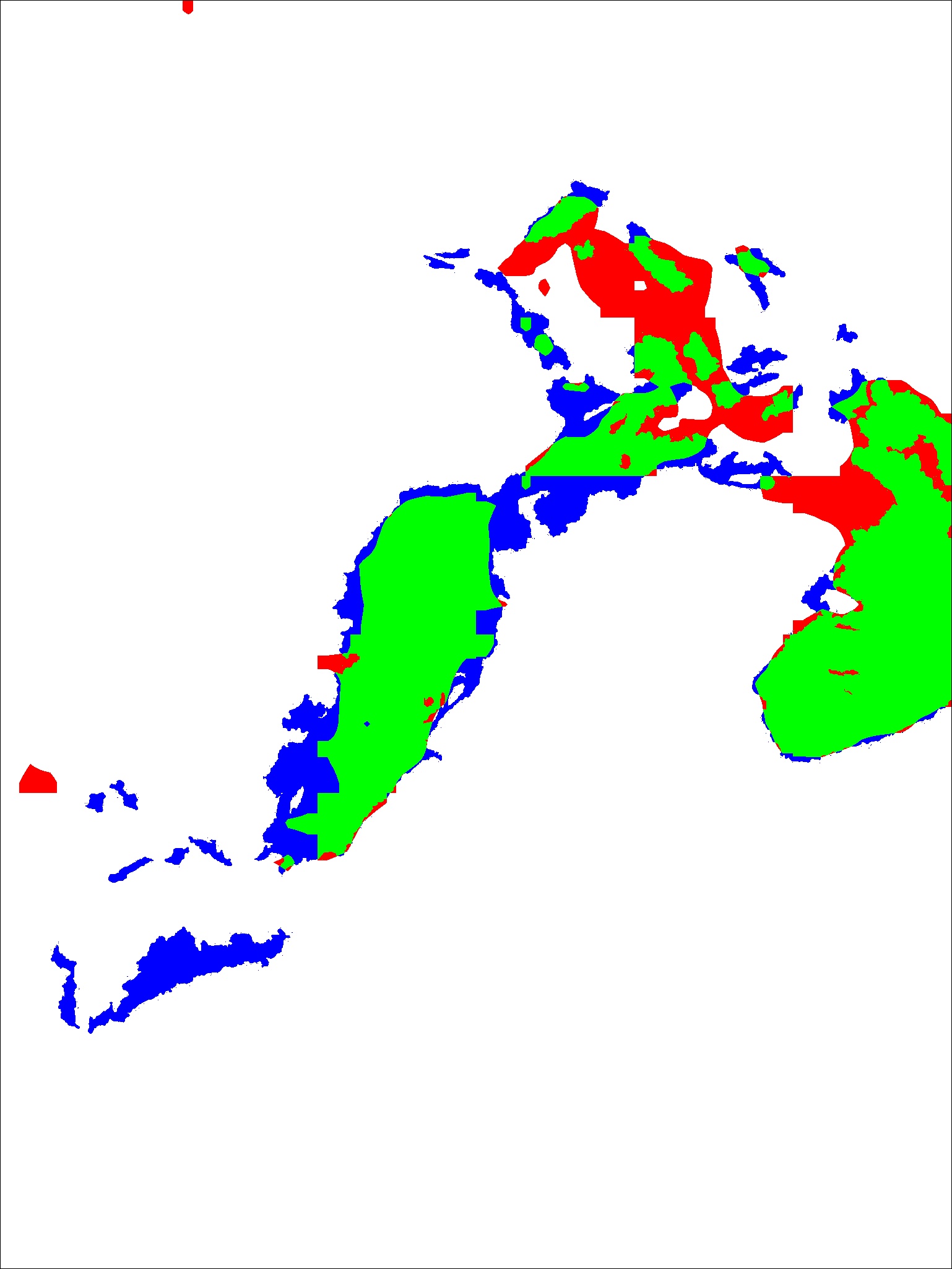}%
              \label{fig1:A2}%
           } 
\hspace{1pt}
\subfloat[BIT]{%
              \includegraphics[width=.193\linewidth]{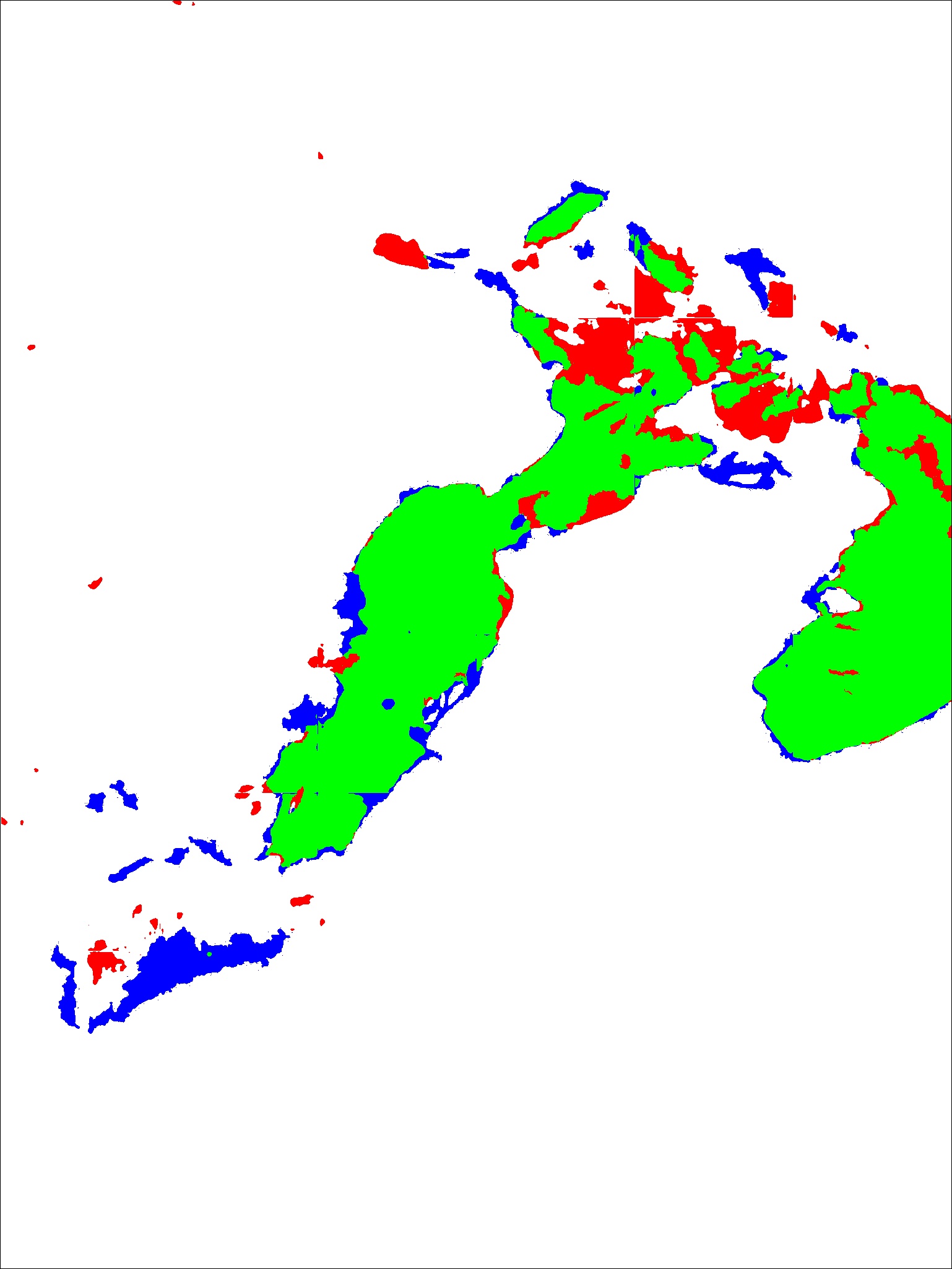}%
              \label{fig1:BIT}%
           } 
\hspace{1pt}
\subfloat[ChangeFormer]{%
              \includegraphics[width=.193\linewidth]{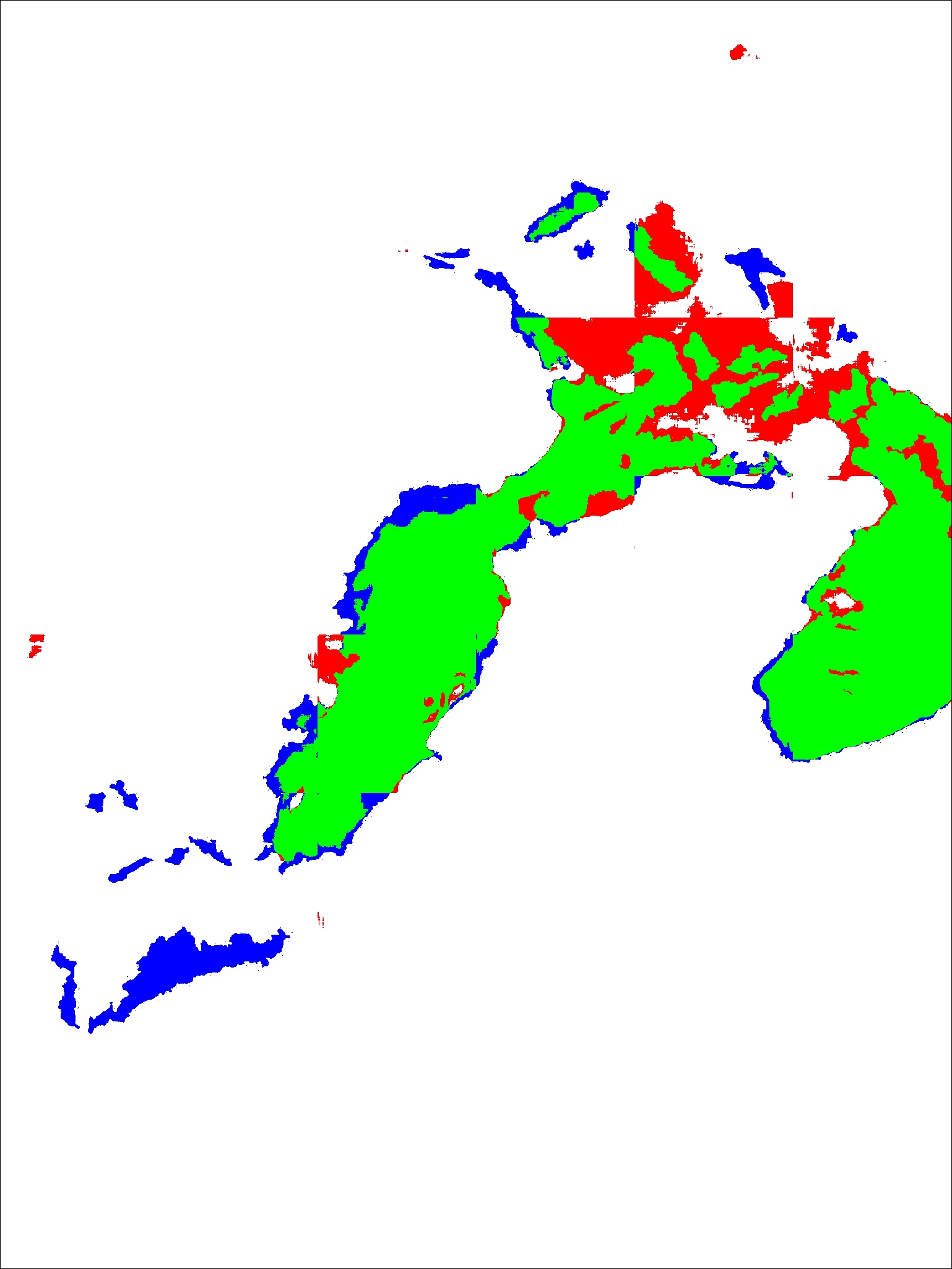}%
              \label{fig1:CF}%
           } 
\hspace{1pt}
\subfloat[DMINet]{%
              \includegraphics[width=.193\linewidth]{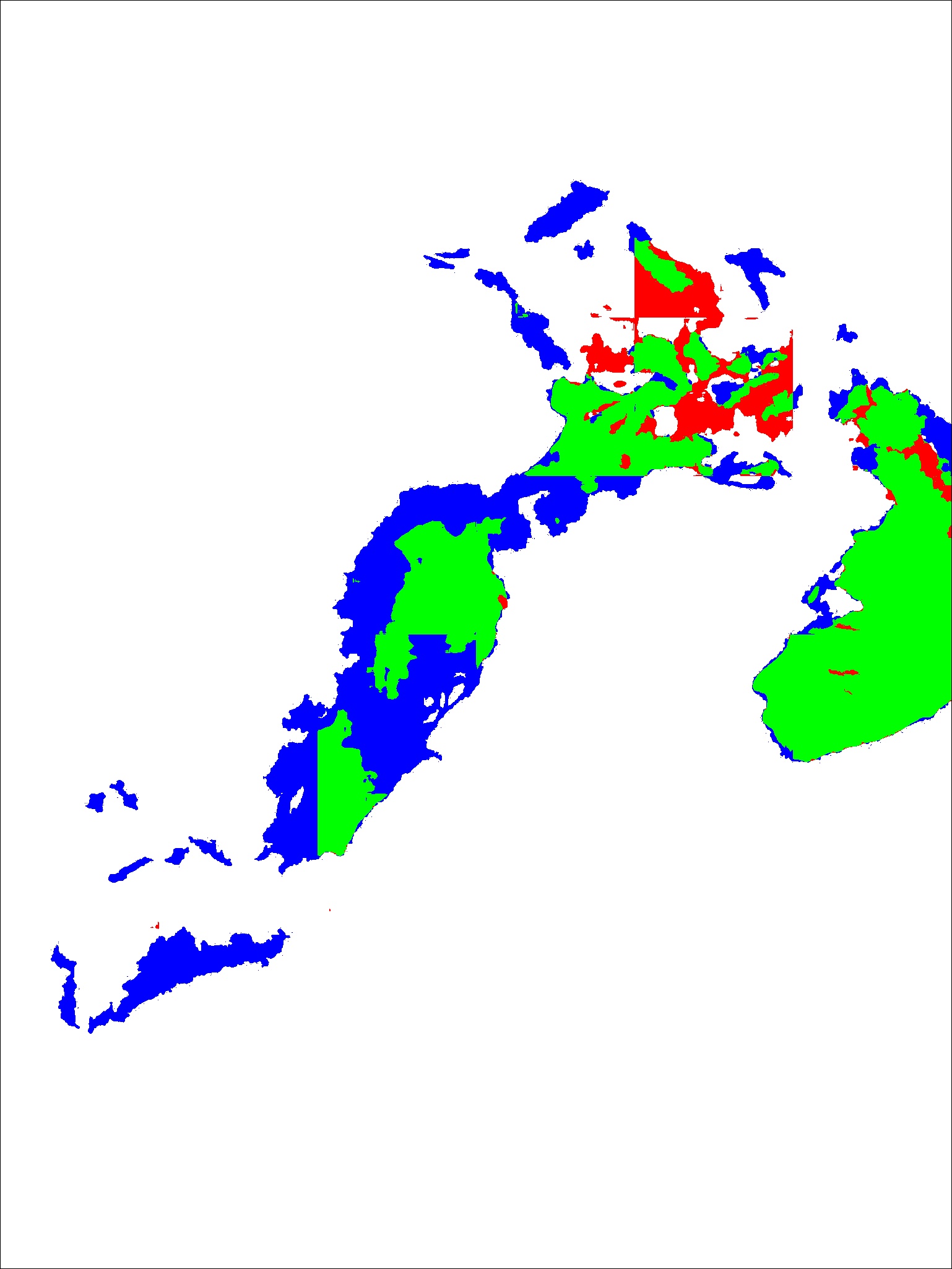}
              \label{fig1:DMI}%
           } 
\hspace{1pt}
\subfloat[FC-EF]{%
              \includegraphics[width=.193\linewidth]{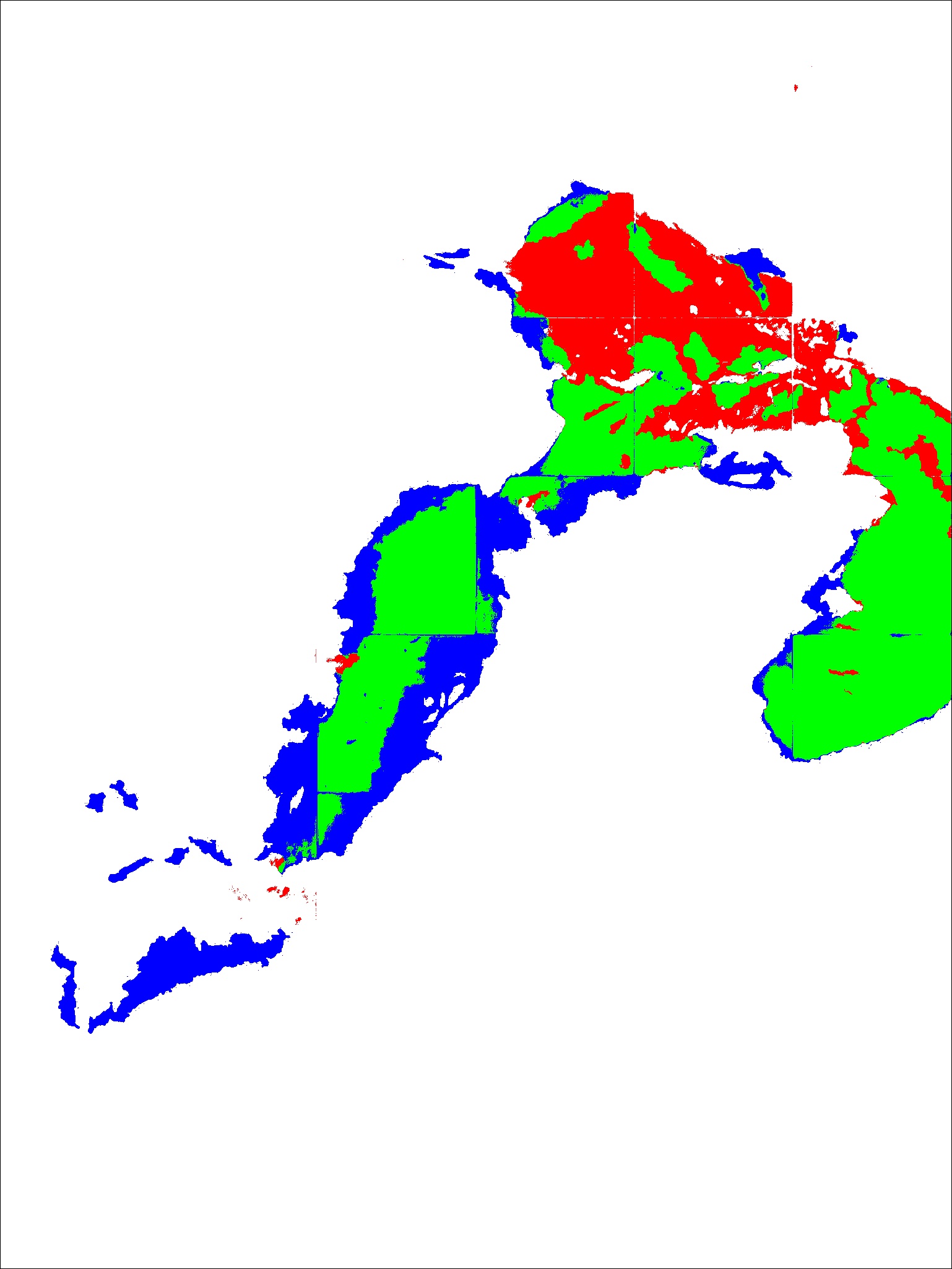}%
              \label{fig1:FCEF}%
           } 
\hspace{1pt}
\subfloat[FCNPP]{%
              \includegraphics[width=.193\linewidth]{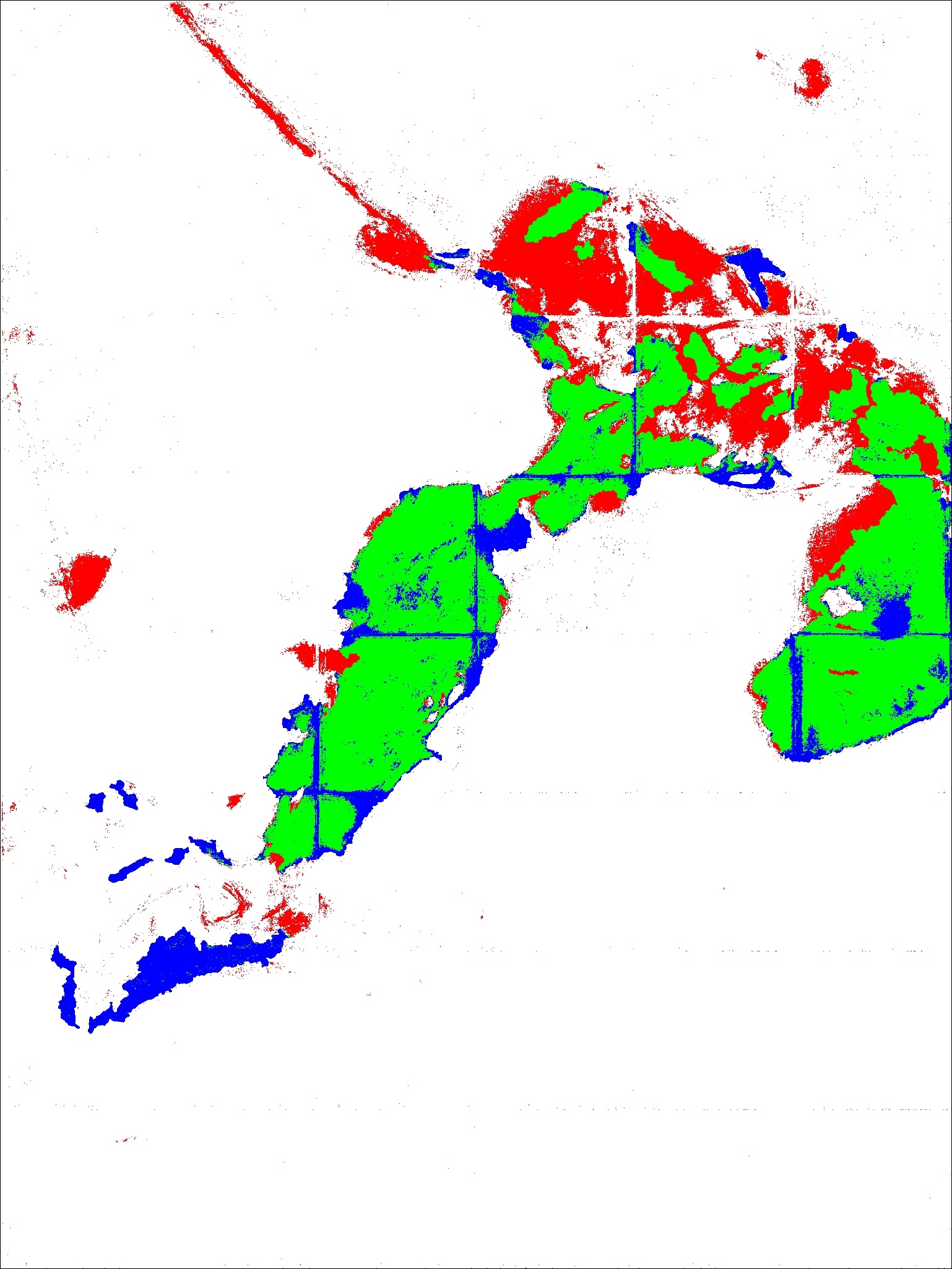}%
              \label{fig1:FCNPP}%
           } 
\hspace{1pt}
\subfloat[ICIFNet]{%
              \includegraphics[width=.193\linewidth]{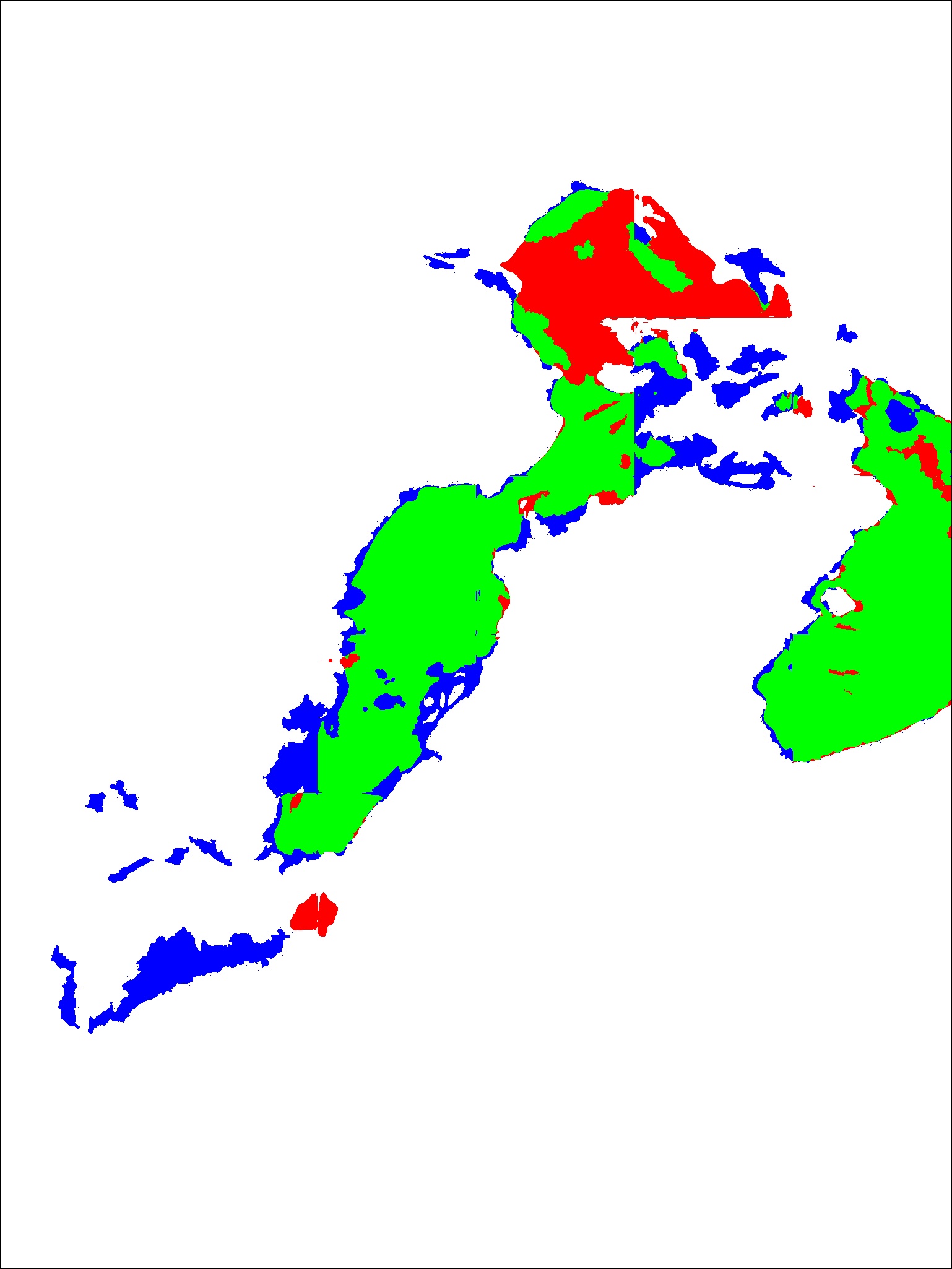}%
              \label{fig1:ICIF}%
           } 
\hspace{1pt}
\subfloat[RDPNet]{%
              \includegraphics[width=.193\linewidth]{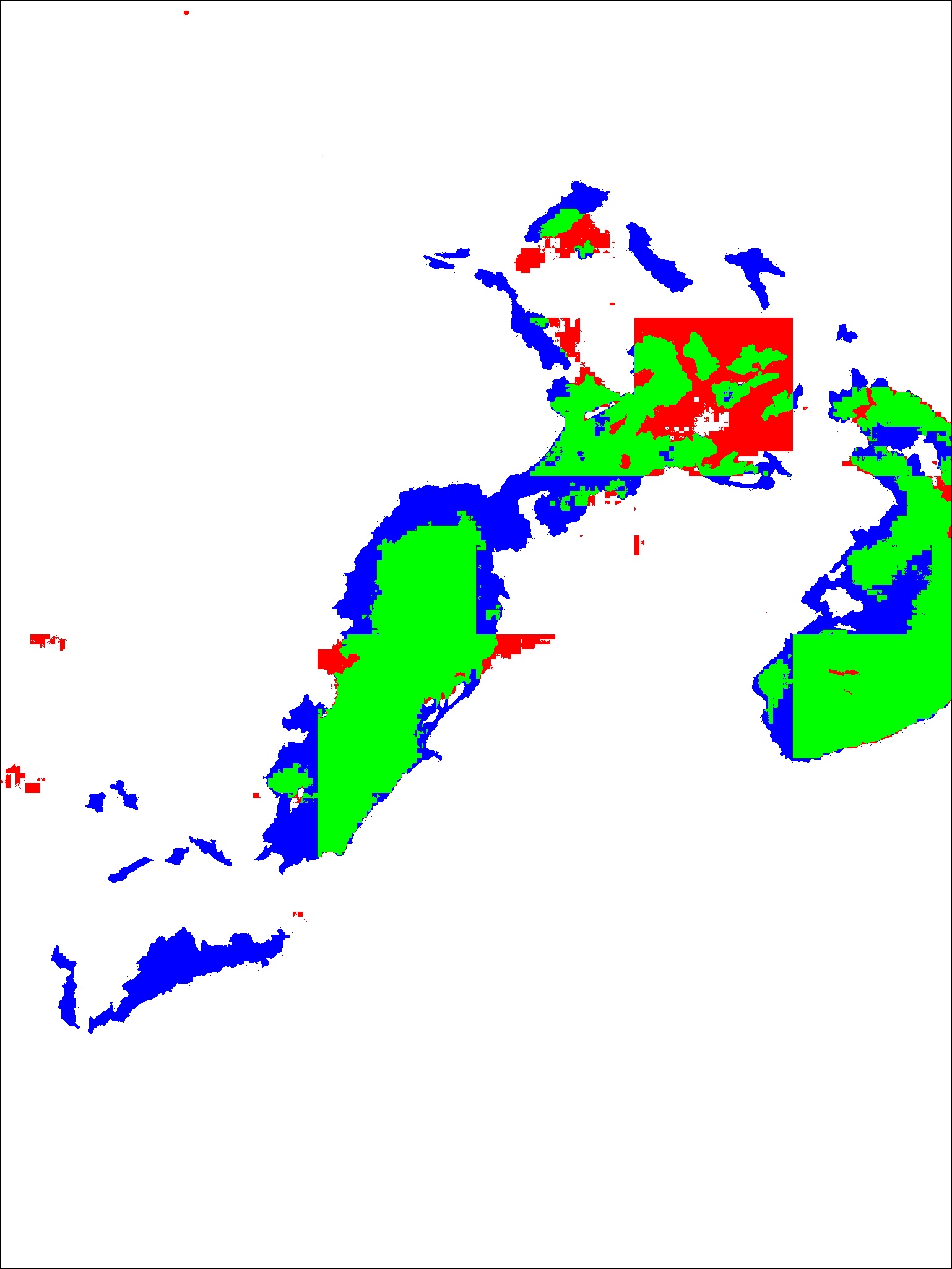}%
              \label{fig1:RDP}%
           } 
\hspace{1pt}
\subfloat[ResUnet]{%
              \includegraphics[width=.193\linewidth]{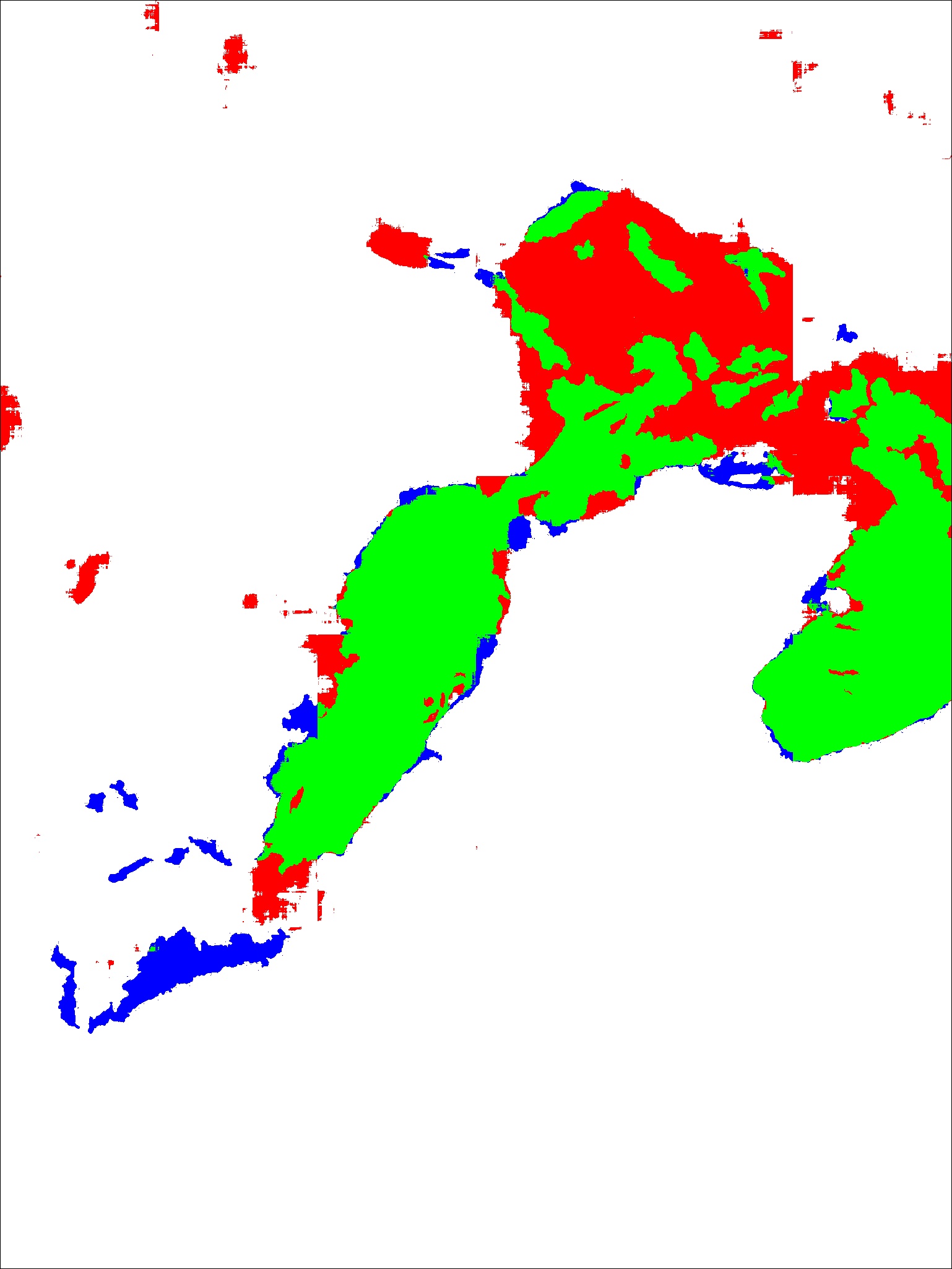}%
              \label{fig1:ResU}%
           } 
\hspace{1pt}
\subfloat[SiamUnet-Conc]{%
              \includegraphics[width=.193\linewidth]{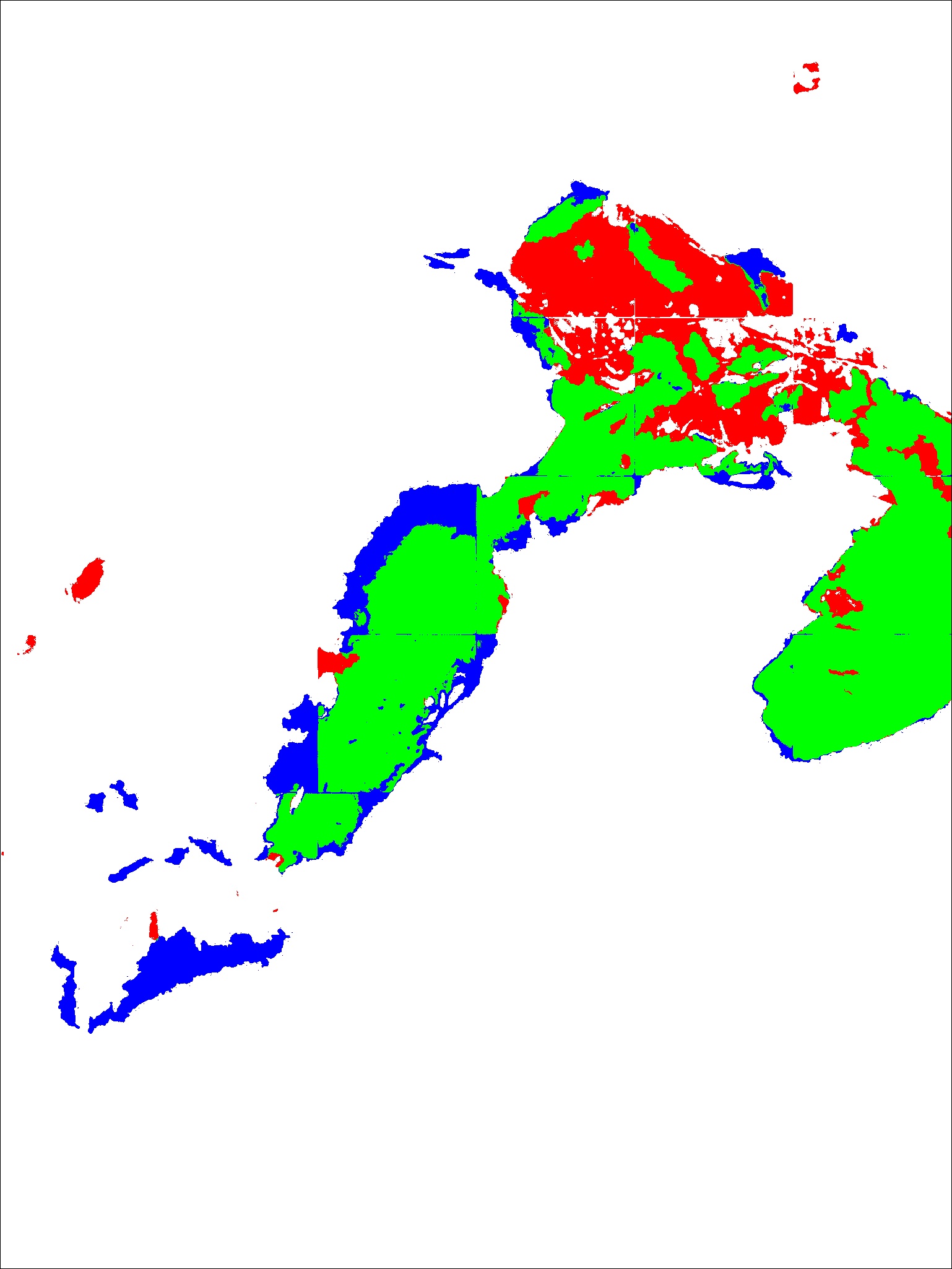}%
              \label{fig1:SUC}%
           } 
\hspace{1pt}
\subfloat[SiamUnet-Diff]{%
              \includegraphics[width=.193\linewidth]{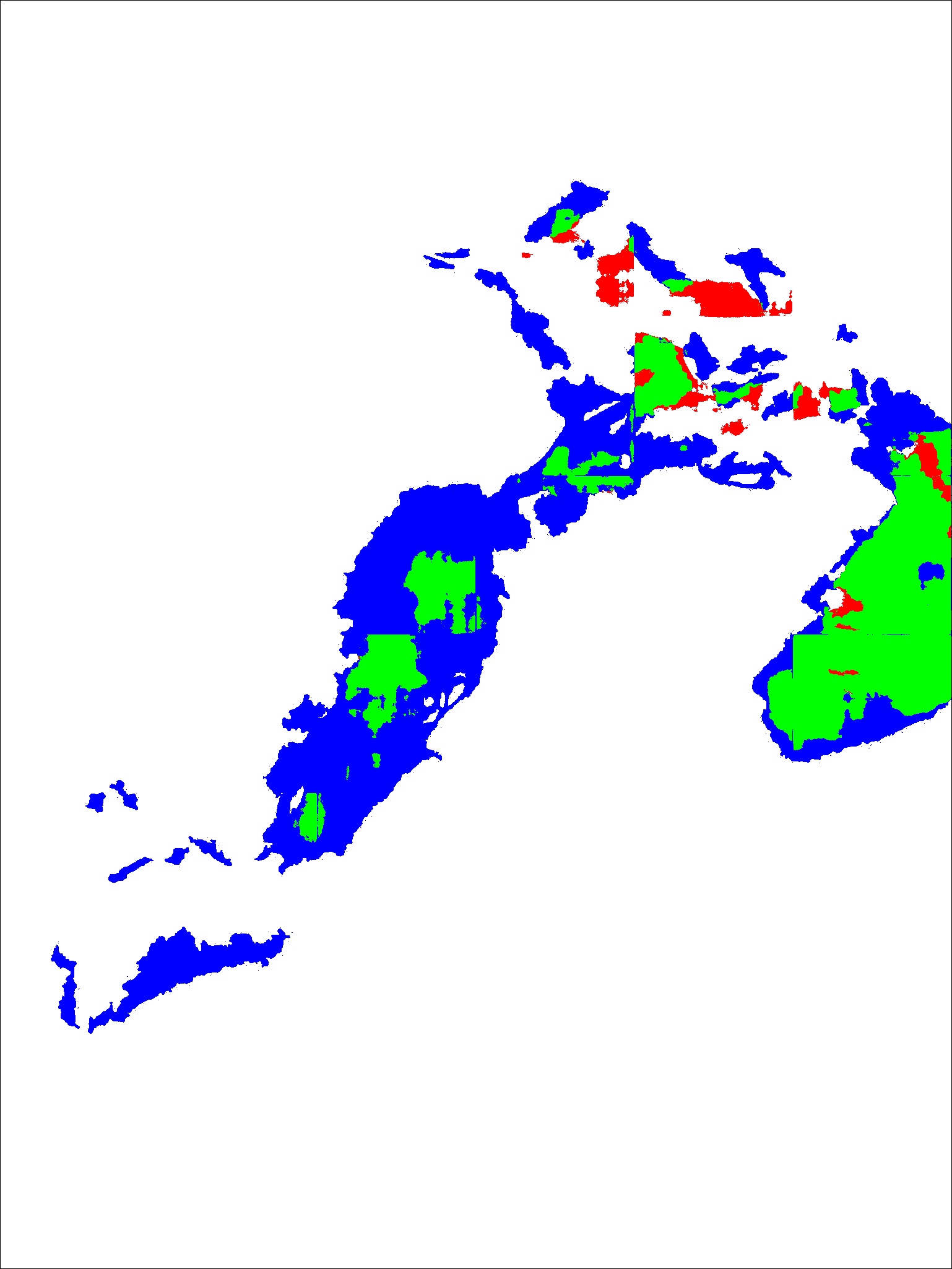}%
              \label{fig1:SUD}%
           } 
\hspace{1pt}
\subfloat[SNUNet]{%
              \includegraphics[width=.193\linewidth]{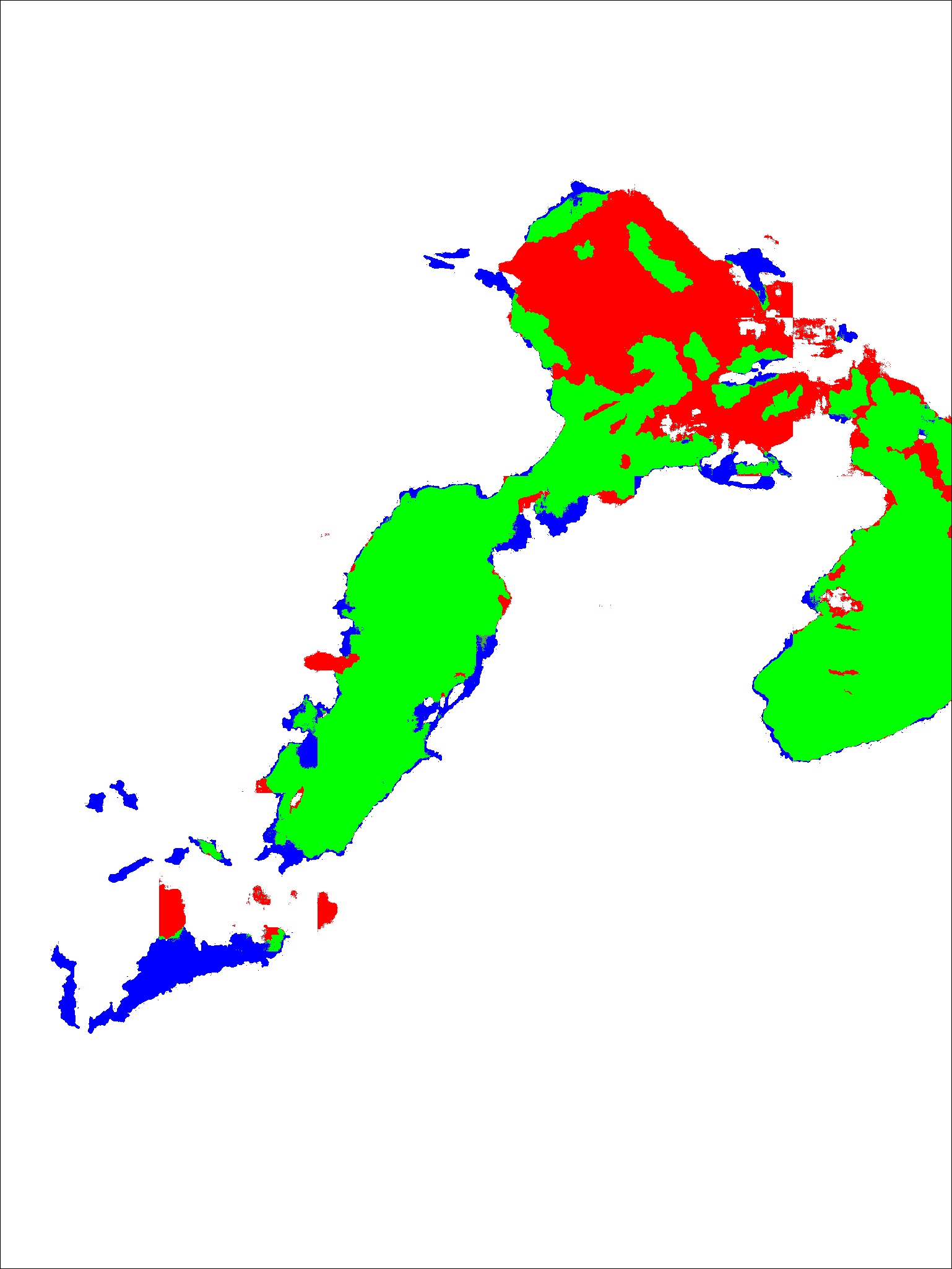}%
              \label{fig1:SNU}%
           } 
\hspace{1pt}
\subfloat[MineNetCD (Ours)]{%
              \includegraphics[width=.193\linewidth]{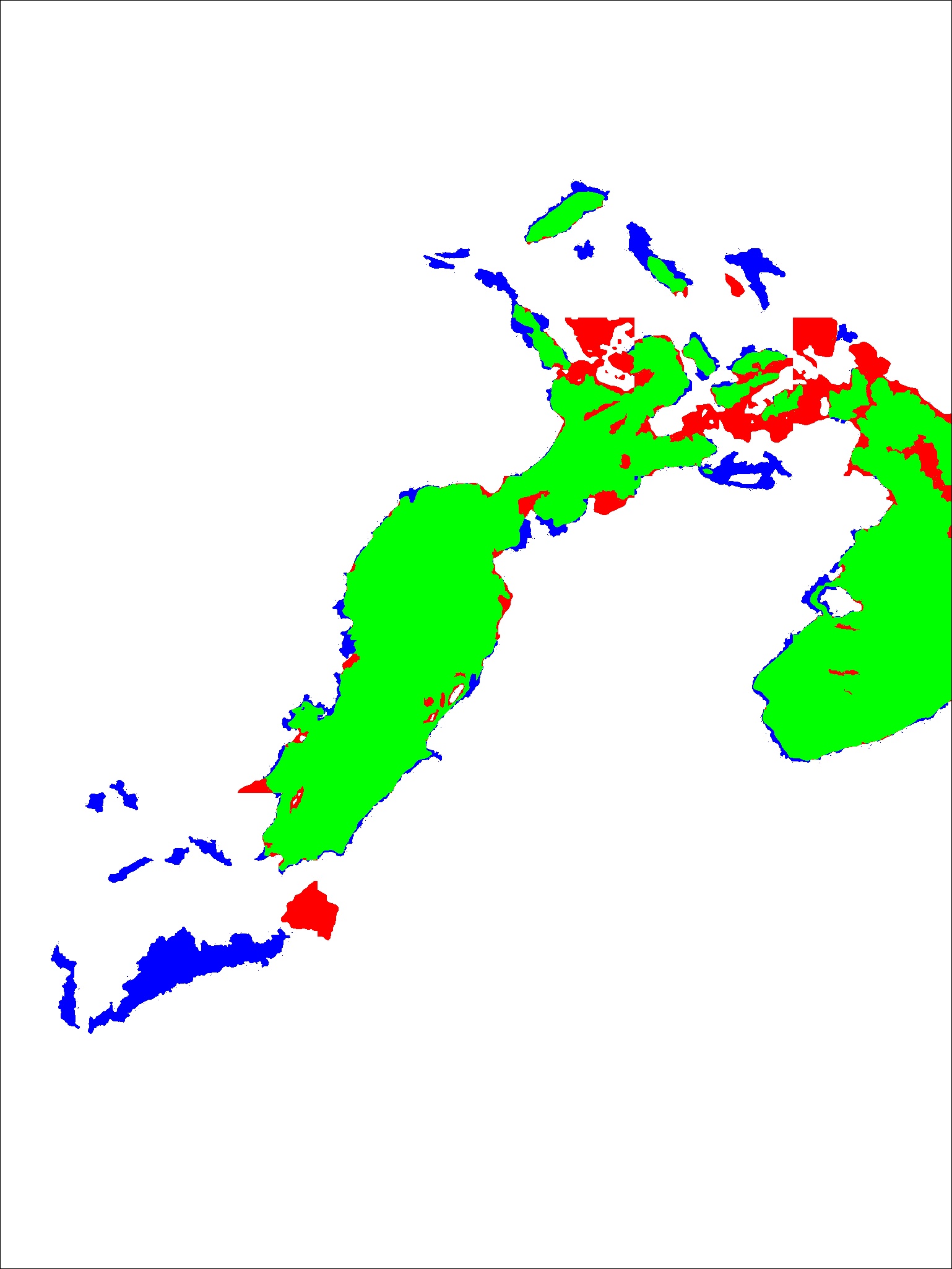}%
              \label{fig1:CFFT}%
           } 
\hspace{1pt}
\caption{Visual comparisons of the proposed method and the state-of-the-art approaches on the MineNetCD dataset. The rendered colors represent true positives (green), true negatives (white), false positives (red), and false negatives (blue).}
\label{fig:quali1}
\end{figure*}

\begin{figure*}
\hspace{1pt}
\subfloat[Pre-change]{%
              \includegraphics[width=.193\linewidth]{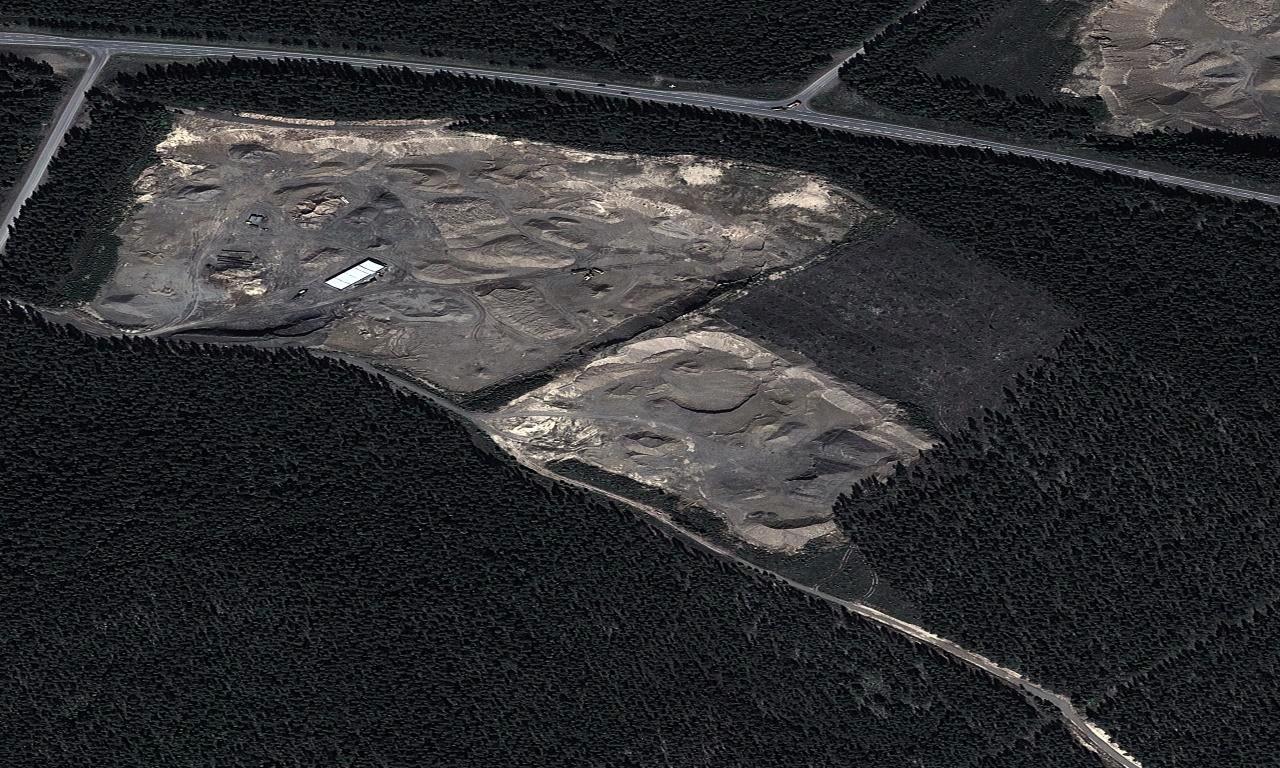}%
              \label{fig2:pre}%
           } 
\hspace{1pt}
\subfloat[Post-change]{%
              \includegraphics[width=.193\linewidth]{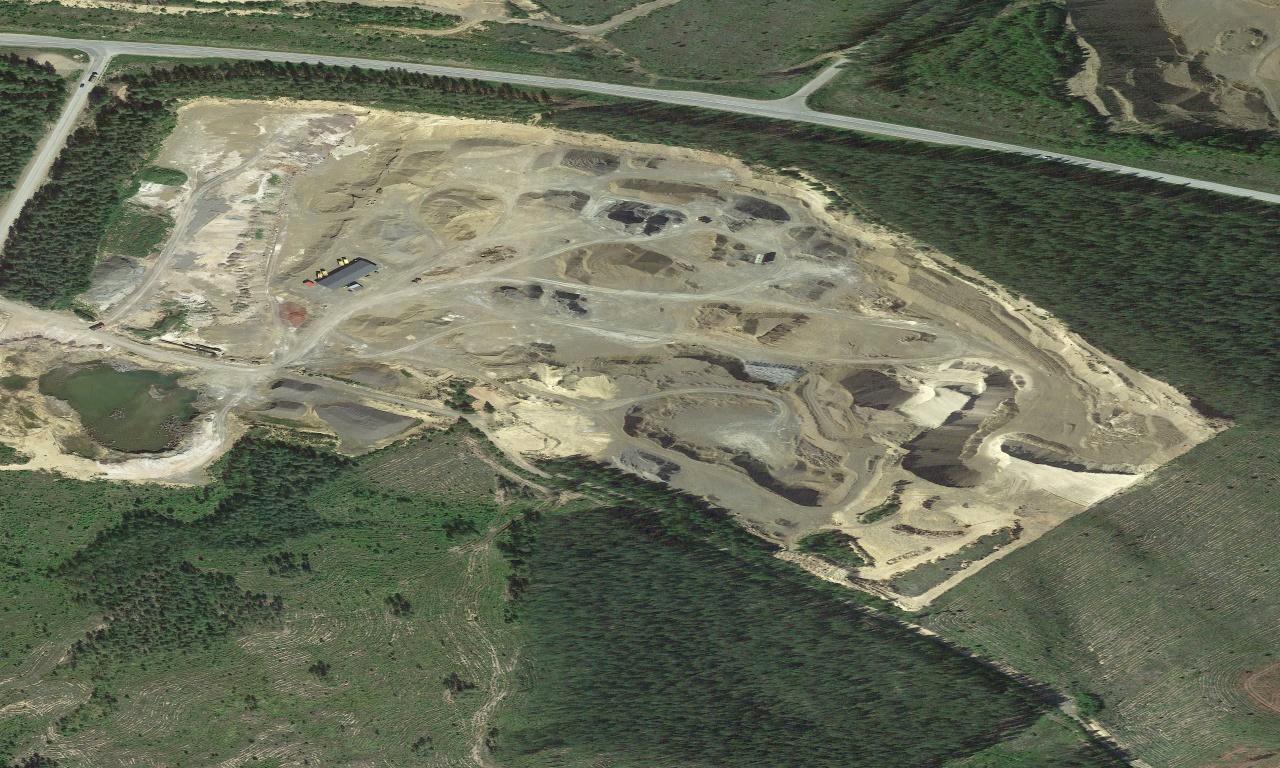}%
              \label{fig2:post}%
           } 
\hspace{1pt}
\subfloat[A2Net]{%
              \includegraphics[width=.193\linewidth]{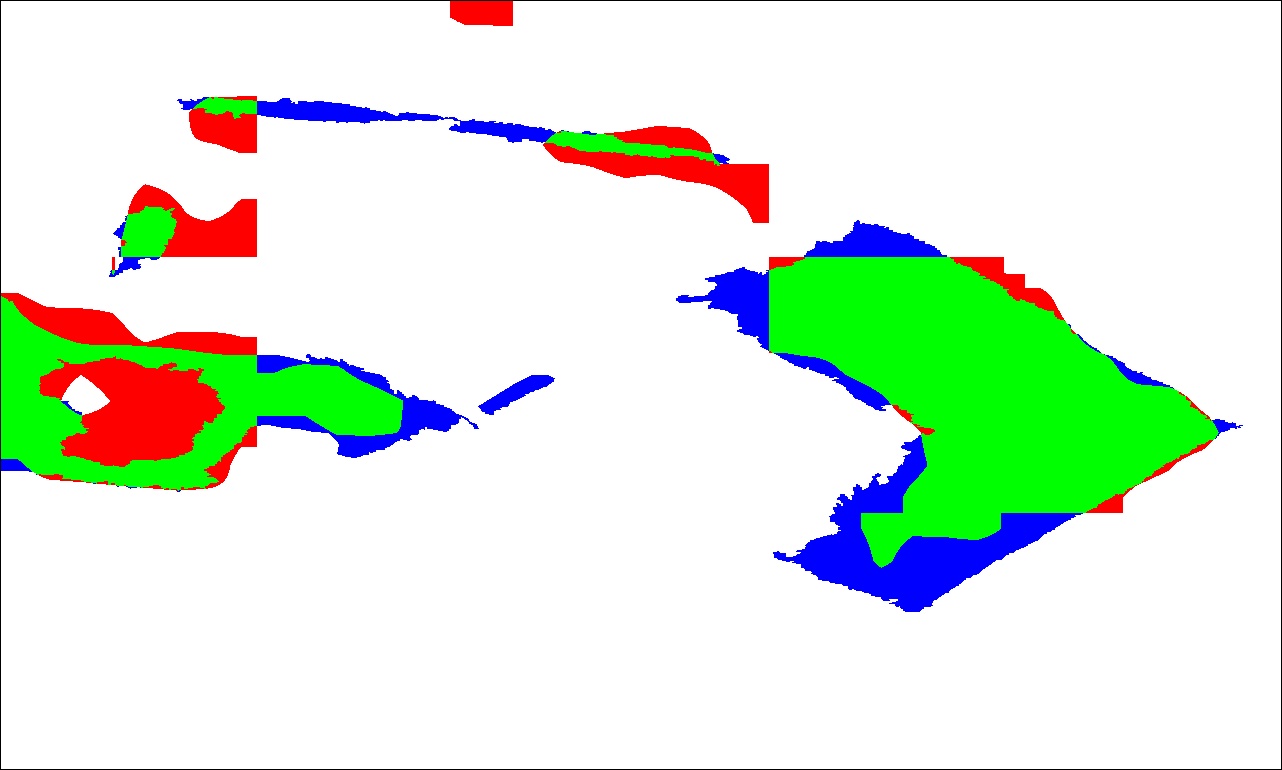}%
              \label{fig2:A2}%
           } 
\hspace{1pt}
\subfloat[BIT]{%
              \includegraphics[width=.193\linewidth]{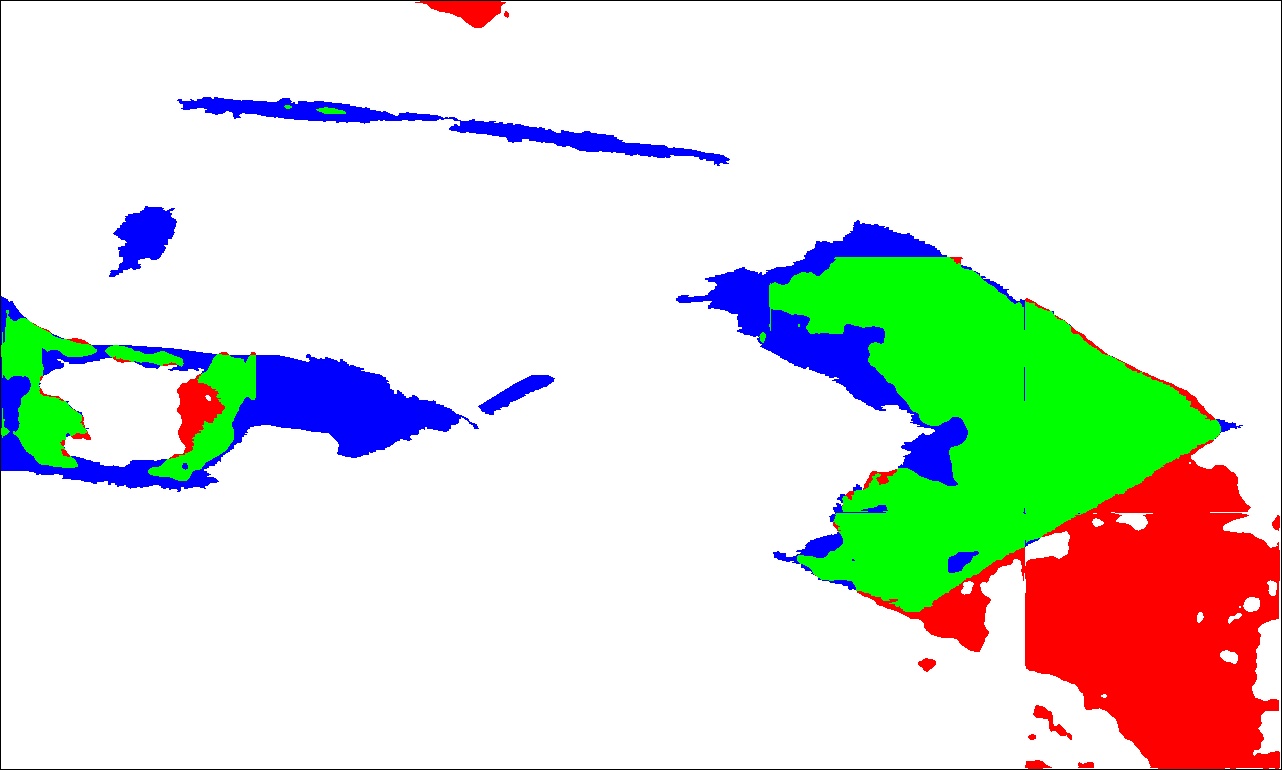}%
              \label{fig2:BIT}%
           } 
\hspace{1pt}
\subfloat[ChangeFormer]{%
              \includegraphics[width=.193\linewidth]{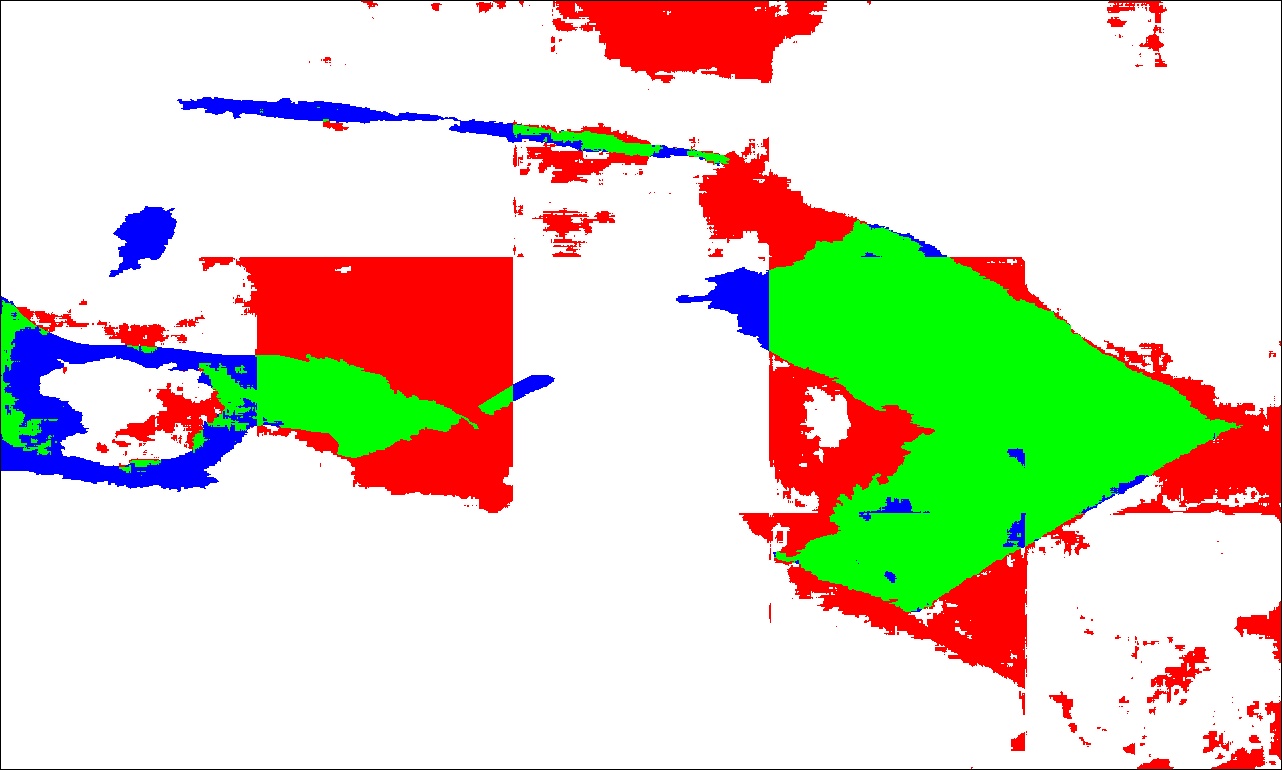}%
              \label{fig2:CF}%
           } 
\hspace{1pt}
\subfloat[DMINet]{%
              \includegraphics[width=.193\linewidth]{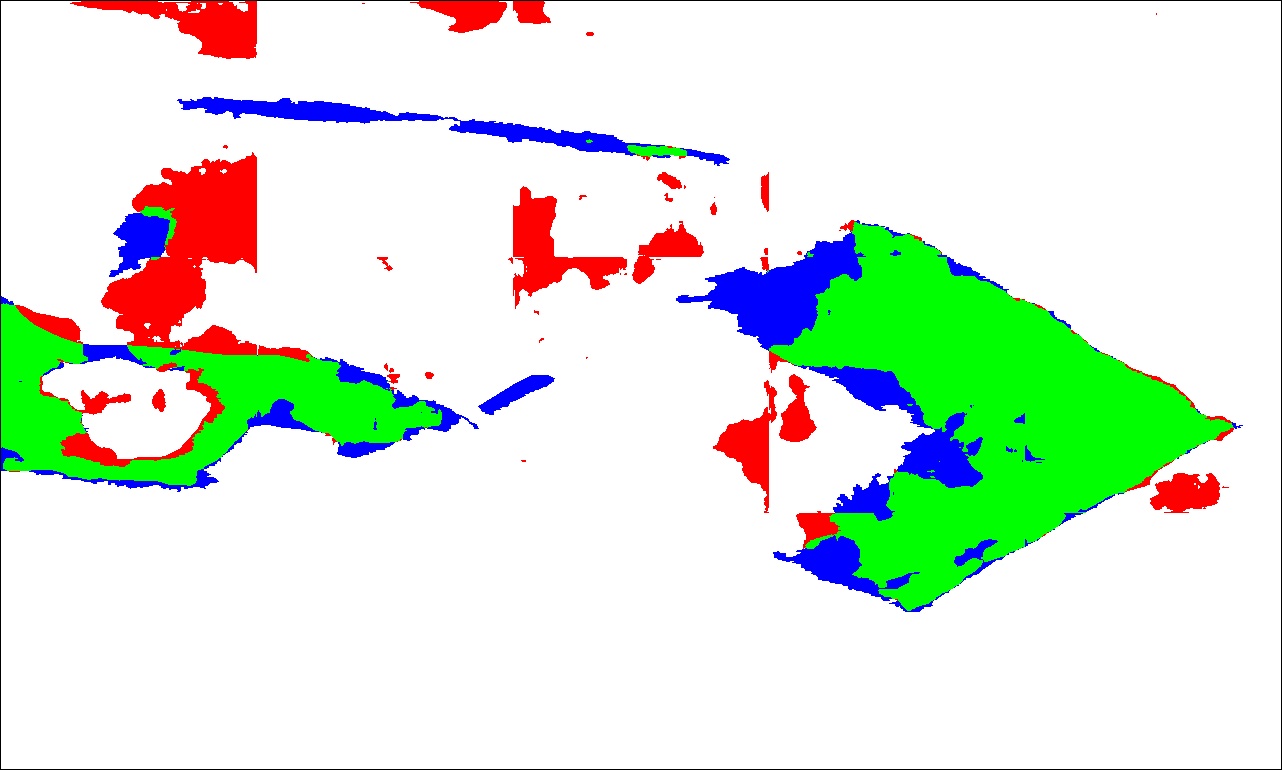}
              \label{fig2:DMI}%
           } 
\hspace{1pt}
\subfloat[FC-EF]{%
              \includegraphics[width=.193\linewidth]{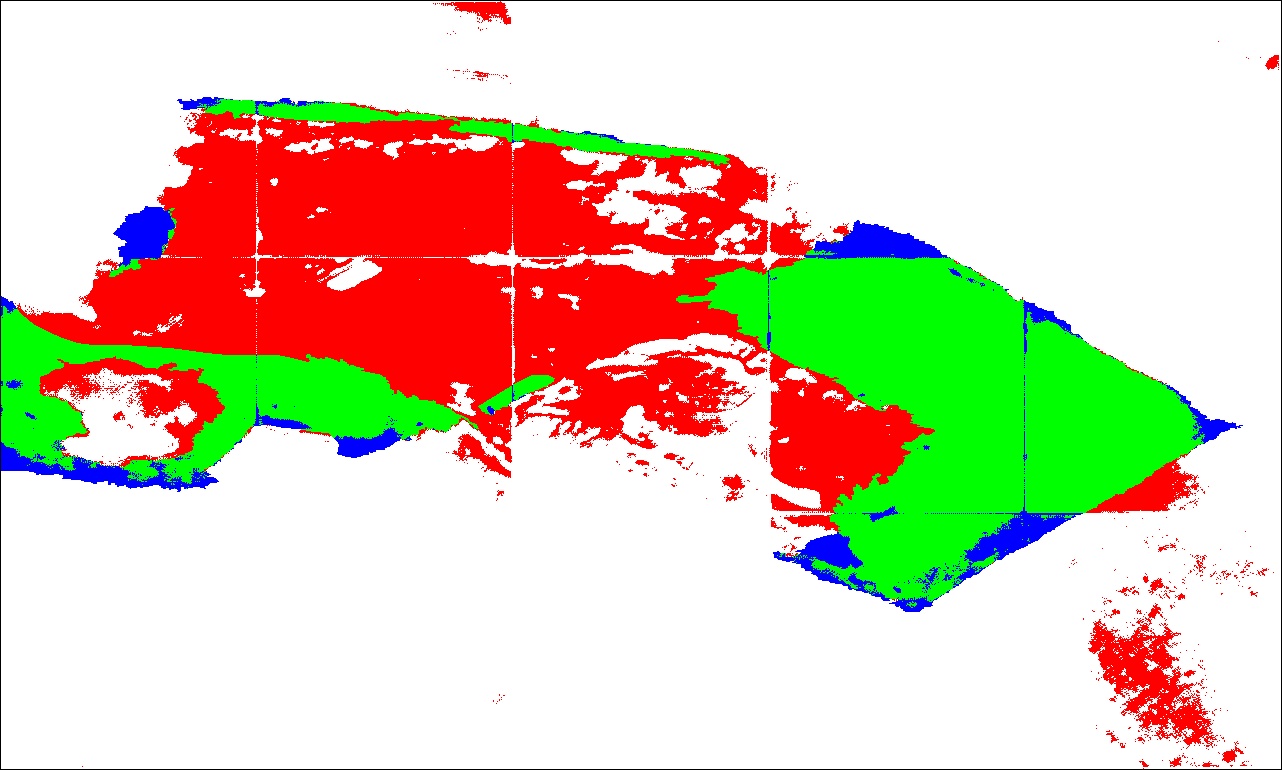}%
              \label{fig2:FCEF}%
           } 
\hspace{1pt}
\subfloat[FCNPP]{%
              \includegraphics[width=.193\linewidth]{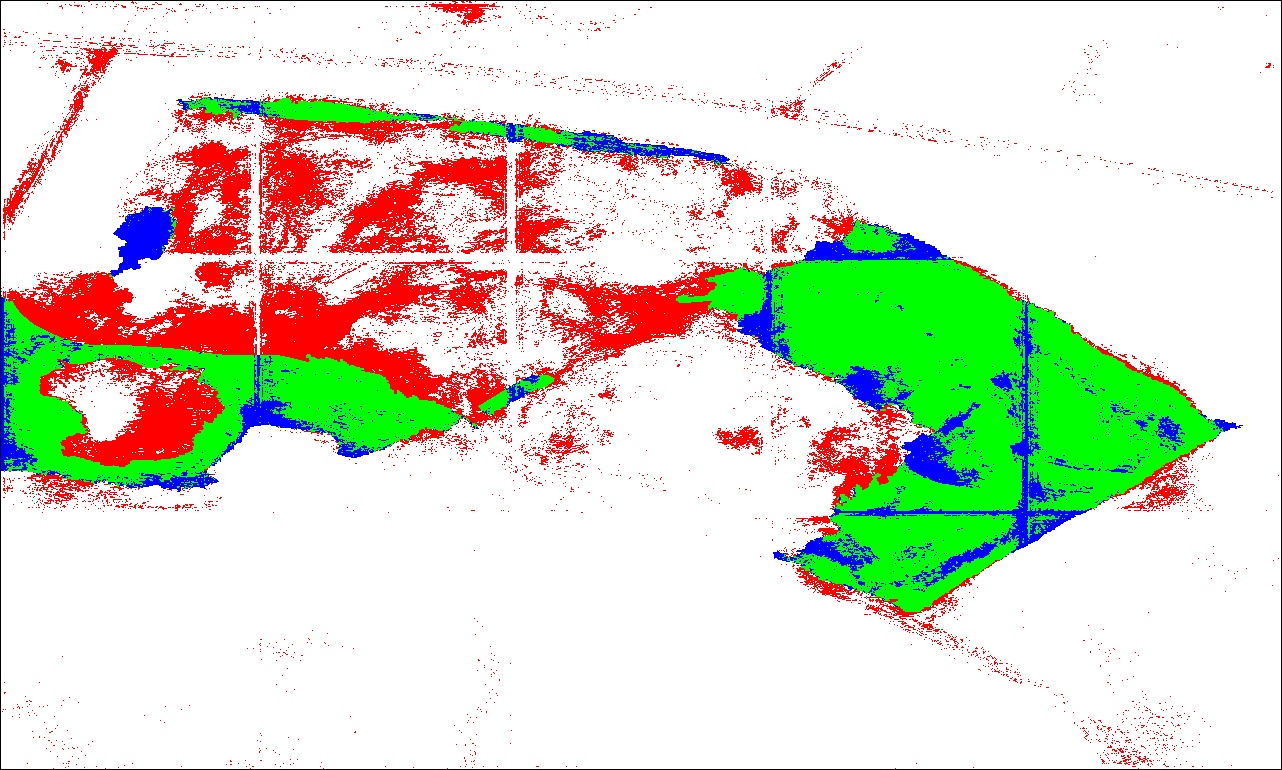}%
              \label{fig2:FCNPP}%
           } 
\hspace{1pt}
\subfloat[ICIFNet]{%
              \includegraphics[width=.193\linewidth]{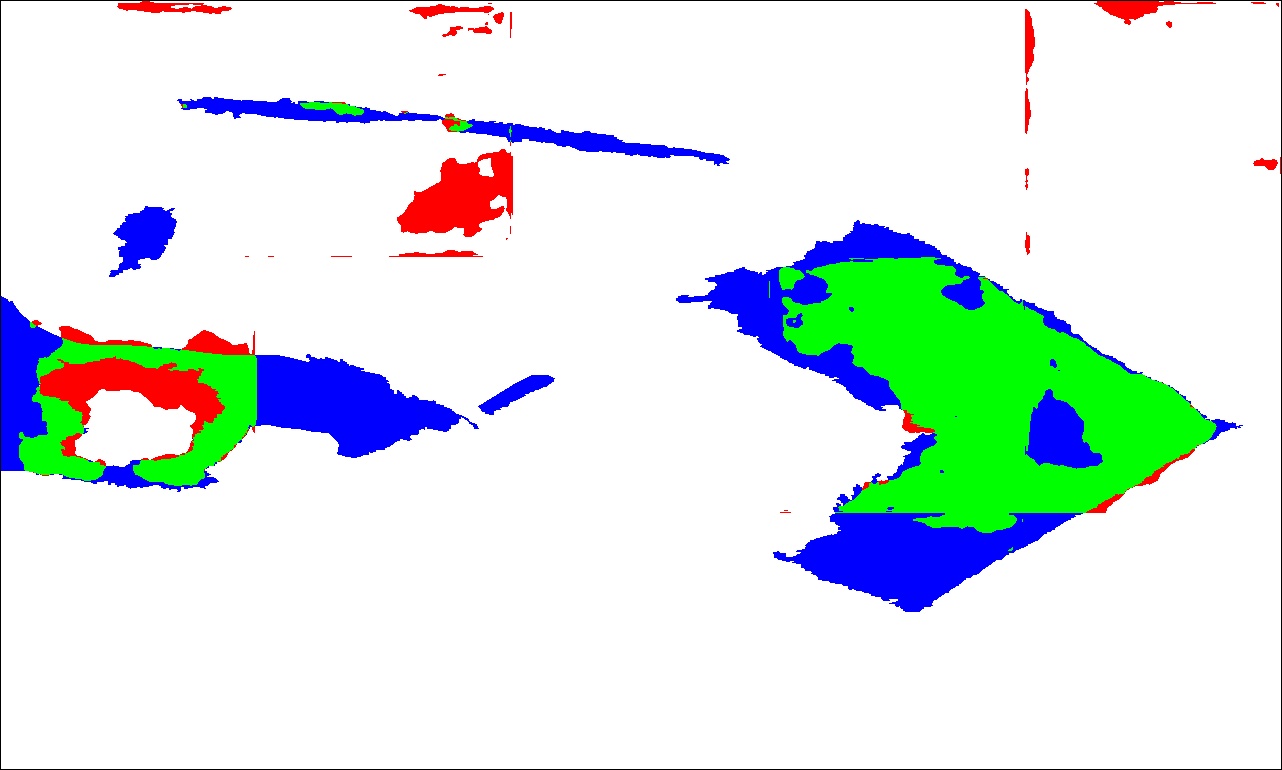}%
              \label{fig2:ICIF}%
           } 
\hspace{1pt}
\subfloat[RDPNet]{%
              \includegraphics[width=.193\linewidth]{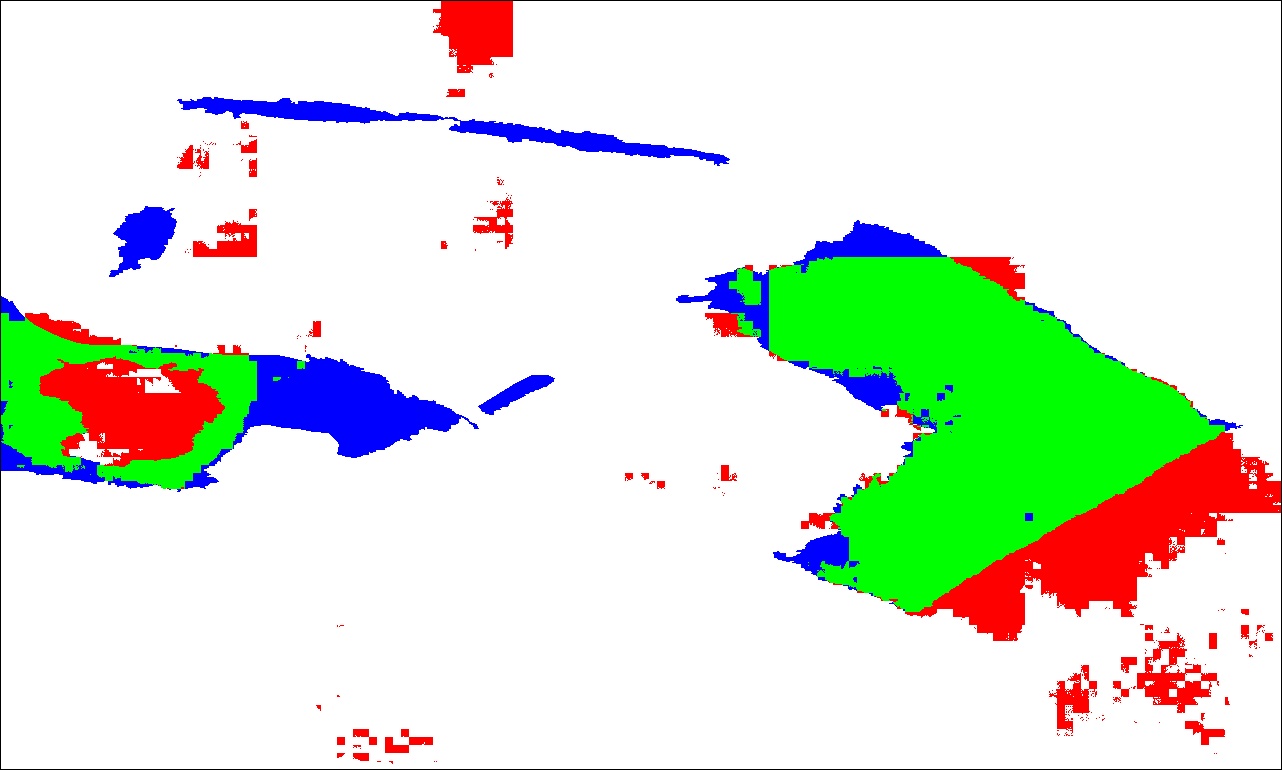}%
              \label{fig2:RDP}%
           } 
\hspace{1pt}
\subfloat[ResUnet]{%
              \includegraphics[width=.193\linewidth]{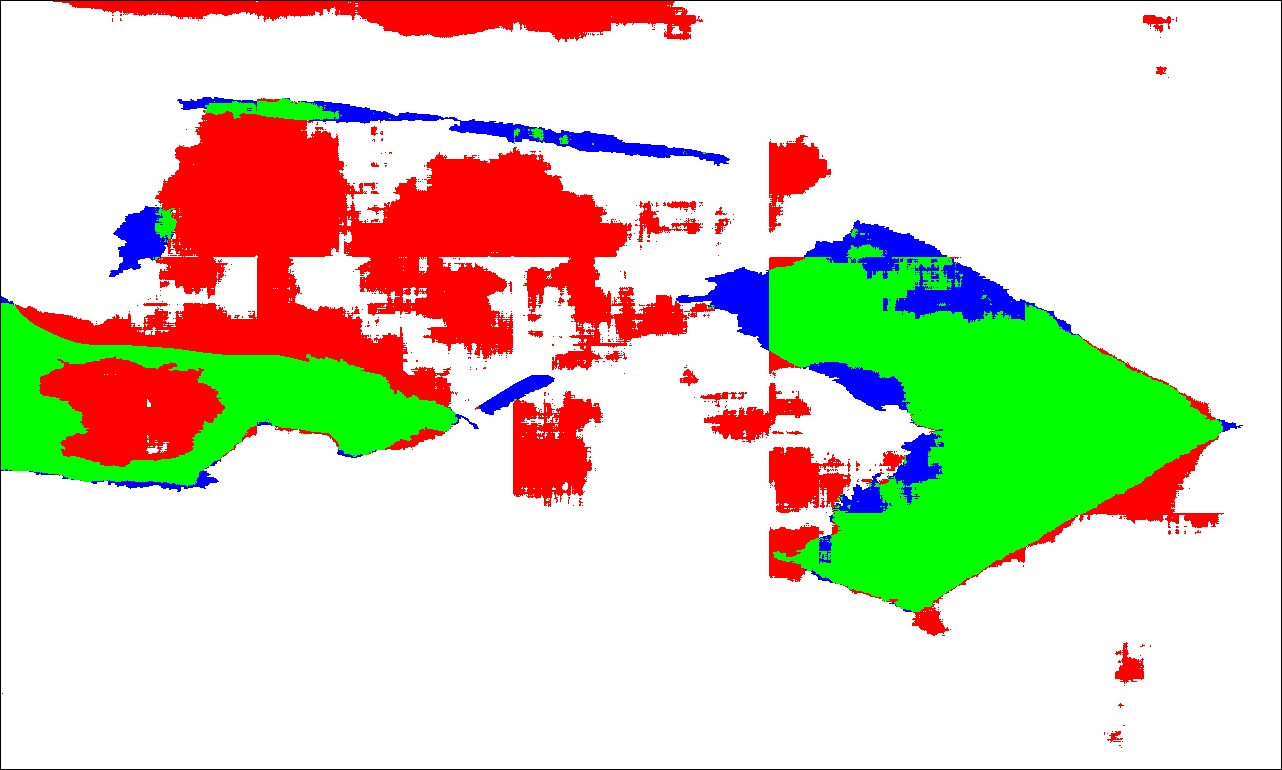}%
              \label{fig2:ResU}%
           } 
\hspace{1pt}
\subfloat[SiamUnet-Conc]{%
              \includegraphics[width=.193\linewidth]{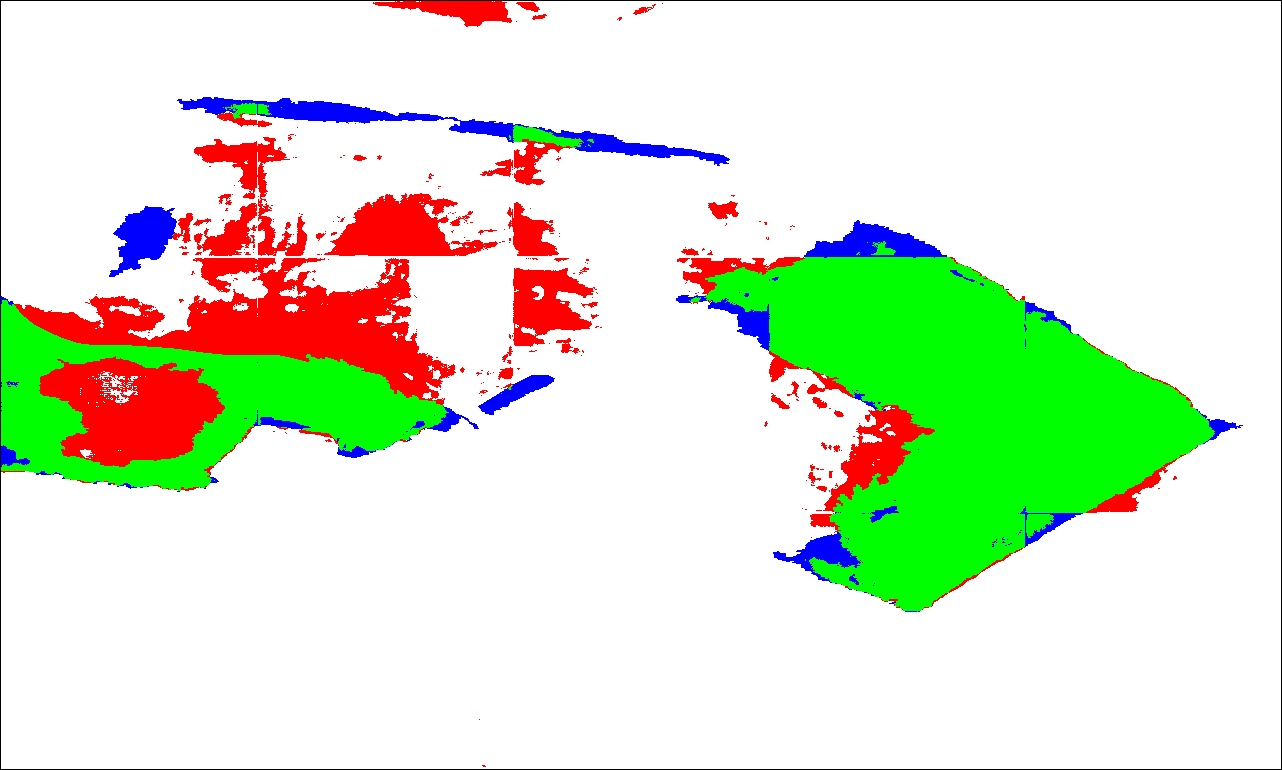}%
              \label{fig2:SUC}%
           } 
\hspace{1pt}
\subfloat[SiamUnet-Diff]{%
              \includegraphics[width=.193\linewidth]{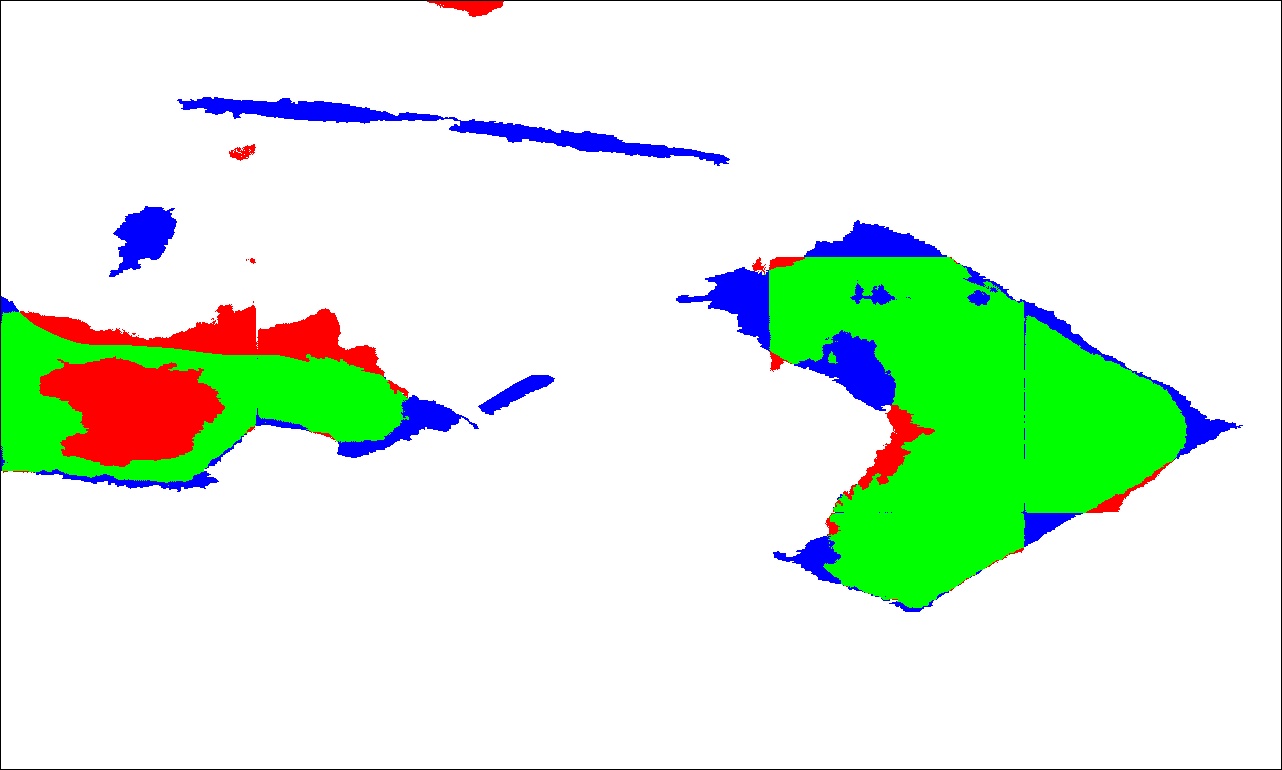}%
              \label{fig2:SUD}%
           } 
\hspace{1pt}
\subfloat[SNUNet]{%
              \includegraphics[width=.193\linewidth]{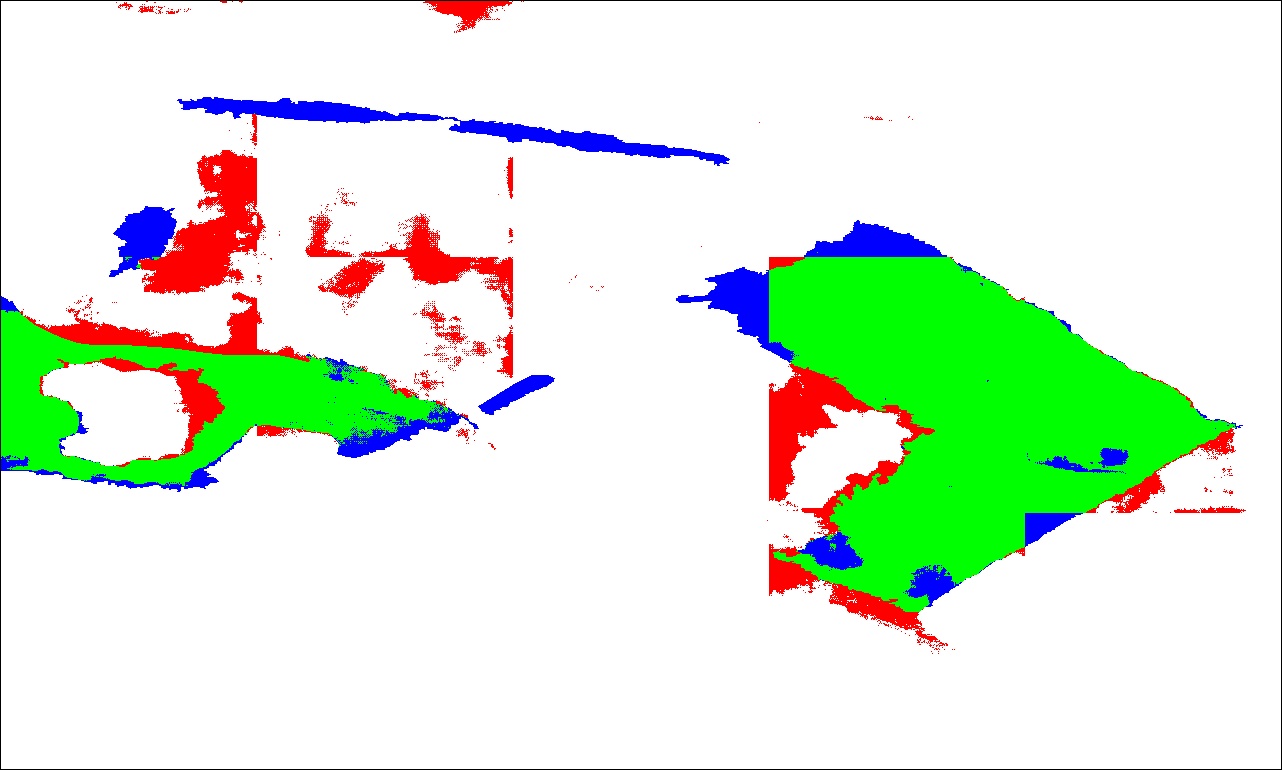}%
              \label{fig2:SNU}%
           } 
\hspace{1pt}
\subfloat[MineNetCD (Ours)]{%
              \includegraphics[width=.193\linewidth]{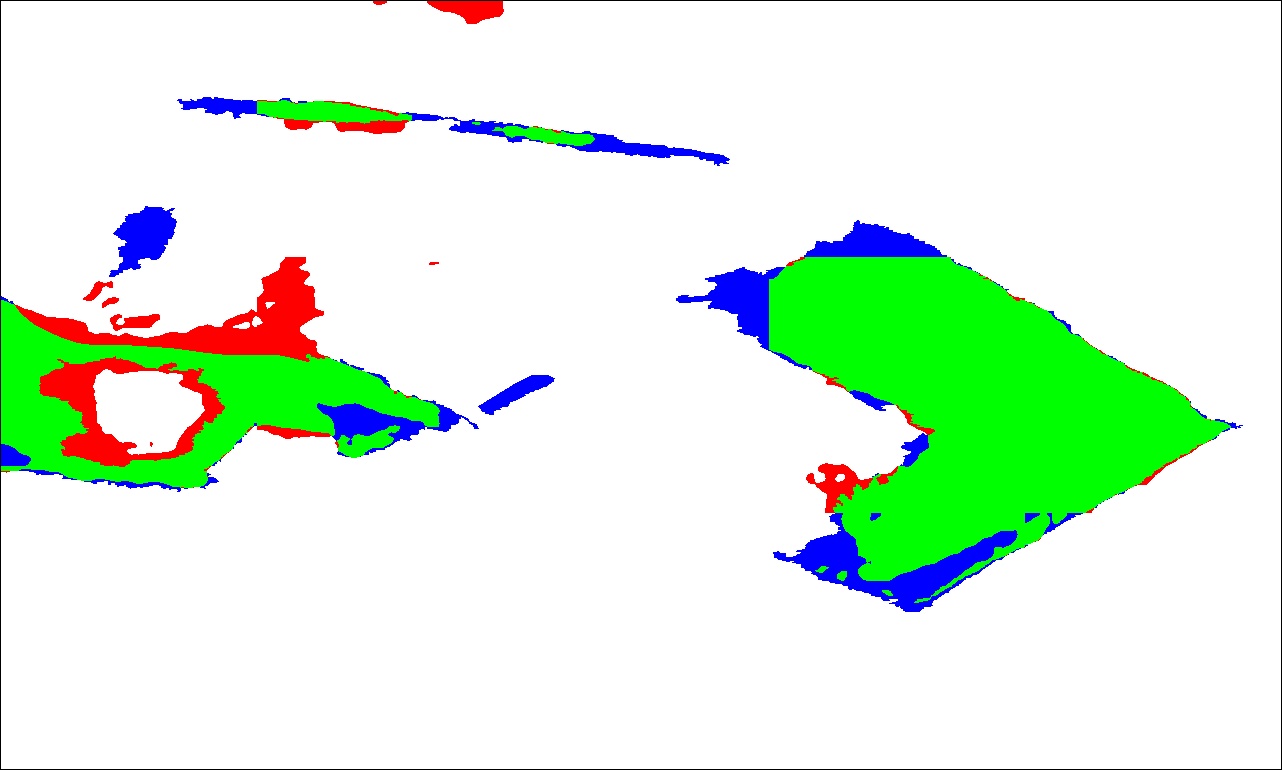}%
              \label{fig2:CFFT}%
           } 
\hspace{1pt}
\caption{Visual comparisons of the proposed method and the state-of-the-art approaches on the MineNetCD dataset. The rendered colors represent true positives (green), true negatives (white), false positives (red), and false negatives (blue).}
\label{fig:quali2}
\end{figure*}
\begin{figure*}
		\centering
		\begin{minipage}{0.065\linewidth}
			\vspace{1pt}
			\centerline{\includegraphics[width=\textwidth]{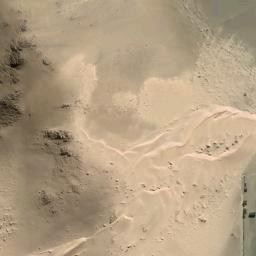}}
			\vspace{1pt}
                \centerline{\includegraphics[width=\textwidth]{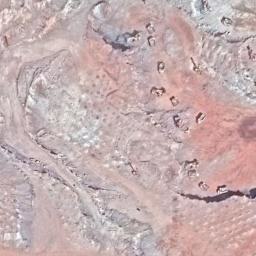}}
                \vspace{1pt}
                \centerline{\includegraphics[width=\textwidth]{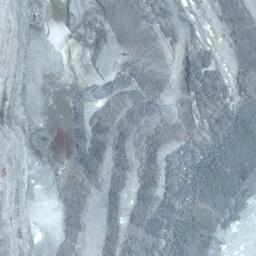}}
                \vspace{1pt}
                \centerline{\includegraphics[width=\textwidth]{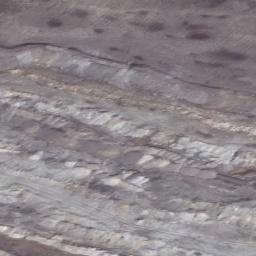}}
                \vspace{1pt}
                \centerline{\includegraphics[width=\textwidth]{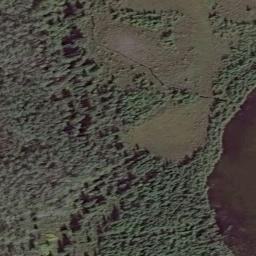}}
                \vspace{1pt}
                \centerline{\includegraphics[width=\textwidth]{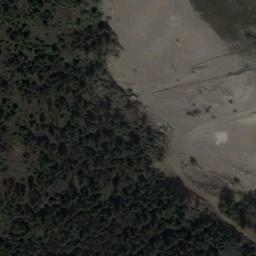}}
                \vspace{1pt}
                \centerline{\includegraphics[width=\textwidth]{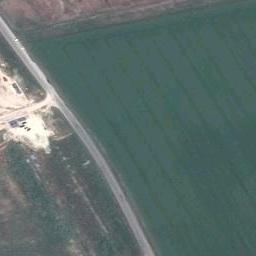}}
                \vspace{1pt}
                \centerline{\includegraphics[width=\textwidth]{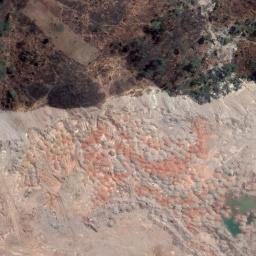}}
                \vspace{1pt}
                \centerline{\includegraphics[width=\textwidth]{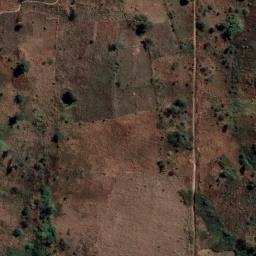}}
                \vspace{1pt}
                \centerline{\includegraphics[width=\textwidth]{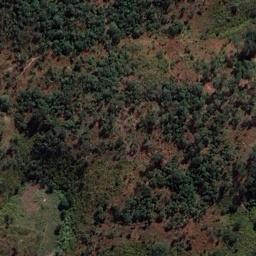}}
			\centerline{(a)}
		\end{minipage}
		\hspace{-6pt}
		\begin{minipage}{0.065\linewidth}
			\vspace{1pt}
			\centerline{\includegraphics[width=\textwidth]{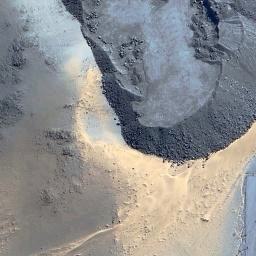}}
			\vspace{1pt}
                \centerline{\includegraphics[width=\textwidth]{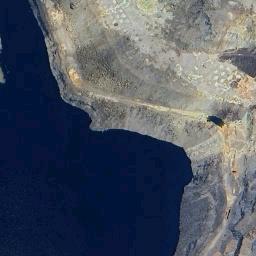}}
                \vspace{1pt}
                \centerline{\includegraphics[width=\textwidth]{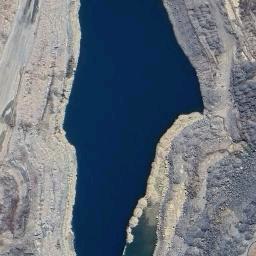}}
                \vspace{1pt}
                \centerline{\includegraphics[width=\textwidth]{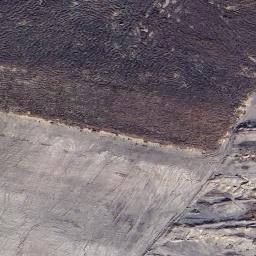}}
                \vspace{1pt}
                \centerline{\includegraphics[width=\textwidth]{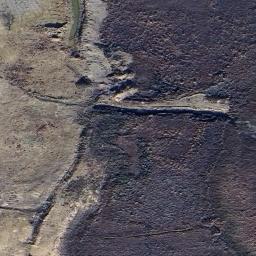}}
                \vspace{1pt}
                \centerline{\includegraphics[width=\textwidth]{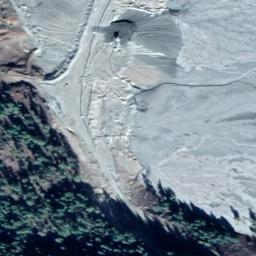}} 
                \vspace{1pt}
                \centerline{\includegraphics[width=\textwidth]{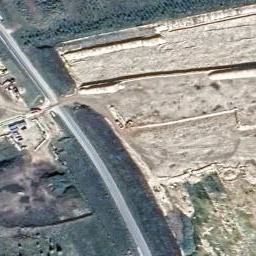}}
                \vspace{1pt}
                \centerline{\includegraphics[width=\textwidth]{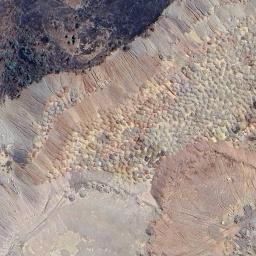}}
                \vspace{1pt}
                \centerline{\includegraphics[width=\textwidth]{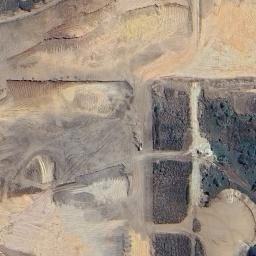}}
                \vspace{1pt}
                \centerline{\includegraphics[width=\textwidth]{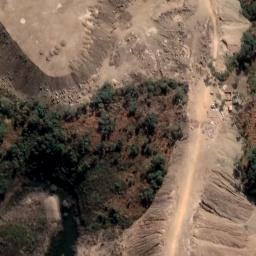}}
			\centerline{(b)}
		\end{minipage}
		\hspace{-6pt}
        \begin{minipage}{0.065\linewidth}
			\vspace{1pt}
			\centerline{\includegraphics[width=\textwidth]{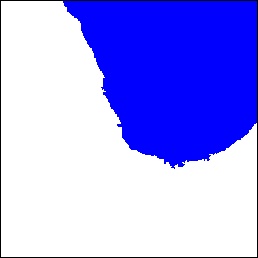}}
			\vspace{1pt}
                \centerline{\includegraphics[width=\textwidth]{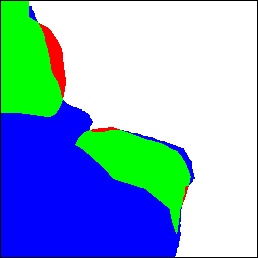}}
                \vspace{1pt}
                \centerline{\includegraphics[width=\textwidth]{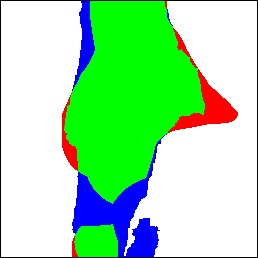}}
                \vspace{1pt}
                \centerline{\includegraphics[width=\textwidth]{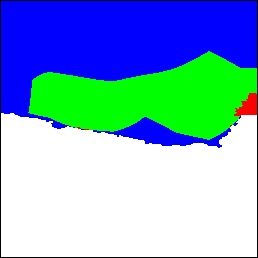}}
                \vspace{1pt}
                \centerline{\includegraphics[width=\textwidth]{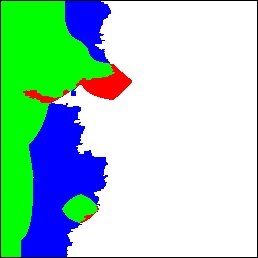}}
                \vspace{1pt}
                \centerline{\includegraphics[width=\textwidth]{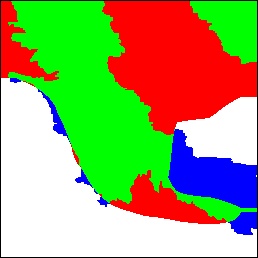}}
                \vspace{1pt}
                \centerline{\includegraphics[width=\textwidth]{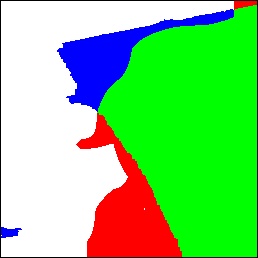}}
                \vspace{1pt}
                \centerline{\includegraphics[width=\textwidth]{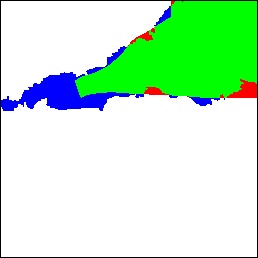}}
                \vspace{1pt}
                \centerline{\includegraphics[width=\textwidth]{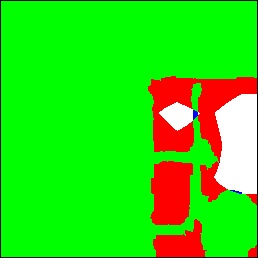}}
                \vspace{1pt}
                \centerline{\includegraphics[width=\textwidth]{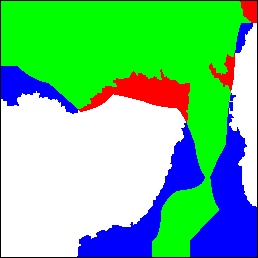}}
			\centerline{(c)}
		\end{minipage}
		\hspace{-6pt}
  \begin{minipage}{0.065\linewidth}
    \vspace{1pt}
    \centerline{\includegraphics[width=\textwidth]{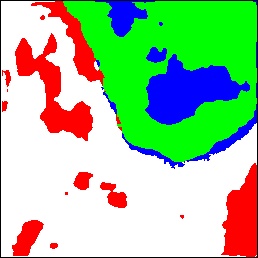}}
    \vspace{1pt}
        \centerline{\includegraphics[width=\textwidth]{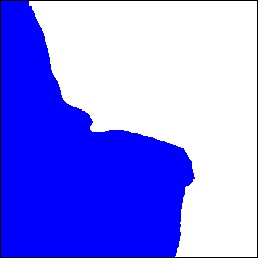}}
        \vspace{1pt}
        \centerline{\includegraphics[width=\textwidth]{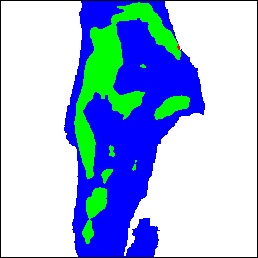}}
        \vspace{1pt}
        \centerline{\includegraphics[width=\textwidth]{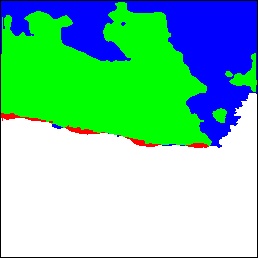}}
        \vspace{1pt}
        \centerline{\includegraphics[width=\textwidth]{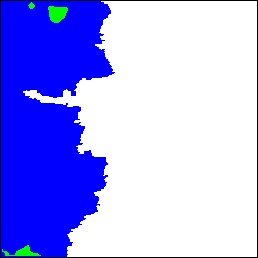}}
        \vspace{1pt}
        \centerline{\includegraphics[width=\textwidth]{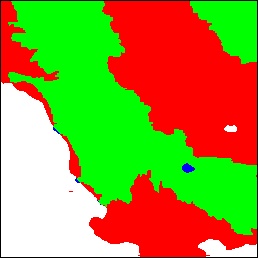}}
        \vspace{1pt}
        \centerline{\includegraphics[width=\textwidth]{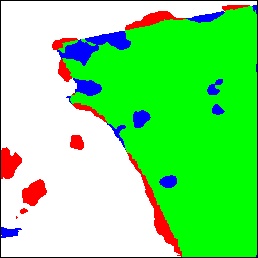}}
        \vspace{1pt}
        \centerline{\includegraphics[width=\textwidth]{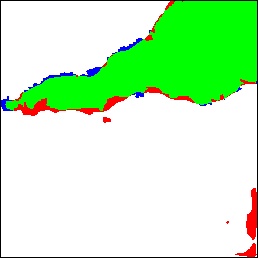}}
        \vspace{1pt}
        \centerline{\includegraphics[width=\textwidth]{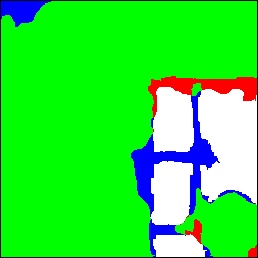}}
        \vspace{1pt}
        \centerline{\includegraphics[width=\textwidth]{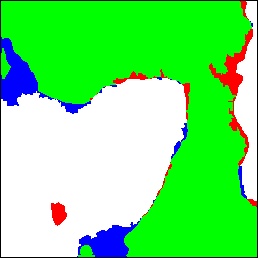}}
    \centerline{(d)}
\end{minipage}
		\hspace{-6pt}
  \begin{minipage}{0.065\linewidth}
    \vspace{1pt}
    \centerline{\includegraphics[width=\textwidth]{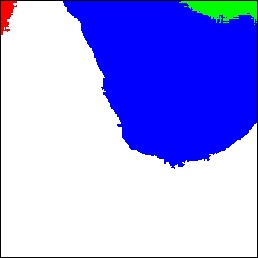}}
    \vspace{1pt}
        \centerline{\includegraphics[width=\textwidth]{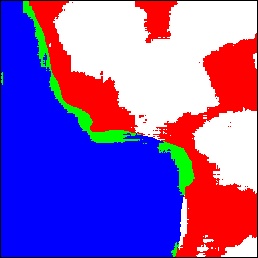}}
        \vspace{1pt}
        \centerline{\includegraphics[width=\textwidth]{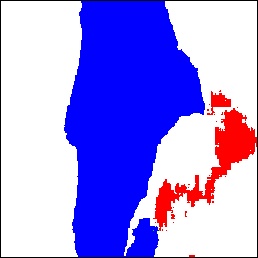}}
        \vspace{1pt}
        \centerline{\includegraphics[width=\textwidth]{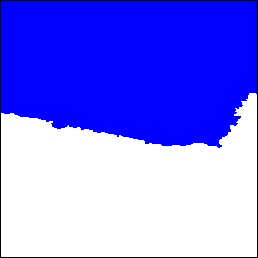}}
        \vspace{1pt}
        \centerline{\includegraphics[width=\textwidth]{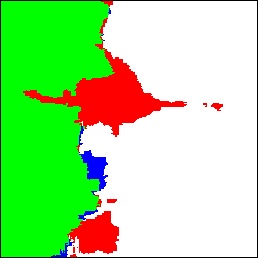}}
        \vspace{1pt}
        \centerline{\includegraphics[width=\textwidth]{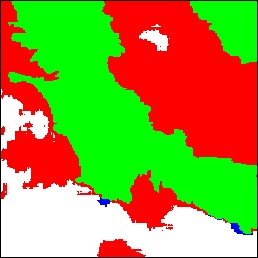}}
        \vspace{1pt}
        \centerline{\includegraphics[width=\textwidth]{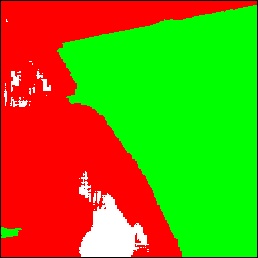}}
        \vspace{1pt}
        \centerline{\includegraphics[width=\textwidth]{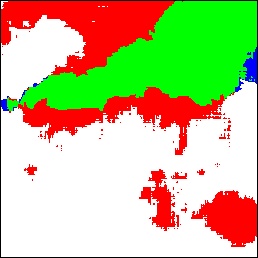}}
        \vspace{1pt}
        \centerline{\includegraphics[width=\textwidth]{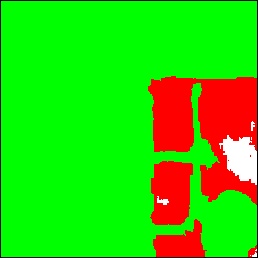}}
        \vspace{1pt}
        \centerline{\includegraphics[width=\textwidth]{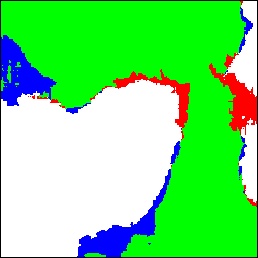}}
    \centerline{(e)}
\end{minipage}
		\hspace{-6pt}
  \begin{minipage}{0.065\linewidth}
    \vspace{1pt}
    \centerline{\includegraphics[width=\textwidth]{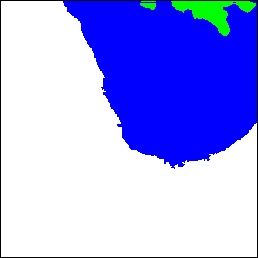}}
    \vspace{1pt}
        \centerline{\includegraphics[width=\textwidth]{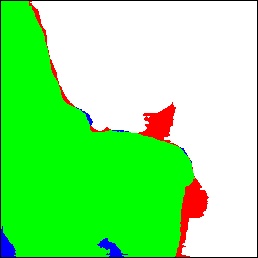}}
        \vspace{1pt}
        \centerline{\includegraphics[width=\textwidth]{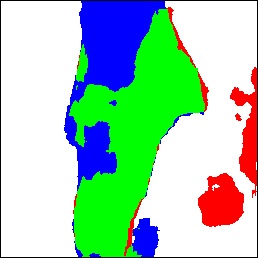}}
        \vspace{1pt}
        \centerline{\includegraphics[width=\textwidth]{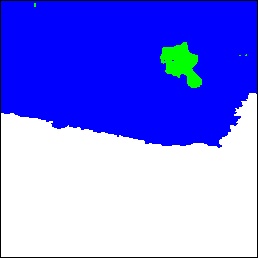}}
        \vspace{1pt}
        \centerline{\includegraphics[width=\textwidth]{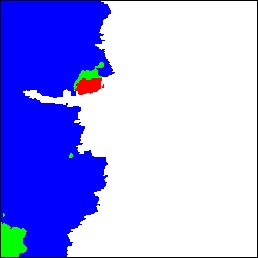}}
        \vspace{1pt}
        \centerline{\includegraphics[width=\textwidth]{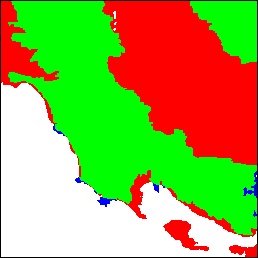}}
        \vspace{1pt}
        \centerline{\includegraphics[width=\textwidth]{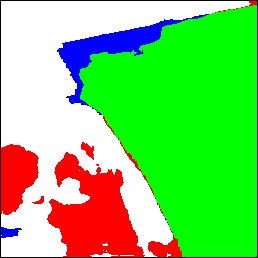}}
        \vspace{1pt}
        \centerline{\includegraphics[width=\textwidth]{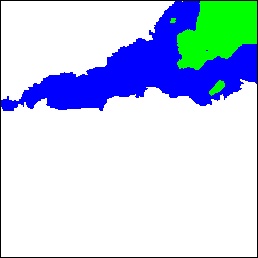}}
        \vspace{1pt}
        \centerline{\includegraphics[width=\textwidth]{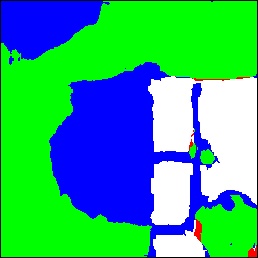}}
        \vspace{1pt}
        \centerline{\includegraphics[width=\textwidth]{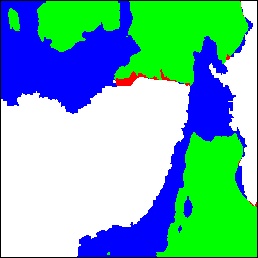}}
    \centerline{(f)}
\end{minipage}
		\hspace{-6pt}
  \begin{minipage}{0.065\linewidth}
    \vspace{1pt}
    \centerline{\includegraphics[width=\textwidth]{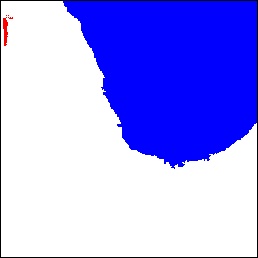}}
    \vspace{1pt}
        \centerline{\includegraphics[width=\textwidth]{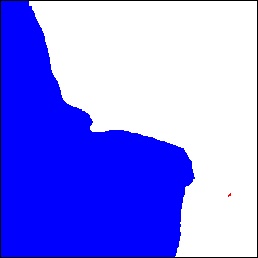}}
        \vspace{1pt}
        \centerline{\includegraphics[width=\textwidth]{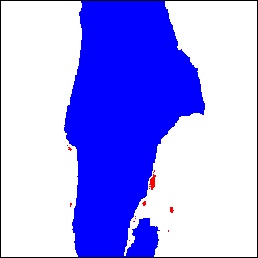}}
        \vspace{1pt}
        \centerline{\includegraphics[width=\textwidth]{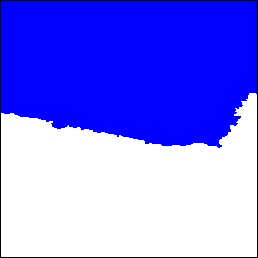}}
        \vspace{1pt}
        \centerline{\includegraphics[width=\textwidth]{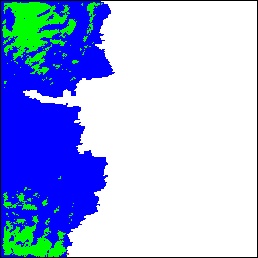}}
        \vspace{1pt}
        \centerline{\includegraphics[width=\textwidth]{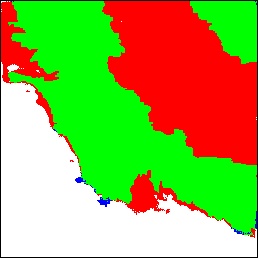}}
        \vspace{1pt}
        \centerline{\includegraphics[width=\textwidth]{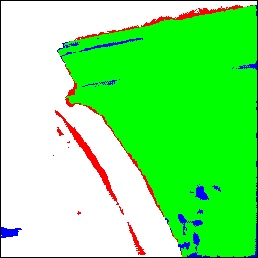}}
        \vspace{1pt}
        \centerline{\includegraphics[width=\textwidth]{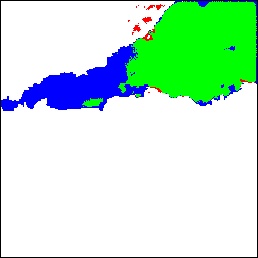}}
        \vspace{1pt}
        \centerline{\includegraphics[width=\textwidth]{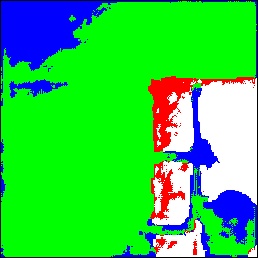}}
        \vspace{1pt}
        \centerline{\includegraphics[width=\textwidth]{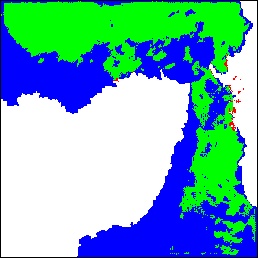}}
    \centerline{(g)}
\end{minipage}
		\hspace{-6pt}
 \begin{minipage}{0.065\linewidth}
    \vspace{1pt}
    \centerline{\includegraphics[width=\textwidth]{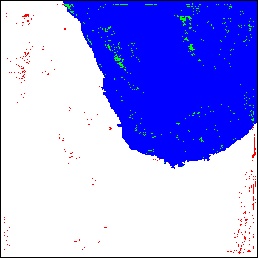}}
    \vspace{1pt}
        \centerline{\includegraphics[width=\textwidth]{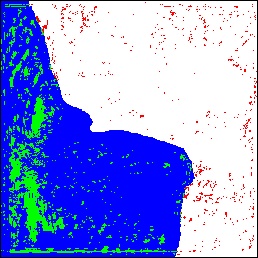}}
        \vspace{1pt}
        \centerline{\includegraphics[width=\textwidth]{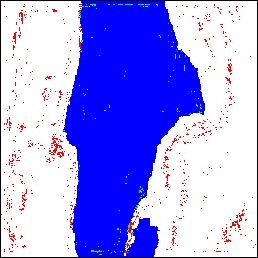}}
        \vspace{1pt}
        \centerline{\includegraphics[width=\textwidth]{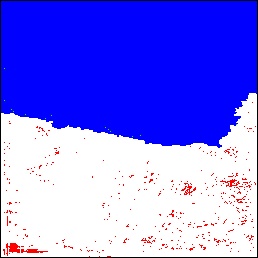}}
        \vspace{1pt}
        \centerline{\includegraphics[width=\textwidth]{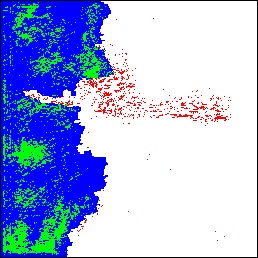}}
        \vspace{1pt}
        \centerline{\includegraphics[width=\textwidth]{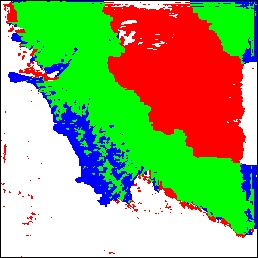}}
        \vspace{1pt}
        \centerline{\includegraphics[width=\textwidth]{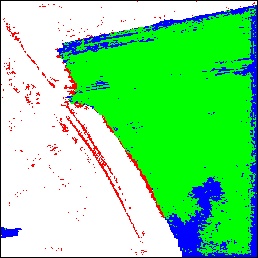}}
        \vspace{1pt}
        \centerline{\includegraphics[width=\textwidth]{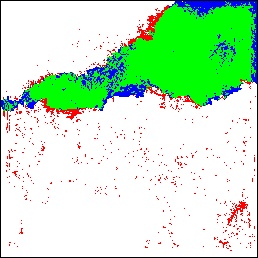}}
        \vspace{1pt}
        \centerline{\includegraphics[width=\textwidth]{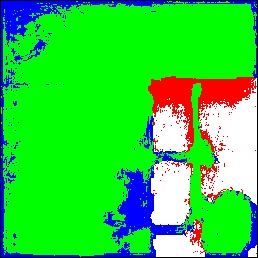}}
        \vspace{1pt}
        \centerline{\includegraphics[width=\textwidth]{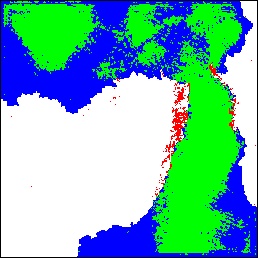}}
    \centerline{(h)}
\end{minipage}
		\hspace{-6pt}
\begin{minipage}{0.065\linewidth}
    \vspace{1pt}
    \centerline{\includegraphics[width=\textwidth]{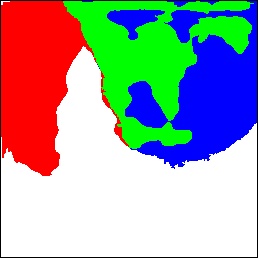}}
    \vspace{1pt}
        \centerline{\includegraphics[width=\textwidth]{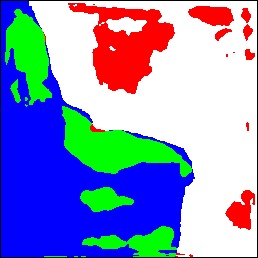}}
        \vspace{1pt}
        \centerline{\includegraphics[width=\textwidth]{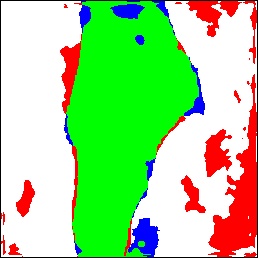}}
        \vspace{1pt}
        \centerline{\includegraphics[width=\textwidth]{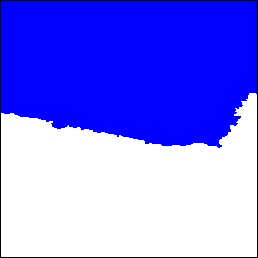}}
        \vspace{1pt}
        \centerline{\includegraphics[width=\textwidth]{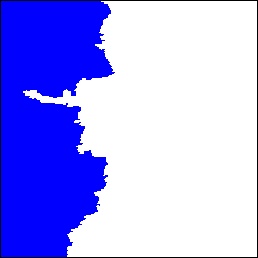}}
        \vspace{1pt}
        \centerline{\includegraphics[width=\textwidth]{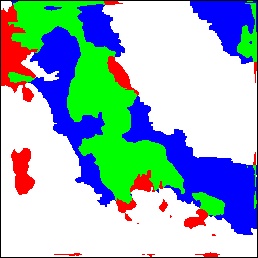}}
        \vspace{1pt}
        \centerline{\includegraphics[width=\textwidth]{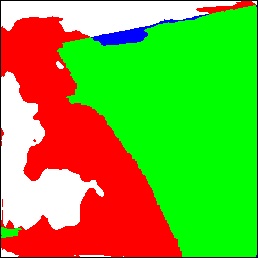}}
        \vspace{1pt}
        \centerline{\includegraphics[width=\textwidth]{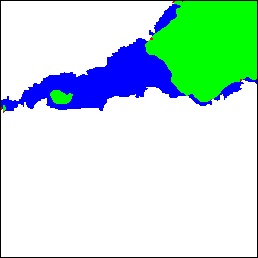}}
        \vspace{1pt}
        \centerline{\includegraphics[width=\textwidth]{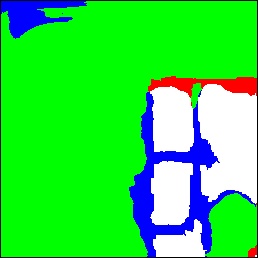}}
        \vspace{1pt}
        \centerline{\includegraphics[width=\textwidth]{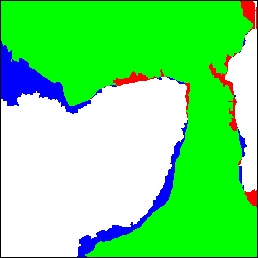}}
    \centerline{(i)}
\end{minipage}
		\hspace{-6pt}
  \begin{minipage}{0.065\linewidth}
    \vspace{1pt}
    \centerline{\includegraphics[width=\textwidth]{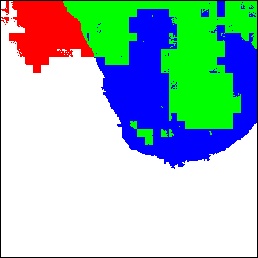}}
    \vspace{1pt}
        \centerline{\includegraphics[width=\textwidth]{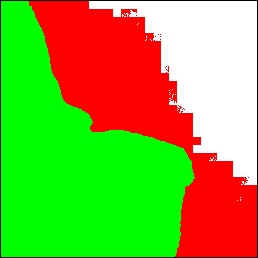}}
        \vspace{1pt}
        \centerline{\includegraphics[width=\textwidth]{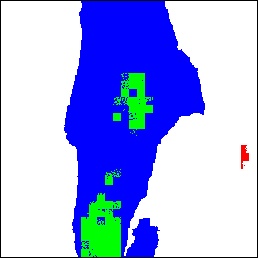}}
        \vspace{1pt}
        \centerline{\includegraphics[width=\textwidth]{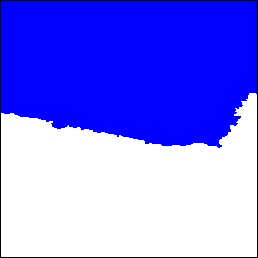}}
        \vspace{1pt}
        \centerline{\includegraphics[width=\textwidth]{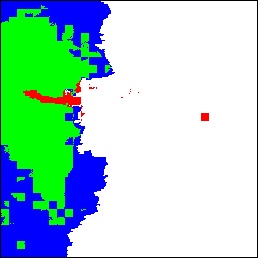}}
        \vspace{1pt}
        \centerline{\includegraphics[width=\textwidth]{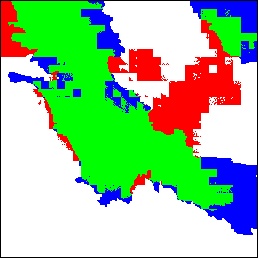}}
        \vspace{1pt}
        \centerline{\includegraphics[width=\textwidth]{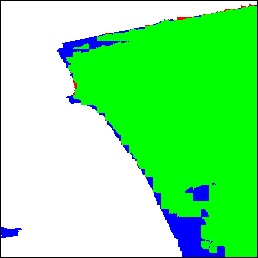}}
        \vspace{1pt}
        \centerline{\includegraphics[width=\textwidth]{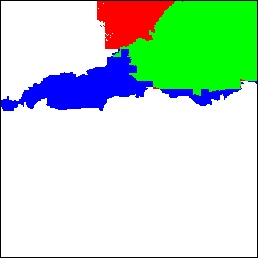}}
        \vspace{1pt}
        \centerline{\includegraphics[width=\textwidth]{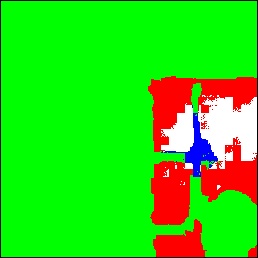}}
        \vspace{1pt}
        \centerline{\includegraphics[width=\textwidth]{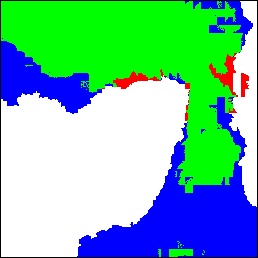}}
    \centerline{(j)}
\end{minipage}
		\hspace{-6pt}
  \begin{minipage}{0.065\linewidth}
    \vspace{1pt}
    \centerline{\includegraphics[width=\textwidth]{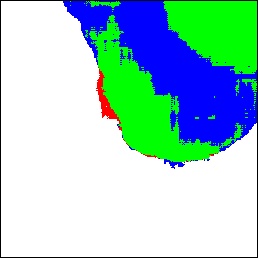}}
    \vspace{1pt}
        \centerline{\includegraphics[width=\textwidth]{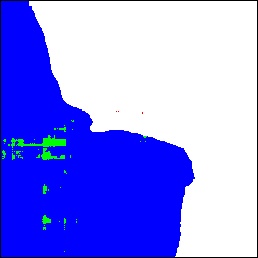}}
        \vspace{1pt}
        \centerline{\includegraphics[width=\textwidth]{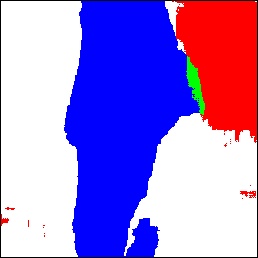}}
        \vspace{1pt}
        \centerline{\includegraphics[width=\textwidth]{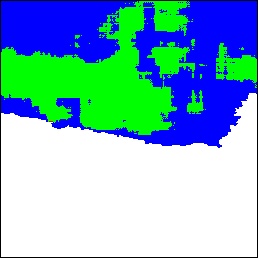}}
        \vspace{1pt}
        \centerline{\includegraphics[width=\textwidth]{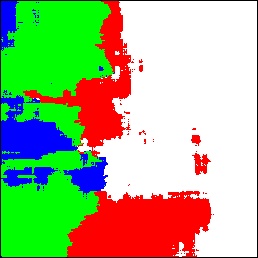}}
        \vspace{1pt}
        \centerline{\includegraphics[width=\textwidth]{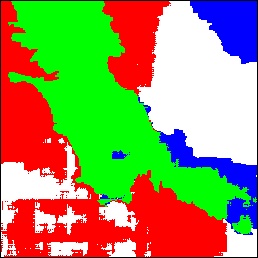}}
        \vspace{1pt}
        \centerline{\includegraphics[width=\textwidth]{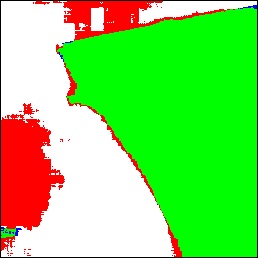}}
        \vspace{1pt}
        \centerline{\includegraphics[width=\textwidth]{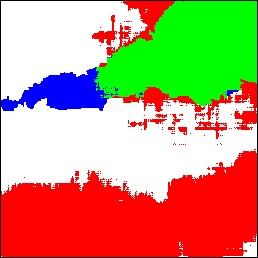}}
        \vspace{1pt}
        \centerline{\includegraphics[width=\textwidth]{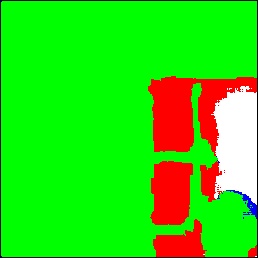}}
        \vspace{1pt}
        \centerline{\includegraphics[width=\textwidth]{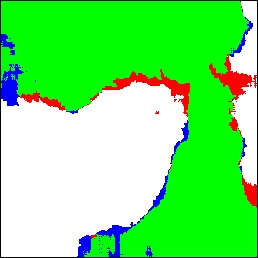}}
    \centerline{(k)}
\end{minipage}
\hspace{-6pt}
\begin{minipage}{0.065\linewidth}
    \vspace{1pt}
    \centerline{\includegraphics[width=\textwidth]{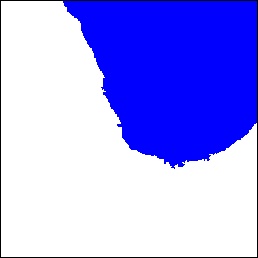}}
    \vspace{1pt}
        \centerline{\includegraphics[width=\textwidth]{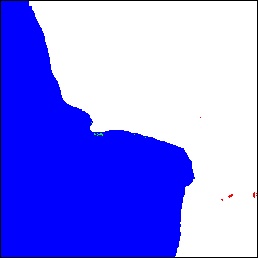}}
        \vspace{1pt}
        \centerline{\includegraphics[width=\textwidth]{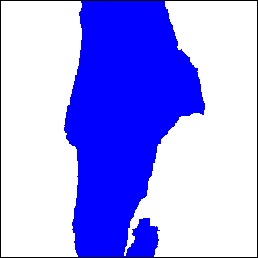}}
        \vspace{1pt}
        \centerline{\includegraphics[width=\textwidth]{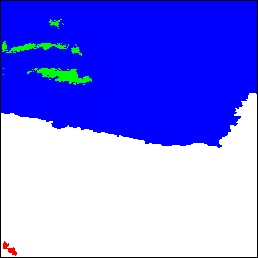}}
        \vspace{1pt}
        \centerline{\includegraphics[width=\textwidth]{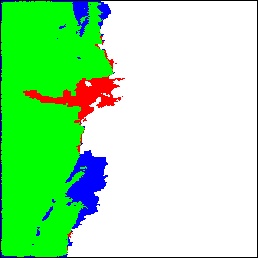}}
        \vspace{1pt}
        \centerline{\includegraphics[width=\textwidth]{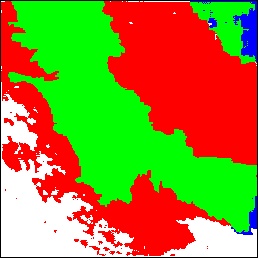}}
        \vspace{1pt}
        \centerline{\includegraphics[width=\textwidth]{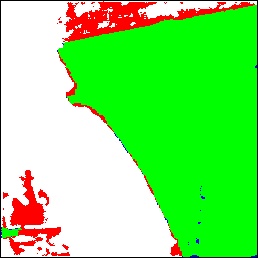}}
        \vspace{1pt}
        \centerline{\includegraphics[width=\textwidth]{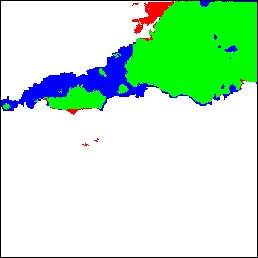}}
        \vspace{1pt}
        \centerline{\includegraphics[width=\textwidth]{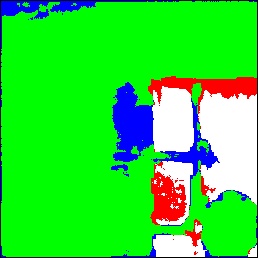}}
        \vspace{1pt}
        \centerline{\includegraphics[width=\textwidth]{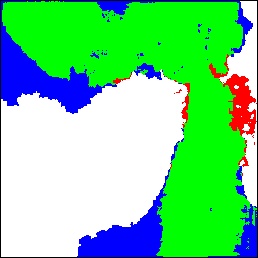}}
    \centerline{(l)}
\end{minipage}
\hspace{-6pt}
\begin{minipage}{0.065\linewidth}
    \vspace{1pt}
    \centerline{\includegraphics[width=\textwidth]{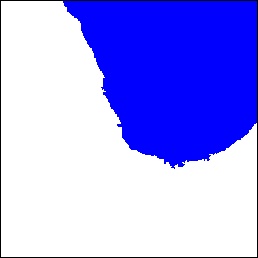}}
    \vspace{1pt}
        \centerline{\includegraphics[width=\textwidth]{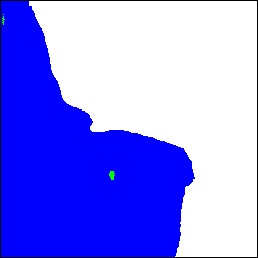}}
        \vspace{1pt}
        \centerline{\includegraphics[width=\textwidth]{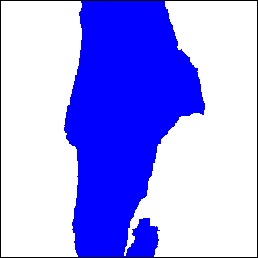}}
        \vspace{1pt}
        \centerline{\includegraphics[width=\textwidth]{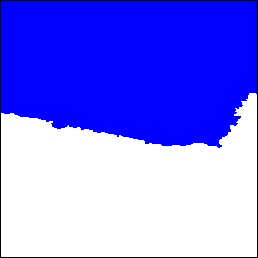}}
        \vspace{1pt}
        \centerline{\includegraphics[width=\textwidth]{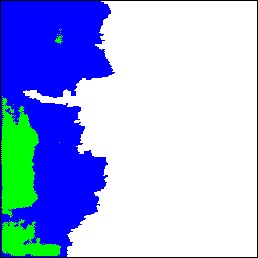}}
        \vspace{1pt}
        \centerline{\includegraphics[width=\textwidth]{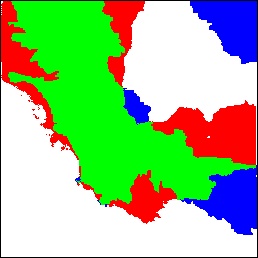}}
        \vspace{1pt}
        \centerline{\includegraphics[width=\textwidth]{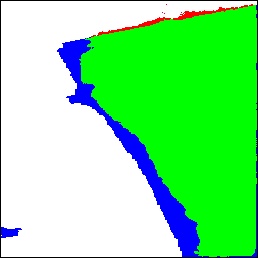}}
        \vspace{1pt}
        \centerline{\includegraphics[width=\textwidth]{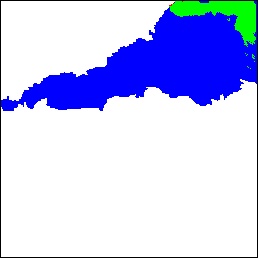}}
        \vspace{1pt}
        \centerline{\includegraphics[width=\textwidth]{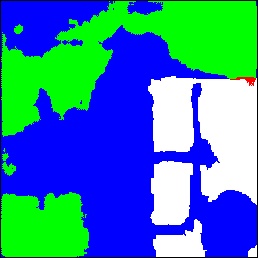}}
        \vspace{1pt}
        \centerline{\includegraphics[width=\textwidth]{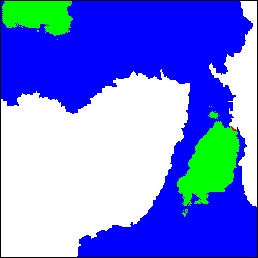}}
    \centerline{(m)}
\end{minipage}
		\hspace{-6pt}
\begin{minipage}{0.065\linewidth}
    \vspace{1pt}
    \centerline{\includegraphics[width=\textwidth]{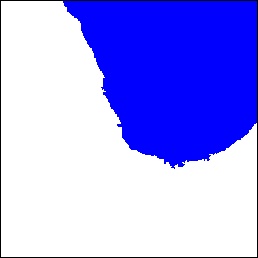}}
    \vspace{1pt}
        \centerline{\includegraphics[width=\textwidth]{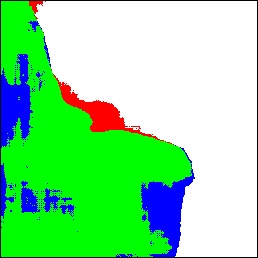}}
        \vspace{1pt}
        \centerline{\includegraphics[width=\textwidth]{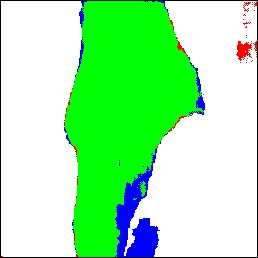}}
        \vspace{1pt}
        \centerline{\includegraphics[width=\textwidth]{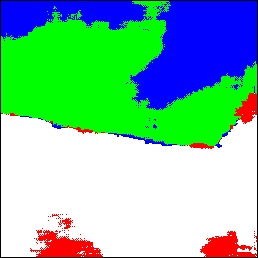}}
        \vspace{1pt}
        \centerline{\includegraphics[width=\textwidth]{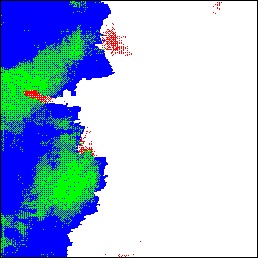}}
        \vspace{1pt}
        \centerline{\includegraphics[width=\textwidth]{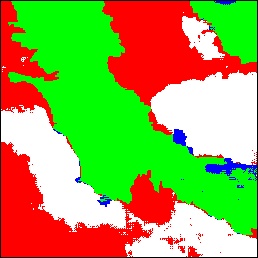}}
        \vspace{1pt}
        \centerline{\includegraphics[width=\textwidth]{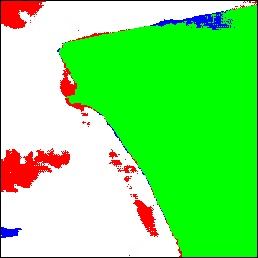}}
        \vspace{1pt}
        \centerline{\includegraphics[width=\textwidth]{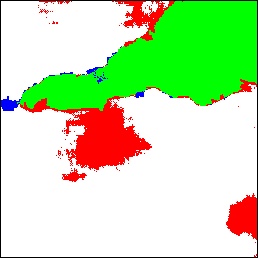}}
        \vspace{1pt}
        \centerline{\includegraphics[width=\textwidth]{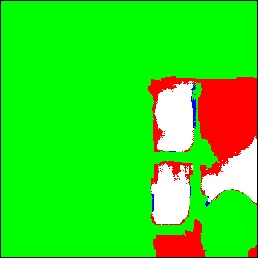}}
        \vspace{1pt}
        \centerline{\includegraphics[width=\textwidth]{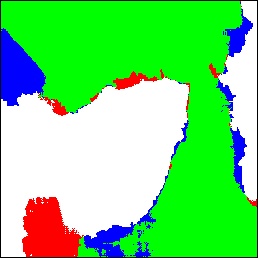}}
    \centerline{(n)}
\end{minipage}
		\hspace{-6pt}
  \begin{minipage}{0.065\linewidth}
    \vspace{1pt}
    \centerline{\includegraphics[width=\textwidth]{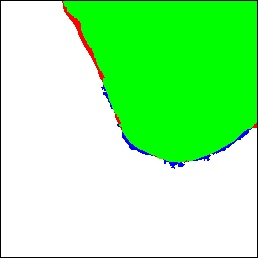}}
    \vspace{1pt}
        \centerline{\includegraphics[width=\textwidth]{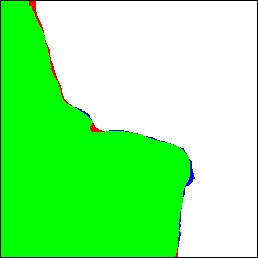}}
        \vspace{1pt}
        \centerline{\includegraphics[width=\textwidth]{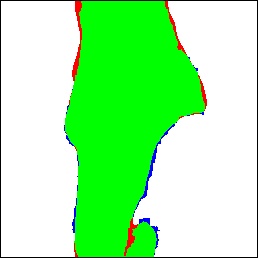}}
        \vspace{1pt}
        \centerline{\includegraphics[width=\textwidth]{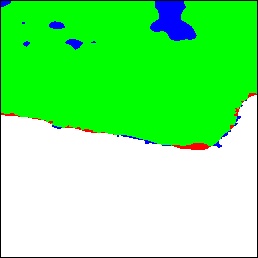}}
        \vspace{1pt}
        \centerline{\includegraphics[width=\textwidth]{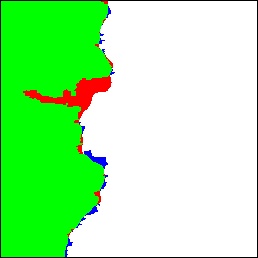}}
        \vspace{1pt}
        \centerline{\includegraphics[width=\textwidth]{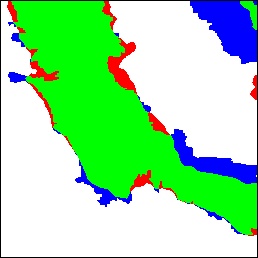}}
        \vspace{1pt}
        \centerline{\includegraphics[width=\textwidth]{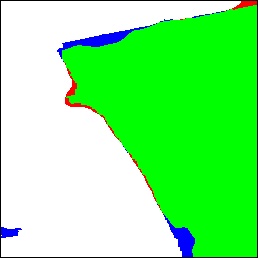}}
        \vspace{1pt}
        \centerline{\includegraphics[width=\textwidth]{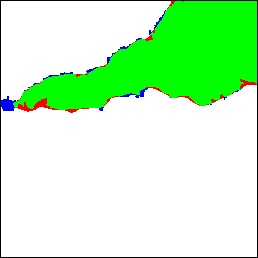}}
        \vspace{1pt}
        \centerline{\includegraphics[width=\textwidth]{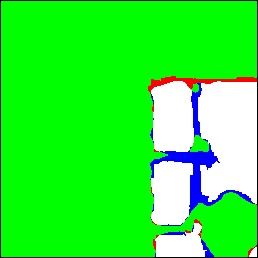}}
        \vspace{1pt}
        \centerline{\includegraphics[width=\textwidth]{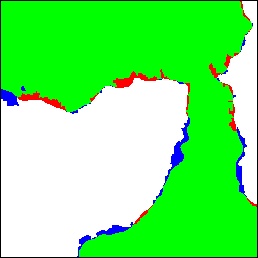}}
    \centerline{(o)}
\end{minipage}
\hspace{-6pt}
		\caption{Visual comparisons of the proposed method and the state-of-the-art approaches on the MineNetCD dataset. (a) Pre-change images. (b) Post-change images. (c) A2Net. (d) BIT. (e) ChangeFormer. (f) DMINet. (g) FC-EF (h) FCNPP (i) ICIFNet (j) RDPNet (k) ResUnet (l) SiamUnet-Conc (m) SiamUnet-Diff (n) SNUNet (o) MineNetCD. The rendered colors represent true positives (green), true negatives (white), false positives (red), and false negatives (blue).} 
		\label{fig:quali256}
	\end{figure*}
Fig. \ref{fig:quali1} and Fig. \ref{fig:quali2} show examples of MineNetCD results obtained using various comparative methods. Additionally, we present the results for original small patches sized $256\times 256$ to enable a comprehensive quantitative comparison. The visualizations reveal that most comparison methods struggle to accurately detect the correct changed areas while avoiding false alarms.

Based on these representative results, the proposed MineNetCD method demonstrates advantages in several aspects:
\begin{itemize}
    \item Better discrimination of changes:
    In the first five rows of Fig. \ref{fig:quali256}, most comparison methods fail to distinguish the changes in bi-temporal images, resulting in change maps with numerous omitted changes (false negatives, shown in blue). 
    Conversely, the MineNetCD method consistently produces accurate change maps from these challenging samples, highlighting its superior ability to differentiate changes in bi-temporal images with heterogeneous spectral characteristics (e.g., rows one to three in Fig. \ref{fig:quali256}) and subtle spectral differences (e.g., rows four and five in Fig. \ref{fig:quali256}).
    This superior discrimination of the MineNetCD model is attributed to the ChangeFFT module, which aligns spatiotemporal feature differences by unifying their spectra in the frequency domain to obtain robust change-aware representations.
    \item Fewer pseudo changes: 
    Compared to other methods, the MineNetCD model significantly reduces pseudo changes—false alarms or irrelevant changes to mining development. In Fig. \ref{fig:quali2}, change maps generated by other methods exhibit extensive pseudo changes (denoted as red patches). In the two six of Fig. \ref{fig:quali256}, most methods predict pseudo changes in mining areas already present in the pre-change image. In contrast, the change maps from the MineNetCD model exhibit fewer pseudo changes, benefiting from the ChangeFFT module's ability to filter out noise in bi-temporal feature differences by processing their feature spectra in the frequency domain.
    \item More precise boundaries:
    Mining change detection often involves irregularly shaped instances due to the nature of open-pit mining developments. It is crucial to accurately capture the extent of the mining area. In Fig. \ref{fig:quali1} and rows eight to ten of Fig. \ref{fig:quali256}, the change maps predicted by the comparison methods appear to have incorrect and unclear instance edges, while the MineNetCD model managed to capture precise edges of change instances in mining areas.
    This precision is achieved through the change-aware representation learning in the ChangeFFT module, which accurately extracts spatiotemporal dependencies in the frequency domain.
\end{itemize}

The changes occurring in the benchmark set are very diverse and relevant for both the monitoring of the mining activities and their relevance for the environment and the local communities. The changes occur in the pits themselves, where extraction occurs, but also in the waste piles and tailings, where wastes are deposited, or the different processing units surrounding the mines. This is important because the impacts of mining vary from commodity to commodity, processing types, and environmental conditions. Only a holistic approach to mining changes can provide sufficient information to monitor and address the different impacts of mining sites.

\subsection{Comparisons of Different Backbone Models for Change Detection}
\begin{figure*}[!htp]
    \captionsetup{singlelinecheck=false}
    \centering
    \includegraphics[width=\linewidth]{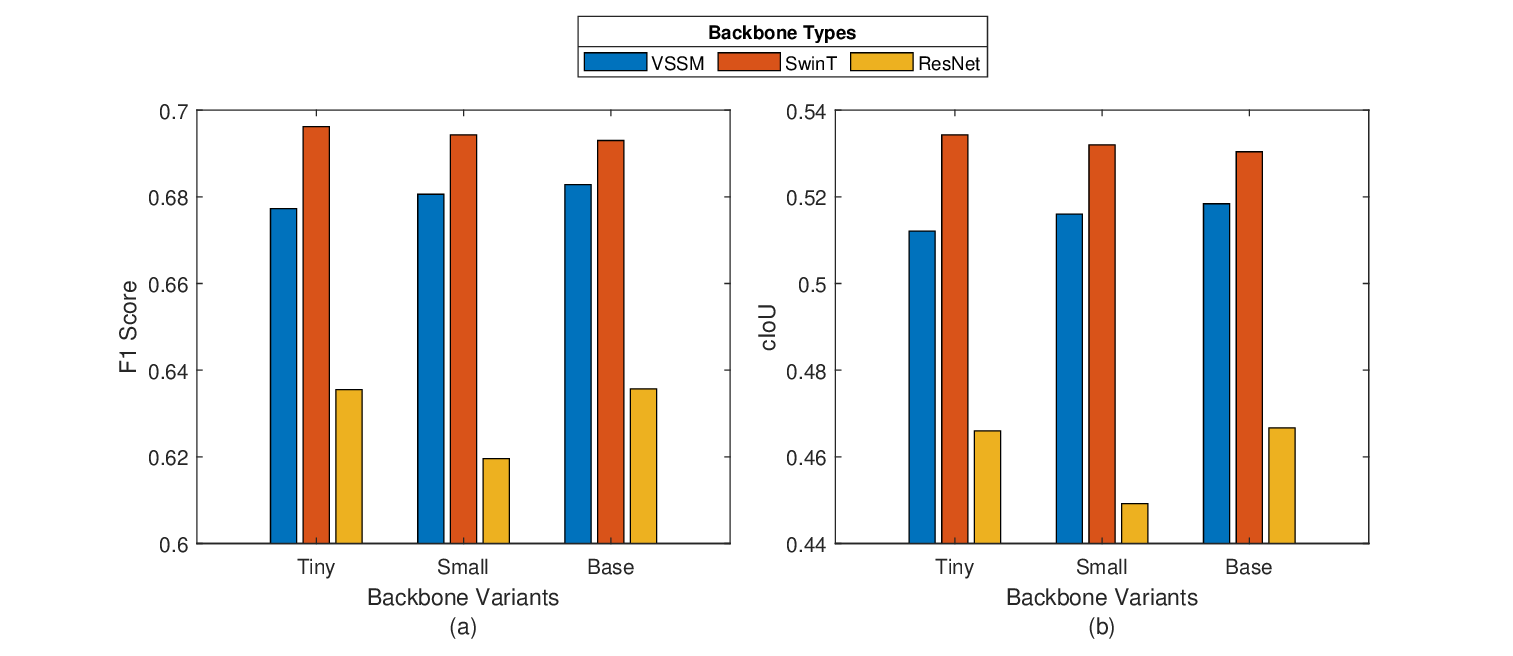}
    \caption{Comparison results of MineNetCD with different backbone models, including VMamba, Swin Transformer (SwinT), and ResNet on the MineNetCD dataset. The tiny, small, and base scales of the ResNet model indicate the ResNet-18, ResNet-50, and ResNet-101, respectively. (a) and (b) display the results in terms of F1 score and cIoU metrics, respectively.}
    \label{fig:backboneanalysis}
\end{figure*}
To explore the influence of different representation learners on the performance of change detection, we apply three distinct branches of backbone models, each with three variants at different scales, to build the modularized Siamese encoder of the MineNet model. As shown in Fig. \ref{fig:backboneanalysis}, the Swin Transformer achieves the best performance on MineNetCD in terms of F1 score and cIoU among all the scales of the backbone variants. In particular, the highest performance is achieved by the Swin Transformer Tiny, indicating that the deep features extracted by a larger self-attention-based backbone may be too abstract to distinguish changes in MineNetCD.

\begin{figure}
\centering
\begin{subfigure}{.12\linewidth}
            \centering
            \includegraphics[width=\textwidth]{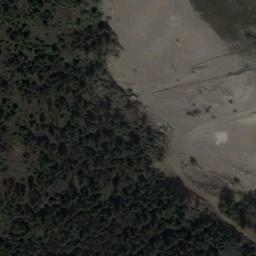}
            \caption{Pre-change}
\end{subfigure}
\begin{subfigure}{.12\linewidth}
            \centering
            \includegraphics[width=\textwidth]{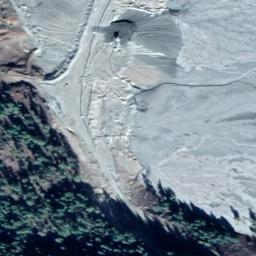}
            \caption{Post-change}
\end{subfigure}
\begin{subfigure}{.12\linewidth}
            \centering
            \includegraphics[width=\textwidth]{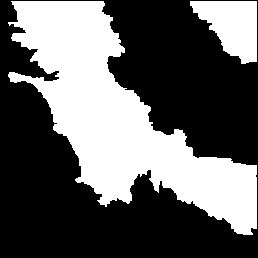}
            \caption{label}
\end{subfigure}
\begin{subfigure}{.12\linewidth}
            \centering
            \includegraphics[width=\textwidth]{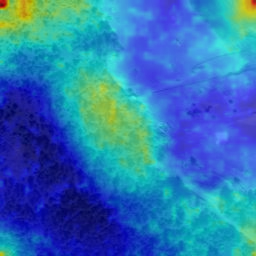}
            \caption{ResNet-18}
\end{subfigure}

\begin{subfigure}{.12\linewidth}
            \centering
            \includegraphics[width=\textwidth]{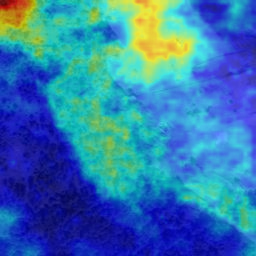}
            \caption{ResNet-50}
\end{subfigure}
\begin{subfigure}{.12\linewidth}
            \centering
         \includegraphics[width=\textwidth]{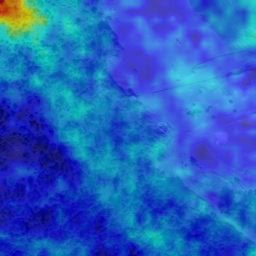}
            \caption{ResNet-101}
\end{subfigure}
\begin{subfigure}{.12\linewidth}
            \centering
            \includegraphics[width=\textwidth]{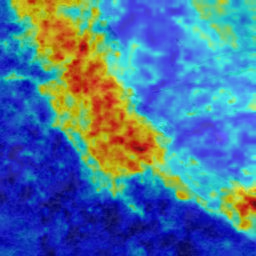}
            \caption{Swin-T}
\end{subfigure}
\begin{subfigure}{.12\linewidth}
            \centering
            \includegraphics[width=\textwidth]{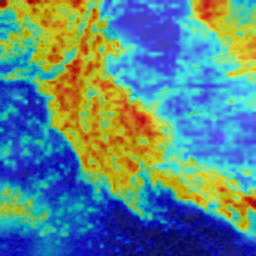}
            \caption{Swin-S}
\end{subfigure}

\begin{subfigure}{.12\linewidth}
            \centering
            \includegraphics[width=\textwidth]{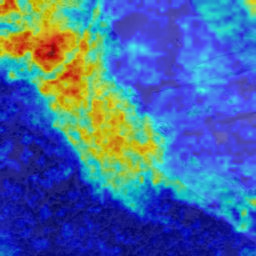}
            \caption{Swin-B}
\end{subfigure}
\begin{subfigure}{.12\linewidth}
            \centering
            \includegraphics[width=\textwidth]{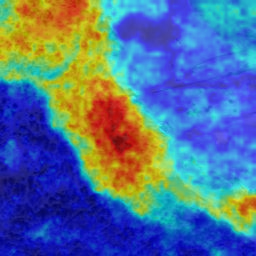}
            \caption{VMamba-T}
\end{subfigure}
\begin{subfigure}{.12\linewidth}
            \centering
            \includegraphics[width=\textwidth]{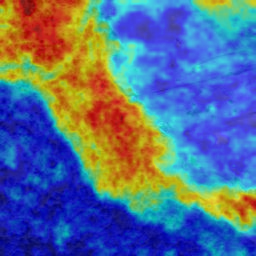}
            \caption{VMamba-S}
\end{subfigure}
\begin{subfigure}{.12\linewidth}
            \centering
            \includegraphics[width=\textwidth]{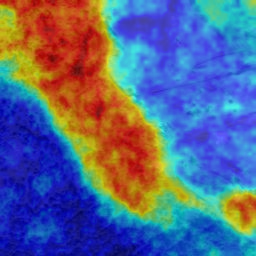}
            \caption{VMamba-B}
\end{subfigure}
\caption{GradCAM visualizations of change-aware representations using different backbone variants on MineNetCD.}
\label{fig:backbone_cam}
\end{figure}

Furthermore, we leverage the GradCAM \cite{selvaraju2017grad} technique to visualize the change-aware representations of different backbone variants, as shown in Fig. \ref{fig:backbone_cam}. It can be observed that the ResNet variants cannot extract effective features among the changed areas, as shown on the label. Furthermore, the heatmaps of the change-aware representations generated by the VMamba series have over-smooth boundaries, which cannot precisely characterize the extent of changed areas. In contrast, the Swin Transformer series can precisely capture the change-aware representations with sharp boundaries, indicating its ability to process the bi-temporal RS images into effective features with the ChangeFFT module.

\subsection{Ablation studies}
\begin{figure}
		\centering
		\begin{minipage}{0.10\linewidth}
			\vspace{1pt}
			\centerline{\includegraphics[width=\textwidth]{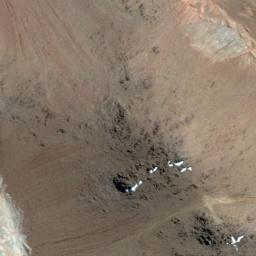}}
			\vspace{1pt}
                \centerline{\includegraphics[width=\textwidth]{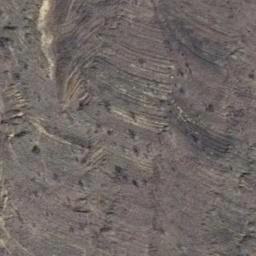}}
                \vspace{1pt}
                \centerline{\includegraphics[width=\textwidth]{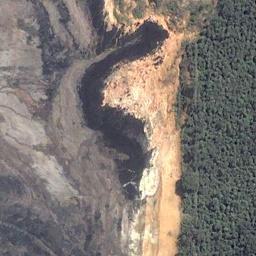}}
                \vspace{1pt}
                \centerline{\includegraphics[width=\textwidth]{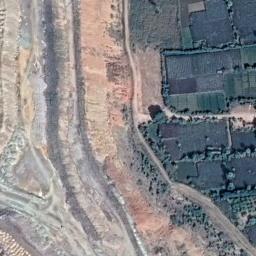}}
                \vspace{1pt}
                \centerline{\includegraphics[width=\textwidth]{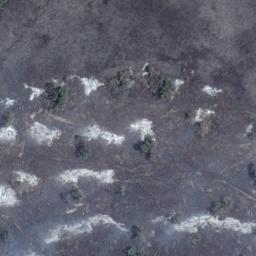}}
			\centerline{(a)}
		\end{minipage}
        \begin{minipage}{0.10\linewidth}
			\vspace{1pt}
			\centerline{\includegraphics[width=\textwidth]{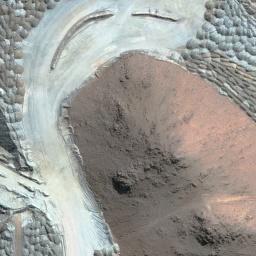}}
			\vspace{1pt}
                \centerline{\includegraphics[width=\textwidth]{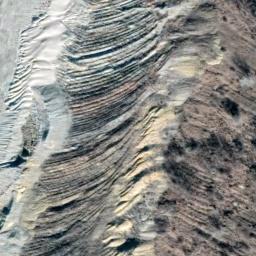}}
                \vspace{1pt}
                \centerline{\includegraphics[width=\textwidth]{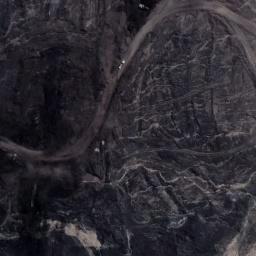}}
                \vspace{1pt}
                \centerline{\includegraphics[width=\textwidth]{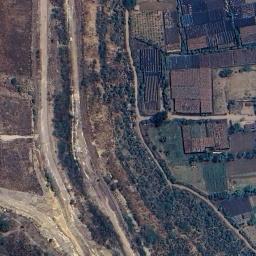}}
                \vspace{1pt}
                \centerline{\includegraphics[width=\textwidth]{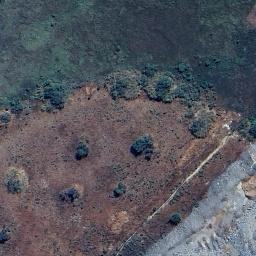}}
			\centerline{(b)}
		\end{minipage}
        \begin{minipage}{0.10\linewidth}
			\vspace{1pt}
			\centerline{\includegraphics[width=\textwidth]{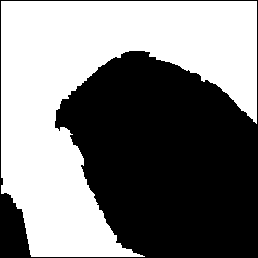}}
			\vspace{1pt}
                \centerline{\includegraphics[width=\textwidth]{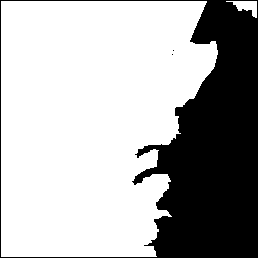}}
                \vspace{1pt}
                \centerline{\includegraphics[width=\textwidth]{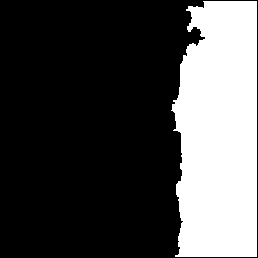}}
                \vspace{1pt}
                \centerline{\includegraphics[width=\textwidth]{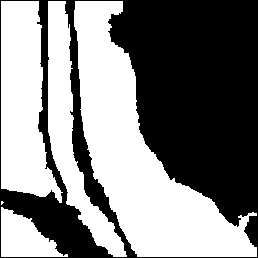}}
                \vspace{1pt}
                \centerline{\includegraphics[width=\textwidth]{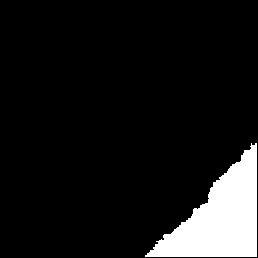}}
			\centerline{(c)}
		\end{minipage}
        \begin{minipage}{0.10\linewidth}
			\vspace{1pt}
			\centerline{\includegraphics[width=\textwidth]{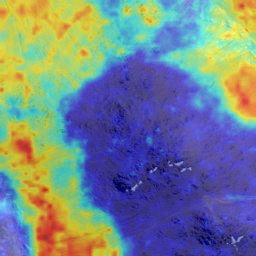}}
			\vspace{1pt}
                \centerline{\includegraphics[width=\textwidth]{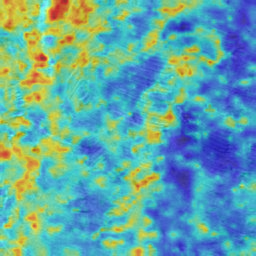}}
                \vspace{1pt}
                \centerline{\includegraphics[width=\textwidth]{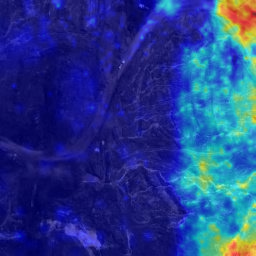}}
                \vspace{1pt}
                \centerline{\includegraphics[width=\textwidth]{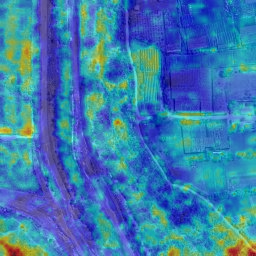}}
                \vspace{1pt}
                \centerline{\includegraphics[width=\textwidth]{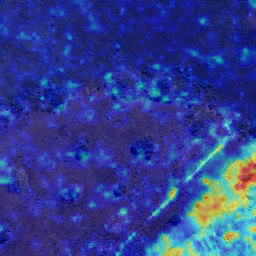}}
			\centerline{(d)}
		\end{minipage}
        \begin{minipage}{0.10\linewidth}
			\vspace{1pt}
			\centerline{\includegraphics[width=\textwidth]{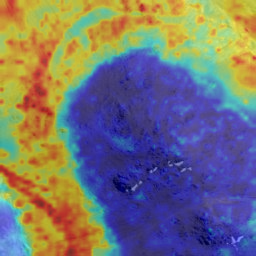}}
			\vspace{1pt}
                \centerline{\includegraphics[width=\textwidth]{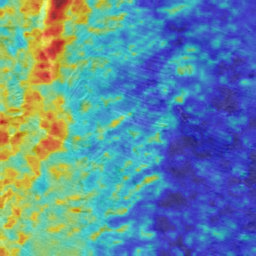}}
                \vspace{1pt}
                \centerline{\includegraphics[width=\textwidth]{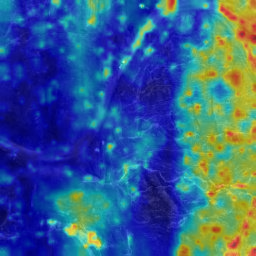}}
                \vspace{1pt}
                \centerline{\includegraphics[width=\textwidth]{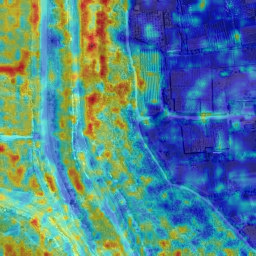}}
                \vspace{1pt}
                \centerline{\includegraphics[width=\textwidth]{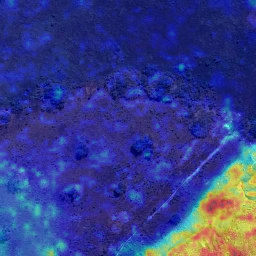}}
			\centerline{(e)}
		\end{minipage}
		\caption{Comparison of GradCAM visualizations of feature differences and change-aware representations: (a) Pre-change images, (b) Post-change images, (c) labels, (d)  Bi-temporal deep feature differences, and (e) Change-aware representations learned by ChangeFFT.} 
\label{fig:fft_ablation}
\end{figure}
\begin{table}[]
\centering
\caption{Performance improvements of ChangeFFT module based on different backbone variants.}
\label{table:ablation_changefft}
\begin{tabular}{@{}c|c|cc@{}}
\toprule
Backbones & Variants & F1 & cIoU \\ \midrule
\multirow{3}{*}{VMamba} &  VMamba-S & +0.0163 & +0.0184 \\
 & VMamba-T & +0.0055 & +0.0061 \\
 & VMamba-B & +0.0234 & +0.0262 \\ \midrule
\multirow{3}{*}{Swin Transformer} 
 & Swin-T & +0.0340 & +0.0389 \\
 & Swin-S & +0.0219 & +0.0253 \\
 & Swin-B & +0.0226 & +0.0258 \\ \midrule
\multirow{3}{*}{ResNet} 
 & ResNet-18 & +0.0098 & +0.0106 \\ 
 & ResNet-50 & +0.0098 & +0.0100 \\
 & ResNet-101 & +0.0085 & +0.0089 \\ \bottomrule
\end{tabular}
\end{table}
To evaluate the effectiveness of the ChangeFFT module, we conducted extensive ablation studies to build the MineNetCD model without using the ChangeFFT module. In this case, the multi-scale deep feature differences are directly fed into the UperNet-based change decoder to reconstruct a change map. Our experiments demonstrated that the ChangeFFT module improves performance in all settings based on various backbones, as shown in Table \ref{table:ablation_changefft}. In particular, incorporating the ChangeFFT module improves up to $3.40\%$ F1 score and $3.89\%$ cIoU based on the Swin Transformer base backbone. The ChangeFFT module is also effective on the VMamba series and ResNet series, indicating that the ChangeFFT module possesses superior compatibility and the potential to become a plug-and-play module in more change detection approaches.

We also display the GradCAM visualization to compare the conventional feature difference methods and the change-aware representations based on the Swin Tranformer base backbone, as shown in Fig. \ref{fig:fft_ablation}. It can be seen in the results that the ChangeFFT module significantly refines the feature difference and obtains change-aware representations highly focused on the changed areas. Specifically, the change-aware representations have a more accurate mapping of ground instances and reduce the noises within the feature differences.

\section{Conclusion}\label{sec6}
In this article, we introduce MineNetCD, a comprehensive benchmark for global mining change detection aimed at promoting  the  sustainability goals of the extractive industry. MineNetCD involves a global mining change detection dataset, a change-aware fast Fourier transform (ChangeFFT) module-based baseline model, and a unified change detection (UCD) framework. The dataset includes bi-temporal high-resolution remote sensing images and pixel-level annotations of changed areas from 100 mining sites worldwide. Our baseline model, utilizing the ChangeFFT module, achieves better than state-of-the-art performance by enabling change-aware representation learning in the frequency domain. The UCD framework features over 70 pretrained models based on 5 change detection datasets, democratizing access to deep learning-based change detection methods without requiring specialized expertise.

With this work, we provide a robust solution for mining change detection, which is instrumental for environmental impact assessments in the mining industry using geospatial data. We anticipate that MineNetCD will pioneer advancements in sustainable metal sourcing and inspire future research focused on environmentally responsible mining practices.


%



\section*{Acknowledgment}

The authors would like to thank Mr. Yasong Shi and Mr. Yilong Deng for their efforts in building the MineNetCD dataset. This work was supported by the European Regional Development Fund and the Land of Saxony by providing the high specification Nvidia A100 GPU server that we used in our experiments.

\bibliographystyle{IEEEtranN}
\bibliography{Ref}
\end{document}